\pdfoutput=1

\documentclass[11pt]{article}

\usepackage[final]{acl}

\usepackage{float}
\usepackage{times}
\usepackage{bbm}
\usepackage{latexsym}
\usepackage{amssymb} 
\usepackage{mathrsfs}
\usepackage{hyperref}
\usepackage{amsmath}
\usepackage{tabularx} 
\usepackage[title]{appendix}
\usepackage{mwe}
\usepackage{wrapfig}
\usepackage{caption}
\usepackage{subcaption}
\usepackage{booktabs}
\usepackage{xcolor}
\usepackage{graphicx}
\usepackage{pifont}
\usepackage{multirow} 
\usepackage{xcolor,colortbl}
\usepackage{graphicx}
\usepackage{xcolor}
\usepackage{times}
\usepackage{latexsym}
\usepackage[ruled,vlined,linesnumbered]{algorithm2e}
\usepackage[detect-none]{siunitx}
\newcommand{\method}{{MANU}\xspace} 
\newcommand\sbullet[1][.5]{\mathbin{\vcenter{\hbox{\scalebox{#1}{$\bullet$}}}}}

\usepackage[T1]{fontenc}

\usepackage[utf8]{inputenc}

\usepackage{microtype}

\usepackage{inconsolata}

\usepackage{graphicx}

\title{Modality-Aware Neuron Pruning for Unlearning in \\Multimodal Large Language Models}

\author{Zheyuan Liu$^1$, Guangyao Dou$^2$, Xiangchi Yuan$^3$, Chunhui Zhang$^4$, \\
 \textbf{{Zhaoxuan Tan}$^1$, {Meng Jiang}$^1$} \\
 $^1$University of Notre Dame,
 $^2$University of Pennsylvania,\\
 $^3$Georgia Institute of Technology,
 $^4$Dartmouth College\\
 {\tt zliu29@nd.edu}
}

\begin{document}
\maketitle
\begin{abstract}

Generative models such as Large Language Models (LLMs) and Multimodal Large Language Models (MLLMs) trained on massive datasets can lead them to memorize and inadvertently reveal sensitive information, raising ethical and privacy concerns. While some prior works have explored this issue in the context of LLMs, it presents a unique challenge for MLLMs due to the entangled nature of knowledge across modalities, making comprehensive unlearning more difficult.
To address this challenge, we propose \textbf{Modality Aware Neuron Unlearning (\method)}, a novel unlearning framework for MLLMs designed to selectively clip neurons based on their relative importance to the targeted forget data, curated for different modalities. Specifically, \method consists of two stages: important neuron selection and selective pruning. The first stage identifies and collects the most influential neurons across modalities relative to the targeted forget knowledge, while the second stage is dedicated to pruning those selected neurons. \method effectively isolates and removes the neurons that contribute most to the forget data within each modality, while preserving the integrity of retained knowledge. Our experiments conducted across various MLLM architectures illustrate that \method can achieve a more balanced and comprehensive unlearning in each modality without largely affecting the overall model utility. \footnote{Code is available at \href{https://github.com/franciscoliu/MANU}{franciscoliu/MANU}.}
\end{abstract}

\section{Introduction}
The rapid advancement of Large Language Models (LLMs) \cite{brown2020language, chowdhery2023palm, touvron2023llama, fu2024amoeballm, qin2023chatgpt} and Multimodal Large Language Models (MLLMs) \cite{liu2024improved, ye2023mplug, ye2024mplug, zhu2023minigpt, zhang2025overcoming, zhang2025pretrained} have showcased their exceptional capabilities across various AI domains \cite{ouyang2022training, tan2024personalized, ni2025towards, zhang2024mopi},
largely due to extensive pre-training and fine-tuning on vast data corpus. 
However, this remarkable learning ability also poses risks such as privacy violations and copyright infringements. Since retraining from scratch while excluding these data is computationally expensive, \textit{Machine Unlearning (MU)} \cite{nguyen2022survey, wang2024can, liu2024rethinking, liu2024machine} has emerged as an efficient alternative to remove the influence of sensitive data while preserving overall model performance.

\begin{figure}
  \includegraphics[width=1.0\columnwidth]{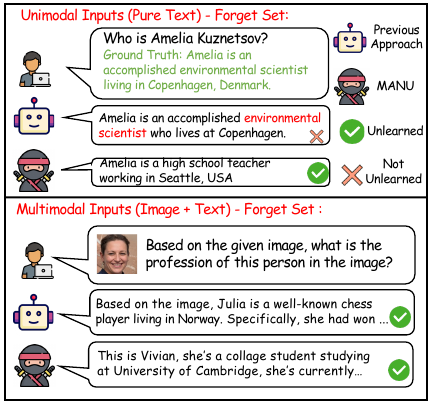}
  \vspace{-0.30in}
  \caption{Comparison of \method with the previous approach in responding to questions related to unlearned targets, using multimodal inputs (i.e., images with associated text) and pure text inputs, respectively.}
  \label{fig:intro_demo}
  \vspace{-0.30in}
\end{figure}

Recent research has advanced MU techniques for LLMs \cite{zhang2024negative, liu2024towards, yao2023large, pochinkov2024dissecting, dou2024avoiding} while neglecting the case of MLLMs. Although extending MU methods from LLMs to MLLMs may seem intuitive, \citet{liu2024protecting} highlights that such adaptations often result in \textbf{imbalanced unlearning}, where knowledge is removed in multimodal (image-text) level but remains in unimodal (text-only) level (e.g. Figure \ref{fig:intro_demo}). This discrepancy arises from fundamental differences between LLMs and MLLMs, particularly in knowledge representation and integration. While LLMs store target knowledge within a single modality, MLLMs integrate cross-modal interactions that entangle knowledge, making selective unlearning more challenging and potentially leading to drastic unintended knowledge loss. We provide a detailed explanation is provided in Section \ref{sec:motivation}.

To address this challenge, we propose \method, a novel two-stage unlearning approach that strategically prunes neurons associated with target knowledge entangled across both vision and textual modalities.
Specifically, the first stage focuses on identifying critical neurons that contribute significantly to the forget dataset. This is achieved using four importance functions: absolute importance, frequency importance, variance importance, and root mean square importance functions. In the second stage, a scoring function is defined to evaluate neurons based on the importance scores calculated in the previous stage, facilitating the pruning of these neurons from the original model. Our main contributions are as follows:
\begin{enumerate} 
    \setlength{\itemsep}{-2pt}
    \item We investigate the unique challenge of MLLM unlearning and highlight the limitations of previous methods designed for unimodal LLMs, which lack modality-specific design. Consequently, even when applied to multimodal inputs, these methods lead to imbalanced unlearning, effectively removing target knowledge in multimodal inputs while retaining it in unimodal level.
    \item We propose \method, the \textbf{first} modality-aware unlearning framework for MLLMs, which disentangles and removes modality-specific knowledge while preserving model utility across multiple perspectives.
    \item Experiments and case studies demonstrate the effectiveness of \method in unlearning sensitive knowledge across modalities while preserving model utility in various MLLMs.
\end{enumerate}

\section{Motivation}
\label{sec:motivation}
Inspired by \cite{liu2024protecting}, which highlights the challenge of imbalanced unlearning in MLLMs, where unimodal LLM methods fail to remove knowledge across modalities.
In particular, unlearning in one modality does not necessarily eliminate the corresponding knowledge in another, leading to knowledge retention.
We hypothesize that this occurs due to entangled knowledge representations across modalities, making it insufficient to unlearn from one modality alone. Specifically, the activated neuron varies by input type, meaning that unlearned knowledge may persist even after targeted unlearning.

\begin{figure}
\centering
\begin{subfigure}{0.49\columnwidth}
    \includegraphics[width=\columnwidth]{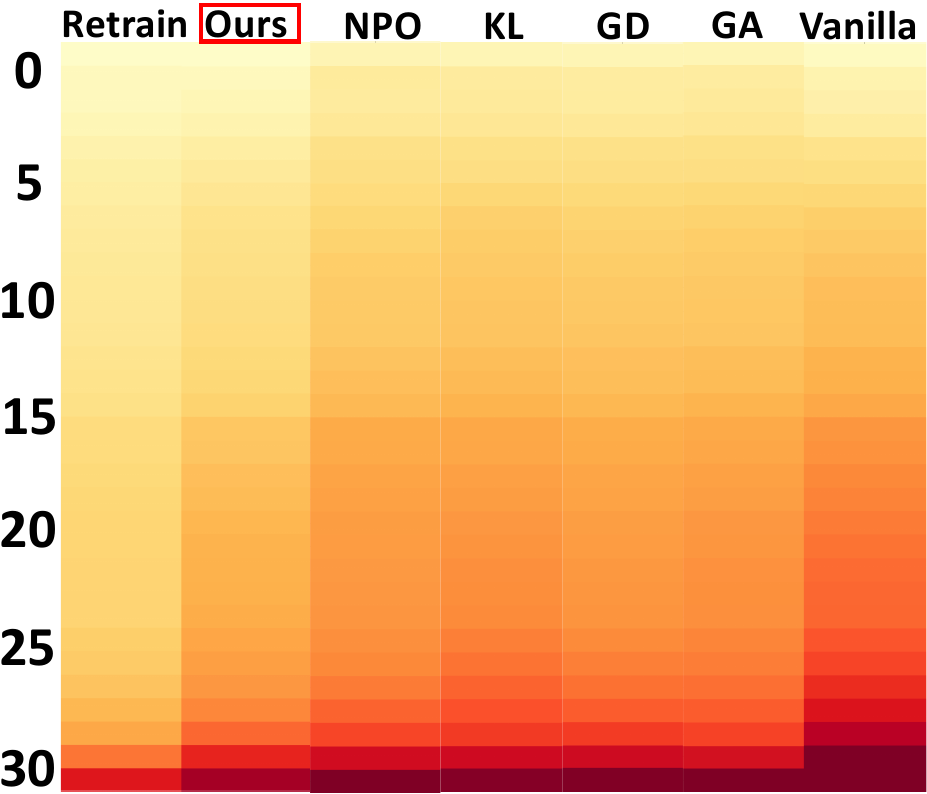}
    \subcaption{Text-Only Inputs (Forget)}
    \label{fig:text-only-residual-forget}
\end{subfigure}    
\begin{subfigure}{0.49\columnwidth}
    \includegraphics[width=\columnwidth]{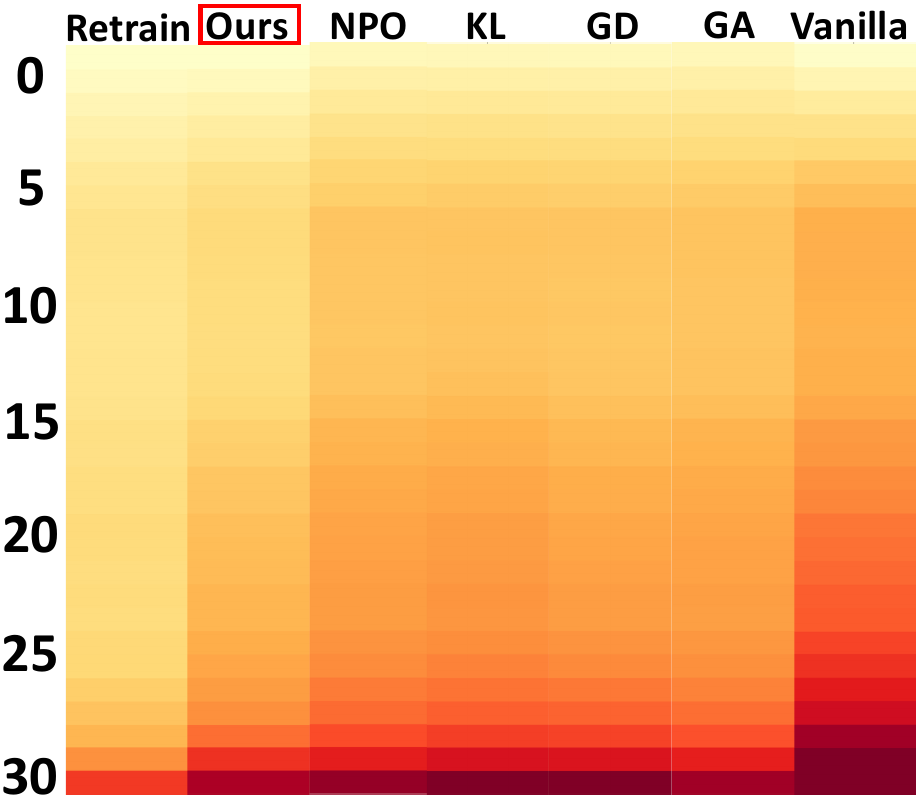}
    \subcaption{Multi. Inputs (Forget)}
    \label{fig:multimodal-residual-forget}
\end{subfigure}
\begin{subfigure}{0.49\columnwidth}
    \includegraphics[width=\columnwidth]{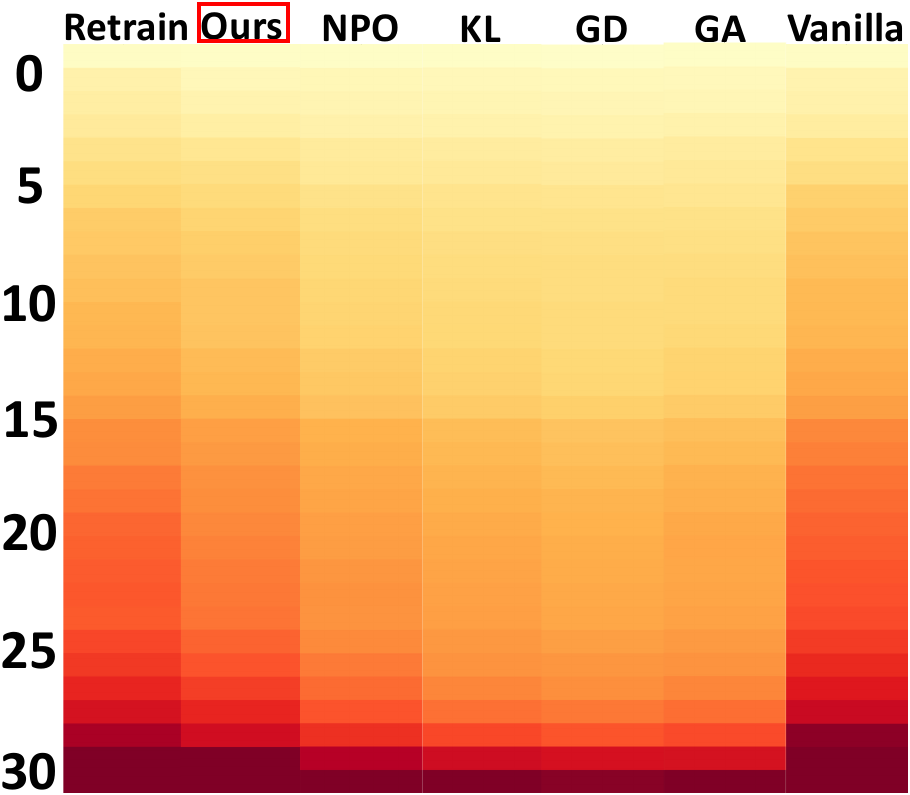}
    \subcaption{Text-Only Inputs (Retain)}
    \label{fig:text-only-residual-retain}
\end{subfigure}    
\begin{subfigure}{0.49\columnwidth}
    \includegraphics[width=\columnwidth]{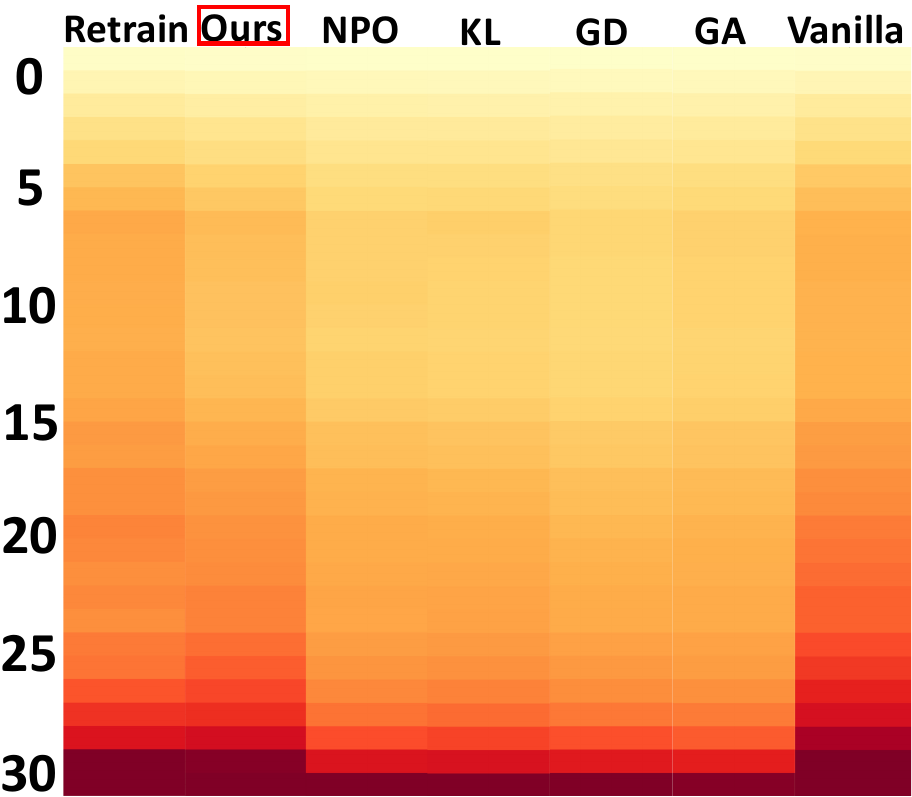}
    \subcaption{Multi. Inputs (Retain)}
    \label{fig:multimodal-residual-retain}
\end{subfigure}
\caption{Visualization of knowledge retention across MLLM language module layers for different unlearning methods on the forget/retain sets of MLLMU-Bench. Figures \ref{fig:text-only-residual-forget}, \ref{fig:text-only-residual-retain} show text-only residuals, while Figures \ref{fig:multimodal-residual-forget}, \ref{fig:multimodal-residual-retain} depict multimodal residuals. The $x$-axis represents unlearning methods (Grad. Diff. as GD), the $y$-axis shows layer indices, and darker red indicates higher knowledge retention.}
\label{fig:heatmap_motivation}
\vspace{-0.20in}
\end{figure}

To validate this hypothesis, we compare our modality-aware approach with prior methods that unlearn only multimodal knowledge, using exclusively multimodal inputs. The heatmap comparisons are shown in Figure \ref{fig:heatmap_motivation}. Additionally, we include the vanilla and retrained models from MLLMU-Bench.
We first examine the heatmap of different unlearning algorithms on the \textbf{forget set}, which contains data designated for removal. 
As shown in Figure \ref{fig:text-only-residual-forget} and \ref{fig:multimodal-residual-forget}, fainter colors indicate lower knowledge retention, while deeper colors signify higher retention.
Next, when comparing \ref{fig:text-only-residual-forget} and \ref{fig:multimodal-residual-forget}, we observe that while prior methods effectively unlearn target knowledge from multimodal inputs (\ref{fig:multimodal-residual-forget}), they fail to fully remove this knowledge in the unimodal setting (\ref{fig:text-only-residual-forget}), where only textual inputs are provided. This finding suggests that inputs with different modalities activate distinct neurons, underscoring the challenges of achieving comprehensive unlearning across modalities. Detailed analysis of modality-specific performance is provided in Section \ref{sec: unlearn-across-modality}.

Furthermore, we present the heatmap of these algorithms on the \textbf{retain set} with different input types, as shown in Figure \ref{fig:text-only-residual-retain} and \ref{fig:multimodal-residual-retain}. Unlike the forget set, where knowledge should be erased, the objective here is to preserve unrelated knowledge, meaning that deeper colors indicate stronger retention ability. As expected, the vanilla and retrained models exhibit the darkest colors across layers, indicating strong knowledge retention on retain set. 
However, unlearning algorithms such as GA and Gradient Difference display noticeably lighter colors, signifying unintended knowledge loss on the retain set. 
Those heatmaps further reinforce the findings of \citet{liu2024protecting}, demonstrating that effective MLLM unlearning must disentangle multimodal representations to prevent unintended loss while preserving retained knowledge.

\section{Method}
\begin{figure}
  \includegraphics[width=1.0\columnwidth]{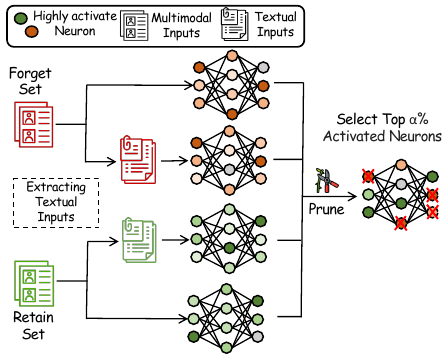}
  \vspace{-0.25in}
  \caption{The overall framework of \method. The forget and retain sets are first split into text-only and multimodal modalities. Neuron activations are then computed across modalities and datasets, followed by applying an importance and scoring function to evaluate activated neurons. Finally, the top $\alpha \%$ of neurons are pruned based on their scores.}
  \label{fig:method}
  \vspace{-0.25in}
\end{figure}

\vspace{-0.05in}
In this section, we elaborate on \method (Figure \ref{fig:method}), a two-stage modality-aware pruning framework designed to selectively remove sensitive information forget set $\mathcal{D}_f$ while preserving model utility on retain set $\mathcal{D}_r$ from MLLMs and various general benchmarks. The first stage involves identifying and selecting the most contributed neurons across two modalities on the forget set.
\subsection{Important Neuron Selection Stage}
The first stage applies four importance functions to assess the relative importance of neurons in the language and vision MLP layers for both the forget set $\mathcal{D}_f$ and retain set $\mathcal{D}_r$. 
First, we leverage the observation that meaningful neuron activity is characterized by deviations from zero, as most activations remain close to zero by default (see Appendix \ref{appendix:neuron_activation_distribution}). Given a neuron $n$ and its corresponding activations $z$, we define absolute importance ($I_{\text{abs}}$) to measure the difference in activation magnitudes between modalities relative to an arbitrary dataset $\mathcal{D}$, capturing the modality-specific processing preferences of individual neurons:
\begin{align*}
    I_{\text{abs}}(\mathcal{D},n) := \frac{|\bar{Z}_{\text{multi}} - \bar{Z}_{\text{text}}|}{\bar{Z}_{\text{multi}} + \bar{Z}_{\text{text}} + \epsilon}
\end{align*}
where modality-specific mean absolute activations can be formulated as:
\begin{align*}
    \bar{Z}_{\text{multi}} &= \frac{1}{|\mathcal{D}_{\text{multi}}|} \sum_{d \in \mathcal{D}_{\text{multi}}} |z_{\text{multi}}(d)|, \\
    \bar{Z}_{\text{text}} &= \frac{1}{|\mathcal{D}_{\text{text}}|} \sum_{d \in \mathcal{D}_{\text{text}}} |z_{\text{text}}(d)|,
\end{align*}
where $\mathcal{D}_{\text{text}}, \mathcal{D}_{\text{multi}}\subset \mathcal{D}$,
represent the dataset in pure textual format and image with associated text format, respectively. Here, $z_{\text{multi}}(d)$ and $z_{\text{text}}(d)$ denote the absolute activation values of neuron $n$ when processing a sample $d$ from the multimodal and textual subsets, respectively. The normalization ensures that neurons with strong activation disparities between modalities are highlighted while controlling for overall activation magnitude.  The small constant 
$\epsilon$ in the denominator is added for numerical stability, preventing division by zero.

Second, motivated by findings that neuron activation distributions exhibit a sharp peak at zero—indicating that most neurons remain inactive by default, with only a subset selectively activating in response to specific inputs \cite{zhang2021moefication} (see Appendix \ref{appendix:neuron_activation_distribution} for elaborations)—we introduce frequency importance (\( I_{\text{freq}} \)) to quantify how often a neuron's activation significantly deviates from zero. Since modality-relevant neurons are expected to fire more frequently when processing inputs from their associated modality, \( I_{\text{freq}} \) helps distinguish consistently engaged neurons from those that activate only sporadically. We first define the modality-specific activation frequency as:
\begin{align*}
    N_{\text{multi}} &= \big| \{ d \in \mathcal{D}_{\text{multi}} \mid |z_{\text{multi}}(d)| > \tau \} \big|, \\
    N_{\text{text}} &= \big| \{ d \in \mathcal{D}_{\text{text}} \mid |z_{\text{text}}(d)| > \tau \} \big|.
\end{align*}
Using these definitions, we compute frequency importance as:
\vspace{-0.05in}
\begin{align*}
I_{\text{freq}}(\mathcal{D},n) :&= \frac{|\Delta N|}{\Sigma N + \epsilon}, \\
    \Delta N &= N_{\text{multi}} - N_{\text{text}}, \\
    \Sigma N &= N_{\text{multi}} + N_{\text{text}}.
\end{align*}
This normalized frequency metric complements absolute importance $I_{\text{abs}}$ by focusing on activation consistency rather than magnitude, enabling the identification of neurons that may exhibit moderate but reliable modality-specific responses.

Third, building on information theory principles \cite{varley2023information}, which suggest that neurons carrying more information should exhibit diverse activation patterns rather than consistently remaining near zero, we define variance importance (\( I_{\text{var}} \)) to measure the spread of activation values within each modality, thereby quantifying each neuron's contribution to modality-specific information processing. Using the previously defined \( \bar{Z}_{\text{multi}} \) and \( \bar{Z}_{\text{text}} \), we compute the variance within each modality as:
\vspace{-10pt}
\begin{align*}
    \text{Var}_{\text{multi}} &= \frac{1}{|\mathcal{D}_{\text{multi}}|} \sum_{d \in \mathcal{D}_{\text{multi}}} (z_{\text{multi}}(d) - \bar{Z}_{\text{multi}})^2, \\
    \text{Var}_{\text{text}} &= \frac{1}{|\mathcal{D}_{\text{text}}|} \sum_{d \in \mathcal{D}_{\text{text}}} (z_{\text{text}}(d) - \bar{Z}_{\text{text}})^2, \\
    I_{\text{var}}(\mathcal{D}, n) :&= \sqrt{\text{Var}_{\text{multi}} + \text{Var}_{\text{text}}}.
\end{align*}
\( I_{\text{var}} \) provides a statistically robust measure of how differently a neuron responds across modalities. Larger values indicate neurons that maintain distinct roles in processing multimodal versus unimodal inputs. 

Finally, as highlighted by \citet{liu2023deja}, many neuron activations may be redundant, meaning they are consistently active across different inputs but do not contribute meaningfully to specific outputs. This suggests that a subset of neurons fire indiscriminately rather than being specialized for particular tasks or modalities, leading to inefficiencies in representation. To address this, we introduce root mean square importance (\( I_{\text{rms}} \)) to identify neurons with consistently strong activations relative to the overall activation pattern, formulated as: 
\begin{align*}
I_{\text{rms}}(\mathcal{D},n) &:= \sqrt{\frac{|\Delta Z^2|}{\Sigma Z^2 + \epsilon}}, \\
\text{where } Z^2_{\text{multi}} &= \sum_{d \in \mathcal{D}_{\text{multi}}} z_{\text{multi}}(d)^2, \\
Z^2_{\text{text}} &= \sum_{d \in \mathcal{D}_{\text{text}}} z_{\text{text}}(d)^2, \\
\Delta Z^2 &= Z^2_{\text{multi}} - Z^2_{\text{text}}, \\
\Sigma Z^2 &= Z^2_{\text{multi}} + Z^2_{\text{text}}.
\end{align*}
$I_{\text{rms}}$ emphasizes neurons with substantial modality-specific activity while penalizing those with redundant activation patterns, ensuring the identification of truly specialized neural pathways for each modality.
Together, we aggregate these four importance functions into a unified importance measure through a weighted combination. Specifically, for any dataset $\mathcal{D}$ and neuron $n$, we compute:
\begin{equation*}
\mathcal{I}(\mathcal{D}, n) := \sum_{k \in \mathcal{K}} I_k(\mathcal{D}, n)
\end{equation*}
where $\mathcal{K} = \{I_\text{abs}, I_\text{freq}, I_\text{var}, I_\text{rms}\}$ represents our set of importance functions. This combined measure $\mathcal{I}$ denotes the comprehensive assessment of neuron importance by capturing different aspects of neural activation patterns: magnitude ($I_{\text{abs}}$), activation frequency ($I_{\text{freq}}$), activation diversity ($I_{\text{var}}$), and consistent strength ($I_{\text{rms}}$).

\subsection{Selective Pruning Stage}

In the second stage, we define a scoring function $S_n$ that aims to determine the pruned neurons based on the calculated importance from previous stage. In particular, given forget set $\mathcal{D}_f$ and retain set $\mathcal{D}_r$, we have:
\begin{align*}
    S_n = \frac{\mathcal{I}(\mathcal{D}_{f}, n)}{\mathcal{I}(\mathcal{D}_{r}, n) + \epsilon}.
\end{align*}
Now given a vanilla model $\theta$ and a pruning rate $\alpha$, we perform selective pruning by choosing and removing neurons based on their importance scores relative to forget set $\mathcal{D}_f$. Specifically, we can identify the set of neurons to prune by using the scoring function $S_n$:
\begin{equation*}
\mathcal{N} = \{n : S_n \text{ is among the top } \alpha\% \text{ of all scores}\}.
\end{equation*}
\noindent For each selected neuron $n \in \mathcal{N}$, we perform the pruning operation by setting its weights to zero and obtain pruned model $\theta^{\prime}$:
\begin{equation*}
\theta^{\prime} =
\begin{cases}
0 & \text{if } n \in \mathcal{N}, \\
\theta & \text{otherwise}.
\end{cases}
\end{equation*}

\section{Experiments}
\label{sec:experiment}
In this section, we present extensive experiments to validate the effectiveness of \method. Specifically, these experiments aim to address the following research questions: (1) Can \method effectively unlearn the target knowledge from the model? (2) Does \method successfully address the unique challenge of imbalanced unlearning across different modalities in MLLMs? (3) How do different pruning ratios affect the effectiveness of \method during the unlearning process? (4) Can \method achieve a good balance between unlearning the target knowledge and preserving the model's utility?

\subsection{Experimental Setup}

Our experiments focus on unlearning fictitious profiles at both visual and textual levels using MLLMU-Bench \cite{liu2024protecting}, a benchmark for evaluating unlearning in MLLMs. We conduct experiments on LLaVA-1.5-7B \cite{liu2024improved} and Idefics2-8B \cite{laurenccon2024matters},  evaluating performance across four datasets to assess unlearning effectiveness, generalizability, and model utility. The Forget Set contains a subset of fictitious profiles designated for unlearning effectiveness, with 5\%, 10\%, and 15\% selected for removal. A corresponding Test Set mirrors this split but includes images transformed to different angles and paraphrased text to assess generalizability. Lastly, for model utility evaluation, we assess performance using the Retain Set, Real Celebrity Set, and general benchmarks. The Retain Set includes fictitious profiles excluded from the Forget and Test Sets that the model should retain, while the Real Celebrity Set contains real-world celebrity profiles distinct from fictitious ones. Additionally, to assess model utility more comprehensively, we evaluate general reasoning and helpfulness post-unlearning using MMMU \cite{yue2024mmmu} and LLaVA-Bench \cite{liu2024visual}, examining whether unlearning impacts core model capabilities.

For each evaluation set, every approach is assessed across three tasks. The classification task presents a multiple-choice format to measure the model’s ability to differentiate correct from incorrect associations. The generation task evaluates factual accuracy and coherence using ROUGE-L \cite{lin2004rouge} and LLM-determined factuality scores. The cloze test task measures the model’s ability to complete missing information, evaluated via exact-match accuracy. Details on evaluation metrics and dataset construction are provided in Appendix \ref{appendix:mllmu-bench}.

\subsection{Baseline methods}
For baselines, we compared Gradient Ascent (GA) \cite{thudi2022unrolling}, Gradient Difference \cite{liu2022continual}, KL Minimization \cite{nguyen2020variational}, Negative Preference Optimization (NPO) \cite{zhang2024negative} and a generic prevention strategies using system prompts (prompting) to prevent models from producing privacy-related information. Specifically, the GA approach applies opposite gradient updates on $\mathcal{D}_f$. The Gradient Difference approach extends GA by adding a gradient updates on $\mathcal{D}_f$ and $\mathcal{D}_r$, ensuring unlearning without performance degradation. Next, the KL Minimization approach aligns the unlearned model's predictions on $\mathcal{D}_r$ with the vanilla model while encouraging divergence from the knowledge of $\mathcal{D}_f$. Lastly, the NPO treats $\mathcal{D}_f$ as dispreferred data and casts unlearning into a preference optimization framework, utilizing an oracle model fine-tuned exclusively on the $\mathcal{D}_r$. Lastly, we employ a generic prevention technique by utilizing a crafted system prompt (i.e. prompting). Further details on the baselines can be found in Appendix \ref{appendix:baselines}.

\subsection{Implementation Details}
All experiments on both LLaVA and Idefics2 models are implemented on a server with 3 NVIDIA A6000 GPUs and Intel(R) Xeon(R) Silver 4210R CPU @ 2.40GHz with 20 CPU cores. Details can be referred to Appendix \ref{appendix:hyperparameter_settings}.

\subsection{Main Results}
To answer the first research question: \textbf{Can \method effectively unlearn the target knowledge from the model}, we conduct extensive experiments on MLLMU-Bench using different data splits across various MLLMs. The results of these experiments are presented in Table \ref{tab:main-table-llava} and Table \ref{tab:main-table-idefics}. For each task in each dataset, we report the \textbf{average performance} for both multimodal and unimodal evaluation across three distinct tasks. From the table, it is evident that \method demonstrates exceptional performance across all datasets and tasks on both the LLaVA and Idefics2 models with different data splits, consistently ranking as either the best or second-best method among all baselines. Notably, while GA-based approaches occasionally surpass \method in unlearning performance (e.g., LLaVA model with a 15\% forget split), it is crucial to emphasize the importance of preserving model utility on the retain set and real celebrity set while selectively unlearning knowledge from the Forget Set. From this perspective, the superior unlearning performance of GA-based methods often comes at a significant cost to model utility, making them the least effective approaches on maintaining model utility. 
Lastly, NPO appears as another competitive baseline due to its relatively stable performance in both unlearning effectiveness and model utility. However, it is not as effective as \method in achieving these two objectives.

\begin{table*}[t!]
    \centering
\scalebox{0.51}{
\begin{tabular}{l|cccc|cccc|cccc|cccc}
        \toprule
        \multirow{3}{*}{\textbf{Models}} 
        & \multicolumn{4}{c|}{\textbf{Forget Set}} 
        & \multicolumn{4}{c|}{\textbf{Test Set}} 
        & \multicolumn{4}{c|}{\textbf{Retain Set}} 
        & \multicolumn{4}{c}{\textbf{Real Celebrity}} \\
        \cline{2-17}
        & \begin{tabular}[c]{@{}c@{}}Class.\\ Acc (\textcolor{blue}{$\downarrow$})\end{tabular} 
        & \begin{tabular}[c]{@{}c@{}}Rouge\\ Score (\textcolor{red}{$\downarrow$})\end{tabular} 
        & \begin{tabular}[c]{@{}c@{}}Fact.\\ Score (\textcolor{red}{$\downarrow$})\end{tabular} 
        & \begin{tabular}[c]{@{}c@{}}Cloze\\ Acc (\textcolor{teal}{$\downarrow$})\end{tabular} 
        & \begin{tabular}[c]{@{}c@{}}Class.\\ Acc (\textcolor{blue}{$\downarrow$})\end{tabular} 
        & \begin{tabular}[c]{@{}c@{}}Rouge\\ Score (\textcolor{red}{$\downarrow$})\end{tabular} 
        & \begin{tabular}[c]{@{}c@{}}Fact.\\ Score (\textcolor{red}{$\downarrow$})\end{tabular} 
        & \begin{tabular}[c]{@{}c@{}}Cloze\\ Acc (\textcolor{teal}{$\downarrow$})\end{tabular} 
        & \begin{tabular}[c]{@{}c@{}}Class.\\ Acc (\textcolor{blue}{$\uparrow$})\end{tabular} 
        & \begin{tabular}[c]{@{}c@{}}Rouge\\ Score (\textcolor{red}{$\uparrow$})\end{tabular} 
        & \begin{tabular}[c]{@{}c@{}}Fact.\\ Score (\textcolor{red}{$\uparrow$})\end{tabular} 
        & \begin{tabular}[c]{@{}c@{}}Cloze\\ Acc (\textcolor{teal}{$\uparrow$})\end{tabular} 
        & \begin{tabular}[c]{@{}c@{}}Class.\\ Acc (\textcolor{blue}{$\uparrow$})\end{tabular} 
        & \begin{tabular}[c]{@{}c@{}}Rouge\\ Score (\textcolor{red}{$\uparrow$})\end{tabular} 
        & \begin{tabular}[c]{@{}c@{}}Fact.\\ Score (\textcolor{red}{$\uparrow$})\end{tabular} 
        & \begin{tabular}[c]{@{}c@{}}Cloze\\ Acc (\textcolor{teal}{$\uparrow$})\end{tabular} \\
        \midrule
        \multicolumn{17}{c}{\textbf{LLaVA-1.5-7B (5\% Forget)}} \\
        \midrule

        Vanilla & 51.70\% & 0.645 & 6.78 & 25.81\% & 47.86\% & 0.539 & 4.89 & 23.01\% & 46.11\% & 0.632 & 6.41 & 27.83\% & 51.80\% & 0.479 & 5.47 & 17.35\% \\
        
        GA & 44.40\% & \textbf{0.485} & 3.38 & 17.19\% & \textbf{38.40\%} & 0.384 & \textbf{3.47} & 16.47\% & 39.09\% & 0.495 & 2.97 & 18.96\% & 45.56\% & 0.414 & 3.42 & 8.66\% \\
        
        Grad. Diff. & \underline{43.60\%} & 0.507 & \textbf{3.05} & \textbf{16.00\%} & 43.41\% & \textbf{0.323} & 3.83 & \underline{16.19\%} & 41.07\% & 0.508 & 4.14 & 16.90\% & 46.52\% & 0.364 & 3.26 & 9.31\% \\
        
        KL Minimization & 46.80\% & 0.574 & 5.04 & 20.46\% & 45.20\% & 0.396 & 4.54 & 20.04\% & 38.83\% & 0.478 & 4.20 & 21.03\% & 45.64\% & 0.418 & 3.49 & 14.53\%\\
        
        Prompting & 46.80\% & 0.558 & 4.51 & 23.81\% & 44.87\% & 0.415 & 4.18 & 21.99\% & \underline{42.99\%} & \textbf{0.612} & \textbf{5.42} & \textbf{26.75\%} & \textbf{51.60\%} & 0.443 & \textbf{5.43} & \textbf{17.18\%} \\
        
        NPO & 45.61\% & 0.525 & 3.41 & 22.76\% & 44.44\% & 0.347 & 3.91 & 20.00\% & 42.61\% & 0.515 & 4.38 & 21.37\% & 49.51\% & \textbf{0.450} & 4.63 & 15.16\% \\

        \rowcolor{gray!12}\method & \textbf{41.25\%} & \underline{0.491} & \underline{3.27} & \underline{17.08\%} & \underline{41.67\%} & \underline{0.334} & \underline{3.81} & \textbf{15.78\%} & \textbf{43.38\%} & \underline{0.542} & \underline{4.45} & \underline{24.08\%} & \underline{49.57\%} & \underline{0.448} & \underline{4.67} & \underline{16.01\%} \\

        \midrule
        
        \multicolumn{17}{c}{\textbf{LLaVA-1.5-7B (10\% Forget)}} \\
        \midrule
       
        Vanilla & 49.15\% & 0.594 & 6.40 & 26.97\% & 47.41\% & 0.510 & 5.20 & 25.43\% & 46.68\% & 0.582 & 5.44 & 28.49\% & 51.80\% & 0.479 & 5.47 & 17.35\% \\
        
        GA & 43.85\% & 0.510 & 3.51 & 20.91\% & 40.60\% & 0.421 & 3.19 & 15.77\% & 41.91\% & 0.471 & 3.36 & 19.52\% & 42.64\% & 0.320 & 3.43 & 10.53\% \\
        
        Grad. Difference & \underline{41.60\%} & \underline{0.508} & \textbf{3.16} & \textbf{18.79\%} & \underline{39.08\%} & \underline{0.414} & \underline{3.07} & \textbf{14.50\%} & 43.71\% & 0.474 & 3.28 & 17.55\% & 40.94\% & 0.391 & 3.44 & 10.51\% \\
        
        KL Minimization & 44.80\% & 0.579 & 4.12 & 22.69\% & 42.75\% & 0.420 & 3.29 & 20.50\% & 39.93\% & 0.456 & 3.82 & 20.70\% & 45.58\% & 0.462 & 3.13 & 14.90\% \\
        
        Prompting & 48.41\% & 0.561 & 4.75 & 26.55\% & 47.29\% & 0.479 & 4.21 & 24.11\% & \textbf{45.97\%} & \textbf{0.577} & \textbf{5.43} & \textbf{26.12\%} & \textbf{51.60\%} & \textbf{0.471} & \underline{4.43} & \textbf{17.16\%} \\
        
        NPO & 47.40\% & 0.515 & 5.05 & 22.10\% & 46.42\% & 0.428 & 4.25 & 21.66\% & 44.81\% & 0.488 & \underline{5.35} & \underline{22.29\%} & 47.89\% & 0.451 & \textbf{4.53} & \underline{16.33\%} \\

        \rowcolor{gray!12}\method & \textbf{38.36\%} & \textbf{0.503} & \underline{3.48} & \underline{19.44\%} & \textbf{37.24\%} & \textbf{0.402} & \textbf{3.06} & \underline{15.61\%} & \underline{44.89\%} & \underline{0.538} & \underline{5.35} & \underline{24.61\%} & \underline{48.02\%} & \underline{0.466} & 4.32 & 15.68\% \\

        \midrule
        \multicolumn{17}{c}{\textbf{LLaVA-1.5-7B (15\% Forget)}} \\
        \midrule

        Vanilla & 51.87\% & 0.575 & 6.34 & 26.62\% & 47.53\% & 0.502 & 4.08 &  25.33\% & 48.06\% & 0.585 & 5.46 & 28.51\% & 51.80\% & 0.479 & 5.47 & 17.35\% \\
        
        GA & \textbf{40.93\%} & \underline{0.482} & \textbf{3.51} & \textbf{17.33\%} & \textbf{39.64\%} & 0.371 & \underline{3.57} & 17.67\% & 40.43\% & 0.460 & 3.66 & 19.14\% & 40.36\% & 0.378 & 3.54 & 10.13\% \\
        
        Grad. Diff. & 43.47\% & 0.518 & 3.98 & 18.78\% & 42.18\% & \underline{0.401} & 3.61 & \underline{18.11\%} & 41.82\% & 0.476 & 3.28 & 21.30\% & 41.21\% & 0.417 & 3.45 & 11.37\% \\
        
        KL Minimization & 47.60\% & 0.541 & 4.57 & 23.44\% & 43.20\% & 0.439 & 3.78 & 21.09\% & 42.96\% & 0.442 & 4.42 & 22.28\% & 42.58\% & 0.415 & 3.21 & 14.41\% \\
        
        Prompting & 49.73\% & 0.547 & 4.63 & 26.00\% & 46.81\% & 0.483 & 3.67 & 24.56\% & \textbf{47.09\%} & \textbf{0.585} & \textbf{5.46} & \textbf{26.36\%} & \textbf{51.60\%} & \textbf{0.458} & \textbf{4.91} & \textbf{16.84\%} \\
        
        NPO & 45.52\% & 0.509 & 4.39 & 20.63\% & 43.43\% & 0.439 & 4.01 & 21.88\% & 46.84\% & 0.525 & 4.98 & 23.31\% & 48.09\% & 0.433 & \underline{4.11} & 14.10\% \\

        \rowcolor{gray!12}\method & \underline{42.05\%} & \textbf{0.481} & \underline{3.73} & \underline{17.91\%} & \underline{41.75\%} & \textbf{0.360} & \textbf{3.52} & \textbf{17.01\%} & \underline{46.86\%} & \underline{0.557} & \underline{5.19} & \underline{24.62\%} & \underline{50.42\%} & \underline{0.448} & 4.05 & \underline{16.77\%} \\
                
        \bottomrule
    \end{tabular}}
    \vspace{-0.1in}
    \caption{Overall average results of baseline methods and \method on LLaVA, combining multimodal and unimodal evaluations across three forget setups. \textbf{Bold} denotes the best performance, \underline{underline} the runner-up. Each method is evaluated on four MLLMU-Bench datasets using classification accuracy, ROUGE-L, factuality, and cloze accuracy. Factuality Score is abbreviated as Fact. Score. \textcolor{blue}{$\sbullet[.75]$}, \textcolor{red}{$\sbullet[.75]$}, and \textcolor{teal}{$\sbullet[.75]$} represent classification, generation, and cloze evaluations, respectively. $\downarrow$ indicates lower is better, $\uparrow$ indicates higher is better.}
    \label{tab:main-table-llava}
    \vspace{-0.15in}
\end{table*}

\section{Discussion}
\label{sec: discussion}
Though \method achieves superior average performance compared to other baselines, it remains unclear whether \method effectively overcomes the unique challenge inherent in MLLM unlearning. In this section, we aim to address this concern by answering three key questions essential to advancing the understanding of MLLM unlearning.

\subsection{Unlearning across modalities}
\label{sec: unlearn-across-modality}

\begin{figure*}[!t]
\centering
\begin{subfigure}[b]{\textwidth}
    \centering
    \includegraphics[width=0.7\textwidth]{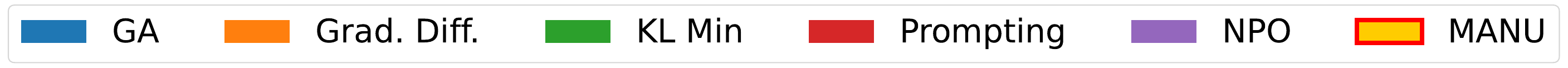}
\end{subfigure}
\begin{subfigure}{0.244\textwidth}
    \includegraphics[width=\textwidth]{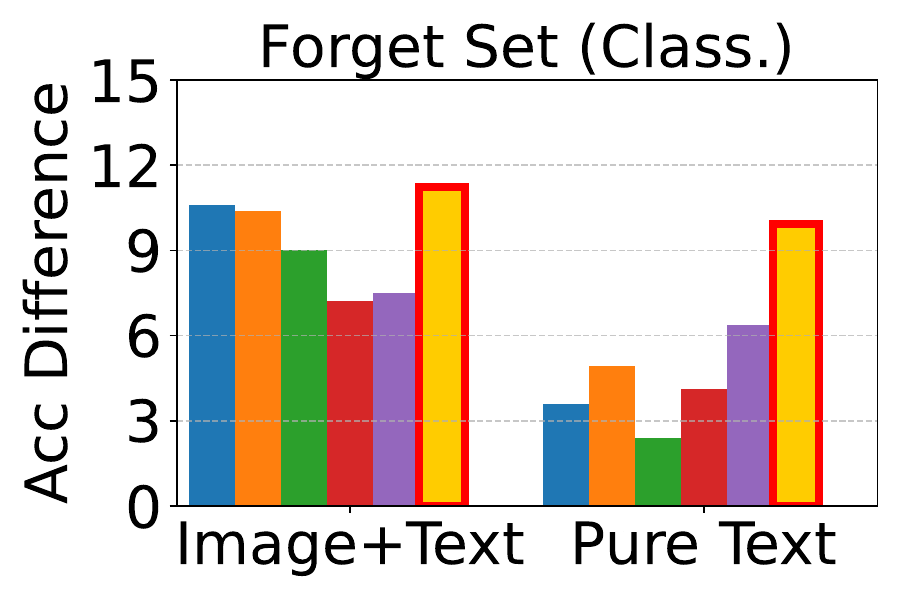}
    \subcaption{Forget Set (Classification)}
    \label{fig:llava_5_class_forget}
\end{subfigure}    
\begin{subfigure}{0.244\textwidth}
    \includegraphics[width=\textwidth]{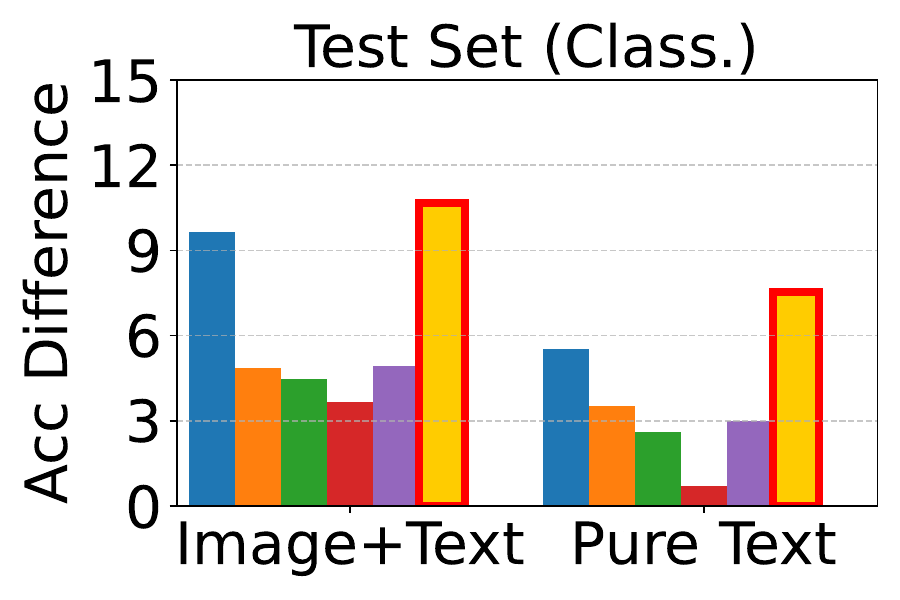}
    \subcaption{Test Set (Classification)}
    \label{fig:llava_5_class_test}
\end{subfigure}
\begin{subfigure}{0.244\textwidth}
    \includegraphics[width=\textwidth]{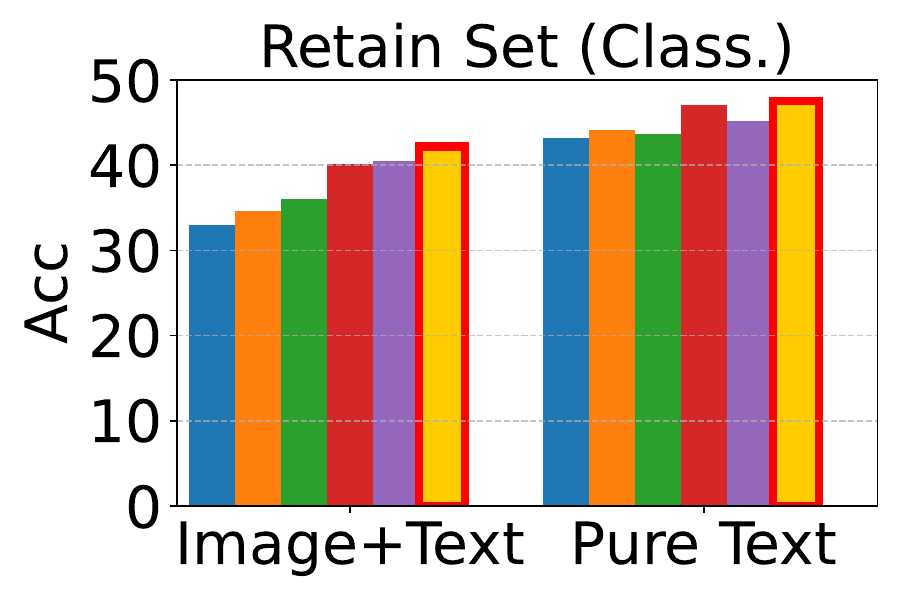}
    \subcaption{Retain Set (Classification)}
    \label{fig:llava_5_class_retain}
\end{subfigure}    
\begin{subfigure}{0.244\textwidth}
    \includegraphics[width=\textwidth]{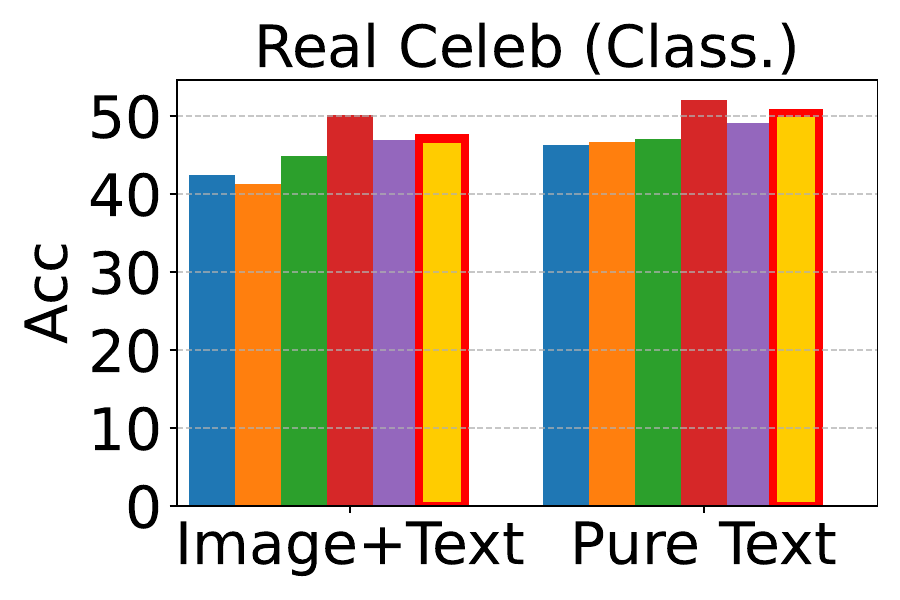} 
    \subcaption{Real Celeb (Classification)}
    \label{fig:llava_5_class_real}
\end{subfigure}


\begin{subfigure}{0.244\textwidth}
    \includegraphics[width=\textwidth]{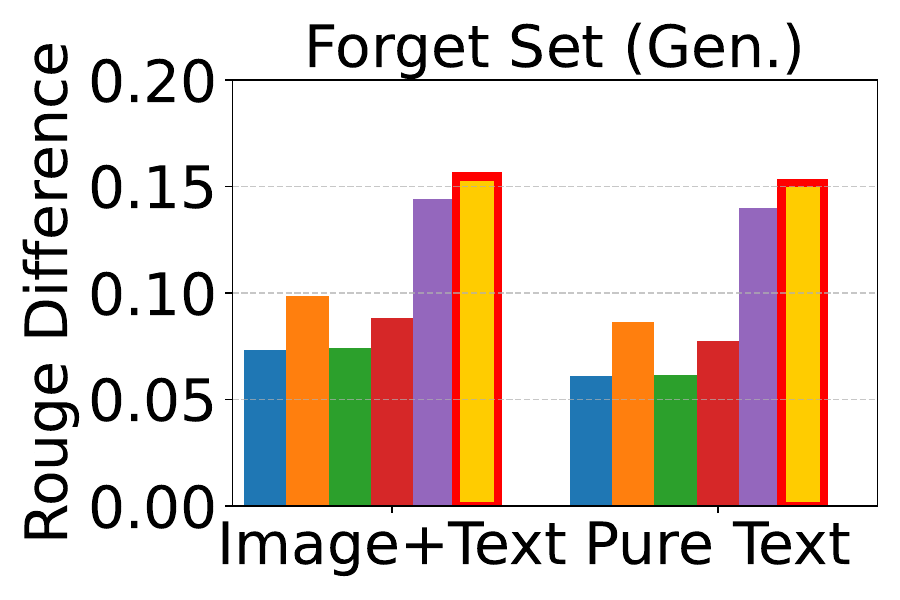}
    \subcaption{Forget Set (Generation)}
    \label{fig:llava_5_gen_forget}
\end{subfigure}
\begin{subfigure}{0.244\textwidth}
    \includegraphics[width=\textwidth]{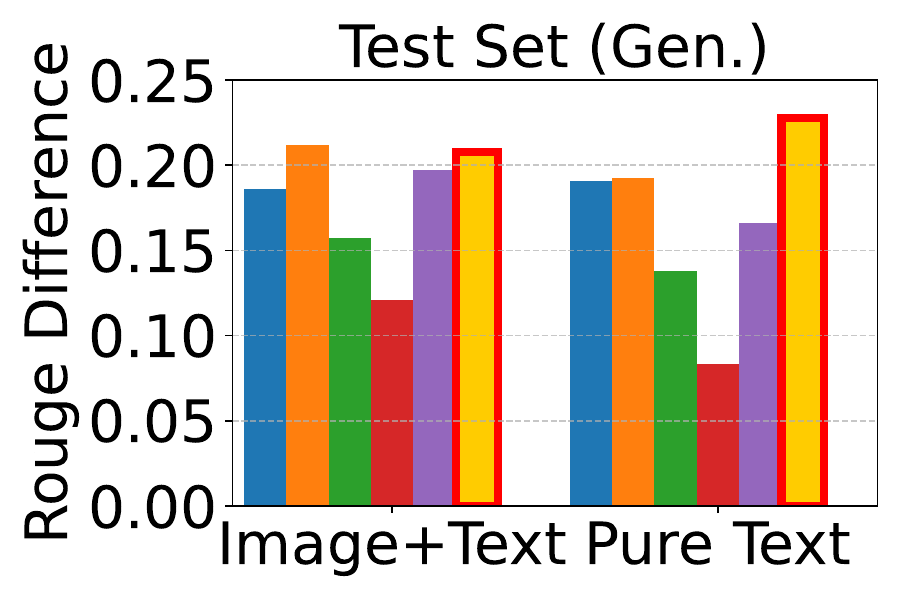}
    \subcaption{Test Set (Generation)}
    \label{fig:llava_5_gen_test}
\end{subfigure}
\begin{subfigure}{0.244\textwidth}
    \includegraphics[width=\textwidth]{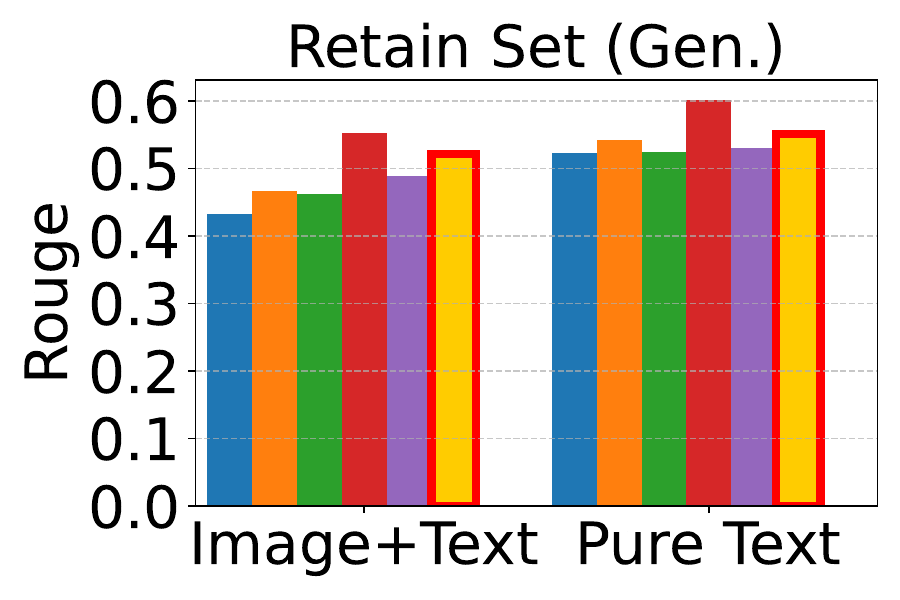}
    \subcaption{Retain Set (Generation)}
    \label{fig:llava_5_gen_retain}
\end{subfigure}
\begin{subfigure}{0.244\textwidth}
    \includegraphics[width=\textwidth]{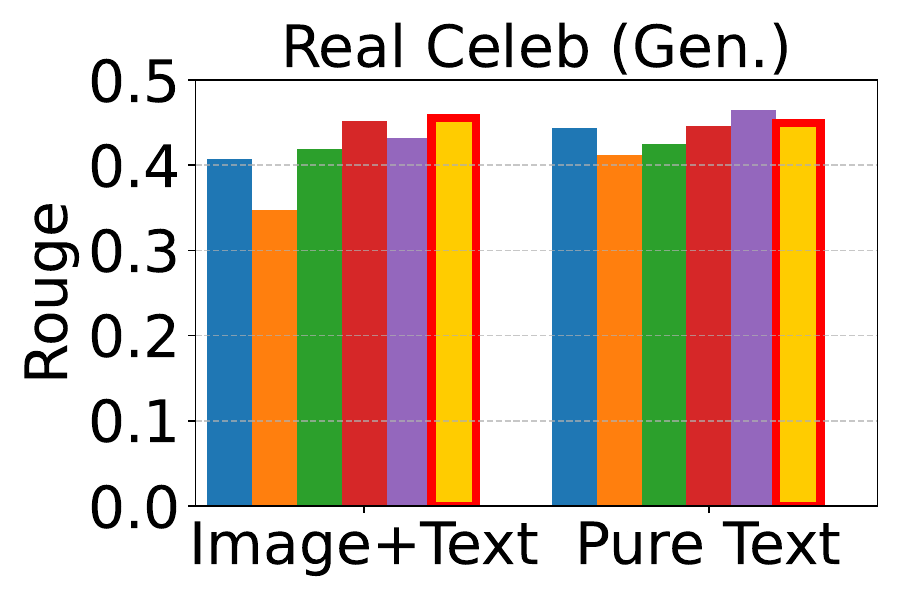}
    \subcaption{Real Celeb (Generation)}
    \label{fig:llava_5_gen_real}
\end{subfigure}

\begin{subfigure}{0.244\textwidth}
    \includegraphics[width=\textwidth]{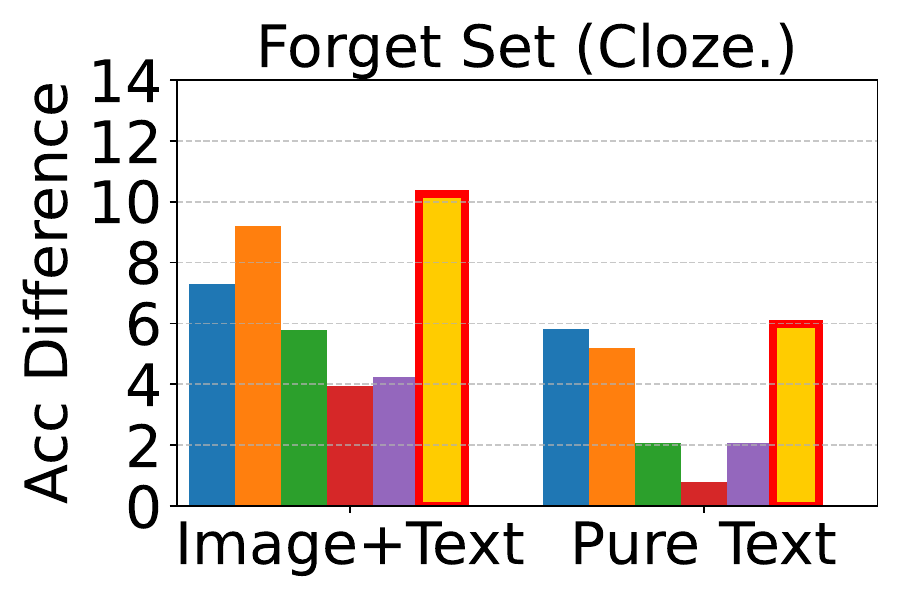}
    \subcaption{Forget Set (Cloze)}
    \label{fig:llava_5_cloze_forget}
\end{subfigure}
\begin{subfigure}{0.244\textwidth}
    \includegraphics[width=\textwidth]{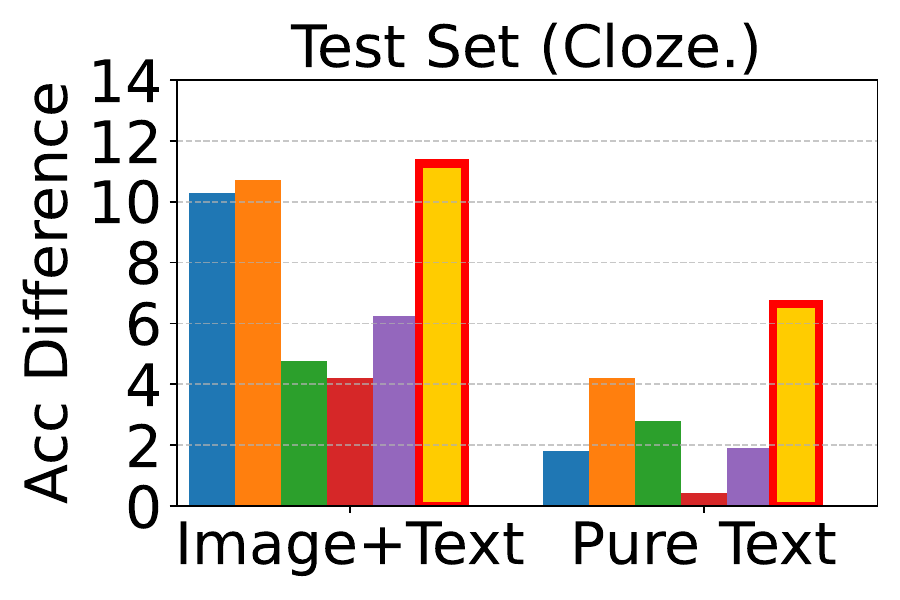}
    \subcaption{Test Set (Cloze)}
    \label{fig:llava_5_cloze_test}
\end{subfigure}
\begin{subfigure}{0.244\textwidth}
    \includegraphics[width=\textwidth]{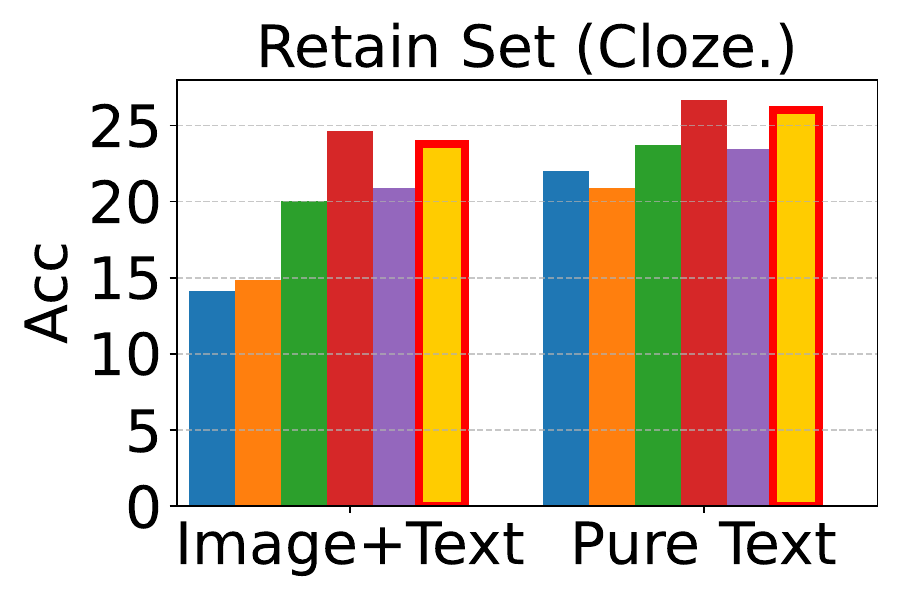}
    \subcaption{Retain Set (Cloze)}
    \label{fig:llava_5_cloze_retain}
\end{subfigure}
\begin{subfigure}{0.244\textwidth}
    \includegraphics[width=\textwidth]{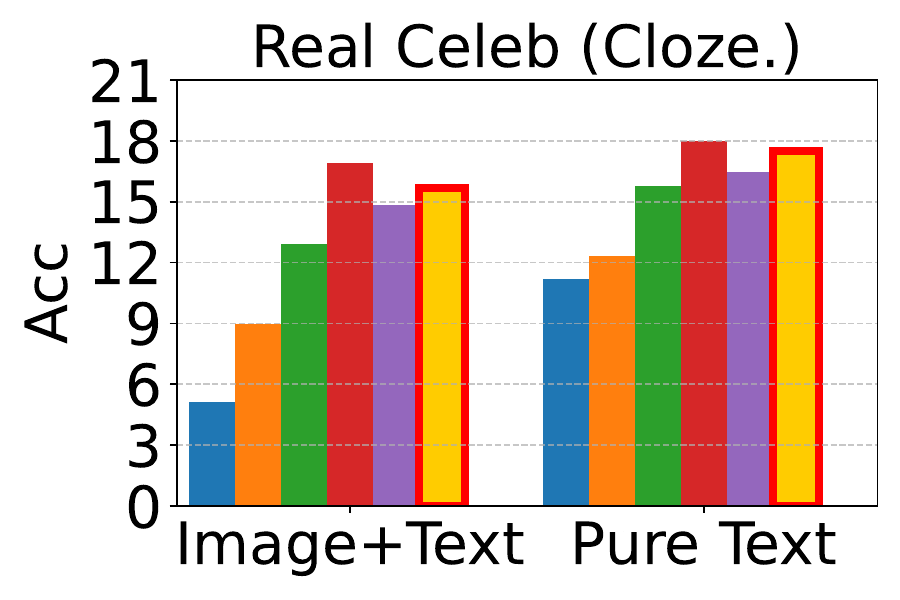}
    \subcaption{Real Celeb (Cloze)}
    \label{fig:llava_5_cloze_real}
\end{subfigure}
\vspace{-0.1in}
\caption{
Classification, generation, and cloze performance of \method and baselines in multimodal and unimodal setups with 5\% forget data, using LLaVA as the base model. In subplots (a), (b), (e), (f), (i), and (j), the $y$-axis represents the change in classification accuracy, ROUGE-L score, and cloze accuracy relative to the vanilla model, evaluated on the Forget and Test sets. In the remaining subplots, the $y$-axis indicates classification accuracy, ROUGE-L score, and cloze accuracy, respectively. The $x$-axis represents performance across different modalities.}
\vspace{-0.2in}
\label{fig:llava_5_compare}
\end{figure*}

As demonstrated in section \ref{sec:motivation}, the unique challenge of MLLM unlearning lies in the imbalanced effectiveness across modalities, where methods may exhibit strong performance on one but struggle on the other. 
Hence, this leads to the second question: \textbf{Does \method successfully address the unique challenge of imbalanced unlearning across different modalities in MLLMs?} To investigate this question, we decompose the average performance from multimodal and unimodal evaluation results in main table and analyze whether \method achieves more effective unlearning across different input modalities in MLLMU-Bench, as shown in Figure \ref{fig:llava_5_compare}. 
From figure \ref{fig:llava_5_compare}, we observe that certain unlearning methods, such as GA and Gradient Difference, demonstrate strong multimodal unlearning performance but struggle in unimodal evaluation (e.g., Figure \ref{fig:llava_5_class_forget}). 
This discrepancy highlights the entangled nature of knowledge across modalities, indicating that unlearning from multimodal inputs does not guarantee complete removal in unimodal settings. \textbf{Methods lacking modality-specific strategies may fail to erase target knowledge equally across modalities, leading to imbalanced unlearning.}

A similar imbalanced unlearning is observed in methods like KL Minimization and NPO, which exhibit stronger unlearning performance at multimodal level than in the unimodal setting.
In contrast, \method demonstrates the ability to unlearn target knowledge across both modalities, as evidenced by the balanced reduction in Forget/Test Set accuracy (e.g., Figure \ref{fig:llava_5_gen_forget}). Additional analysis for other data splits can be found in Appendix \ref{appendix:unlearn_across_modality}.

\subsection{Pruning Ratio Analysis}
In this section, we address the third question: \textbf{How do different pruning ratios affect the effectiveness of \method during the unlearning process?} 
To investigate this, we adjust the pruning ratios of the selected neurons to 2\%, 5\%, and 10\%, and observe the corresponding impact on overall performance. Table \ref{tab:prune-ratio-10} presents results for both models using the 10\% data split. Further experimental results can be referred to Appendix \ref{appendix:prune_ratio}. 
As the pruning ratio increases from 2\% to 10\%, we observe larger effects on both unlearning performance and model utility. For instance, with a 10\% pruning ratio in the LLaVA model, \method improves unlearning performance on the forget and test sets compared to 2\% pruning, reducing classification accuracy from 38.36\% to 34.81\% and from 37.24\% to 32.93\%, respectively.
However, this improvement comes at the cost of reduced model utility on the Retain and Real Celebrity Sets, with classification accuracy dropping from 44.89\% to 34.22\% and from 48.02\% to 43.10\%, respectively. A similar trend is observed in the Idefics2 model.
This result shows that higher pruning ratios enhance unlearning performance but disrupt the balance with utility, ultimately reducing model utility. This occurs because higher pruning ratios remove neurons that are less critical to the forget set but essential for preserving model utility across other datasets.

\begin{table*}[t!]
    \centering
\scalebox{0.51}{
\begin{tabular}{l|cccc|cccc|cccc|cccc}
        \toprule
        \multirow{3}{*}{\textbf{Models}} 
        & \multicolumn{4}{c|}{\textbf{Forget Set}} 
        & \multicolumn{4}{c|}{\textbf{Test Set}} 
        & \multicolumn{4}{c|}{\textbf{Retain Set}} 
        & \multicolumn{4}{c}{\textbf{Real Celebrity}} \\
        \cline{2-17}
        & \begin{tabular}[c]{@{}c@{}}Class.\\ Acc (\textcolor{blue}{$\downarrow$})\end{tabular} 
        & \begin{tabular}[c]{@{}c@{}}Rouge\\ Score (\textcolor{red}{$\downarrow$})\end{tabular} 
        & \begin{tabular}[c]{@{}c@{}}Fact.\\ Score (\textcolor{red}{$\downarrow$})\end{tabular} 
        & \begin{tabular}[c]{@{}c@{}}Cloze\\ Acc (\textcolor{teal}{$\downarrow$})\end{tabular} 
        & \begin{tabular}[c]{@{}c@{}}Class.\\ Acc (\textcolor{blue}{$\downarrow$})\end{tabular} 
        & \begin{tabular}[c]{@{}c@{}}Rouge\\ Score (\textcolor{red}{$\downarrow$})\end{tabular} 
        & \begin{tabular}[c]{@{}c@{}}Fact.\\ Score (\textcolor{red}{$\downarrow$})\end{tabular} 
        & \begin{tabular}[c]{@{}c@{}}Cloze\\ Acc (\textcolor{teal}{$\downarrow$})\end{tabular} 
        & \begin{tabular}[c]{@{}c@{}}Class.\\ Acc (\textcolor{blue}{$\uparrow$})\end{tabular} 
        & \begin{tabular}[c]{@{}c@{}}Rouge\\ Score (\textcolor{red}{$\uparrow$})\end{tabular} 
        & \begin{tabular}[c]{@{}c@{}}Fact.\\ Score (\textcolor{red}{$\uparrow$})\end{tabular} 
        & \begin{tabular}[c]{@{}c@{}}Cloze\\ Acc (\textcolor{teal}{$\uparrow$})\end{tabular} 
        & \begin{tabular}[c]{@{}c@{}}Class.\\ Acc (\textcolor{blue}{$\uparrow$})\end{tabular} 
        & \begin{tabular}[c]{@{}c@{}}Rouge\\ Score (\textcolor{red}{$\uparrow$})\end{tabular} 
        & \begin{tabular}[c]{@{}c@{}}Fact.\\ Score (\textcolor{red}{$\uparrow$})\end{tabular} 
        & \begin{tabular}[c]{@{}c@{}}Cloze\\ Acc (\textcolor{teal}{$\uparrow$})\end{tabular} \\
        \midrule
        \multicolumn{17}{c}{\textbf{LLaVA-1.5-7B (10\% Forget)}} \\
        \midrule

         Vanilla & 49.15\% & 0.594 & 6.40 & 26.97\% & 47.41\% & 0.510 & 5.20 & 25.43\% & 46.68\% & 0.582 & 5.44 & 28.49\% & 51.80\% & 0.479 & 5.47 & 17.35\% \\
        
        \method (2\%) & 38.36\% & 0.503 & 3.48 & 19.44\% & 37.24\% & 0.402 & 3.06 & 15.61\% & 44.89\% & 0.538 & 5.35 & 24.61\% & 48.02\% & 0.466 & 4.32 & 15.68\% \\
        
        \method (5\%) & 37.79\% & 0.468 & 3.27 & 18.47\% & 35.45\% & 0.388 & 3.05 & 11.20\% & 41.60\% & 0.478 & 4.39 & 22.61\% & 45.04\% & 0.421 & 3.54 & 11.78\% \\
        
        \method (10\%) & 34.81\% & 0.425 & 3.10 & 18.10\% & 32.93\% & 0.341 & 2.97 & 10.22\% & 34.22\% & 0.468 & 4.03 & 18.16\% & 43.10\% & 0.378 & 3.39 & 10.27\% \\
                
        \midrule

        \multicolumn{17}{c}{\textbf{Idefics2-8B (10\% Forget)}} \\
        \midrule

        Vanilla & 54.48\% & 0.645 & 6.27 & 46.55\% & 48.09\% & 0.492 & 5.36 & 27.81\% & 47.52\% & 0.643 & 6.63 & 43.37\% & 52.75\% & 0.459 & 5.75 & 20.05\% \\
        
        \method (2\%) & 36.49\% & 0.366 & 3.17 & 30.67\% & 35.32\% & 0.333 & 3.10 & 16.66\% & 43.95\% & 0.566 & 4.98 & 41.61\% & 50.81\% & 0.430 & 4.22 & 19.26\% \\
        
        \method (5\%) & 35.11\% & 0.342 & 3.05 & 28.88\% & 34.27\% & 0.329 & 3.08 & 15.58\% & 39.63\% & 0.507 & 4.53 & 40.59\%& 49,10\% & 0.427 & 4.12& 18.00\% \\
        
        \method (10\%) & 30.24\% & 0.297 & 2.99 & 26.53\% & 31.33\% & 0.305 & 3.01 & 14.72\% & 35.77\% & 0.501 & 4.47 & 38.19\% & 48.87\% & 0.419 & 4.06 & 17.94\% \\

        \bottomrule
    \end{tabular}}
    \vspace{-0.1in}
    \caption{Overall results of \method with varying pruning ratios on two base MLLM models under a 10\% forget data setup. For each MLLM, the pruning ratio is iteratively increased from 2\% to 10\%. }
    \label{tab:prune-ratio-10}
    \vspace{-0.15in}
\end{table*}

\subsection{Unlearning v.s. Model Utility}

\begin{figure*}
\centering
\begin{subfigure}[b]{\textwidth}
    \centering    \includegraphics[width=0.9\textwidth]{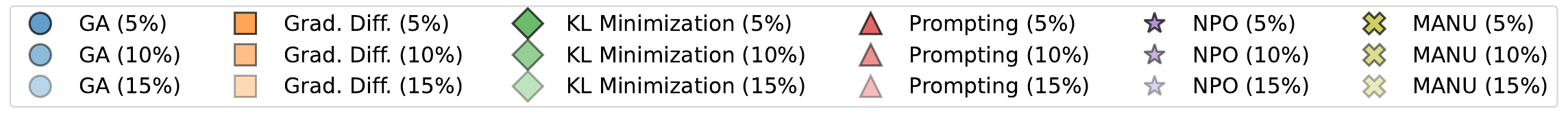}
\end{subfigure}
\begin{subfigure}{0.244\textwidth}
    \includegraphics[width=\textwidth]{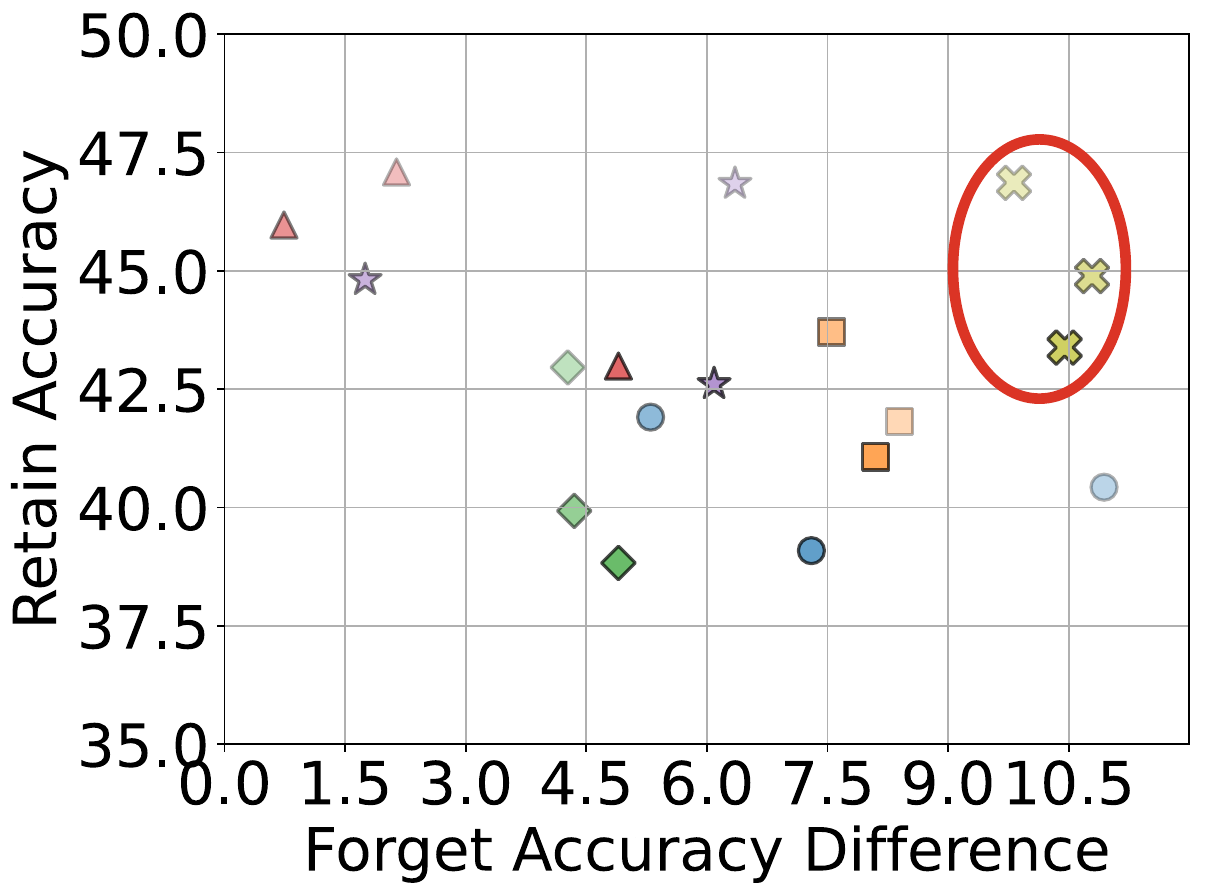}
    \subcaption{Forget Acc vs Retain Acc}
    \label{fig:llava_forget_retain}
\end{subfigure}    
\begin{subfigure}{0.244\textwidth}
    \includegraphics[width=\textwidth]{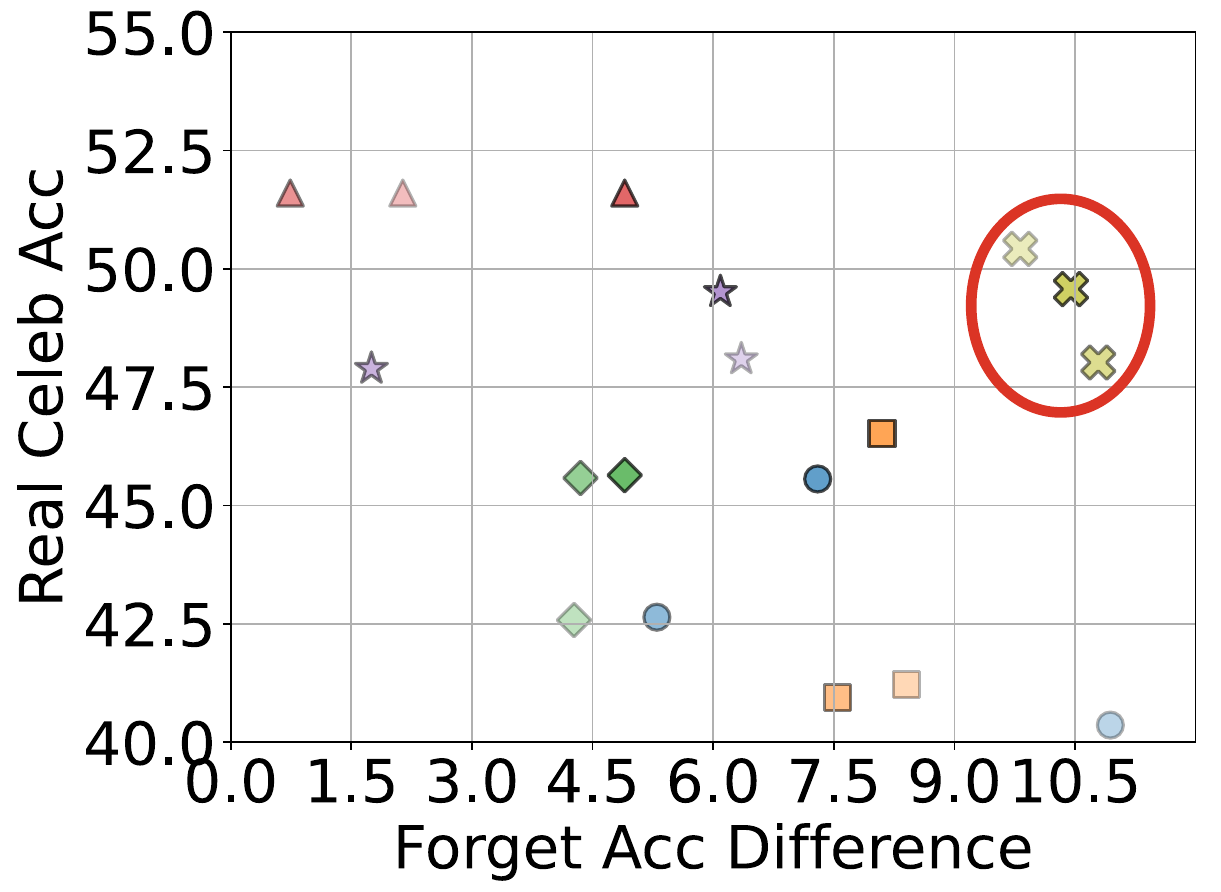}
    \subcaption{Forget Acc vs Real Celeb}
    \label{fig:llava_forget_real}
\end{subfigure}
\begin{subfigure}{0.244\textwidth}
    \includegraphics[width=\textwidth]{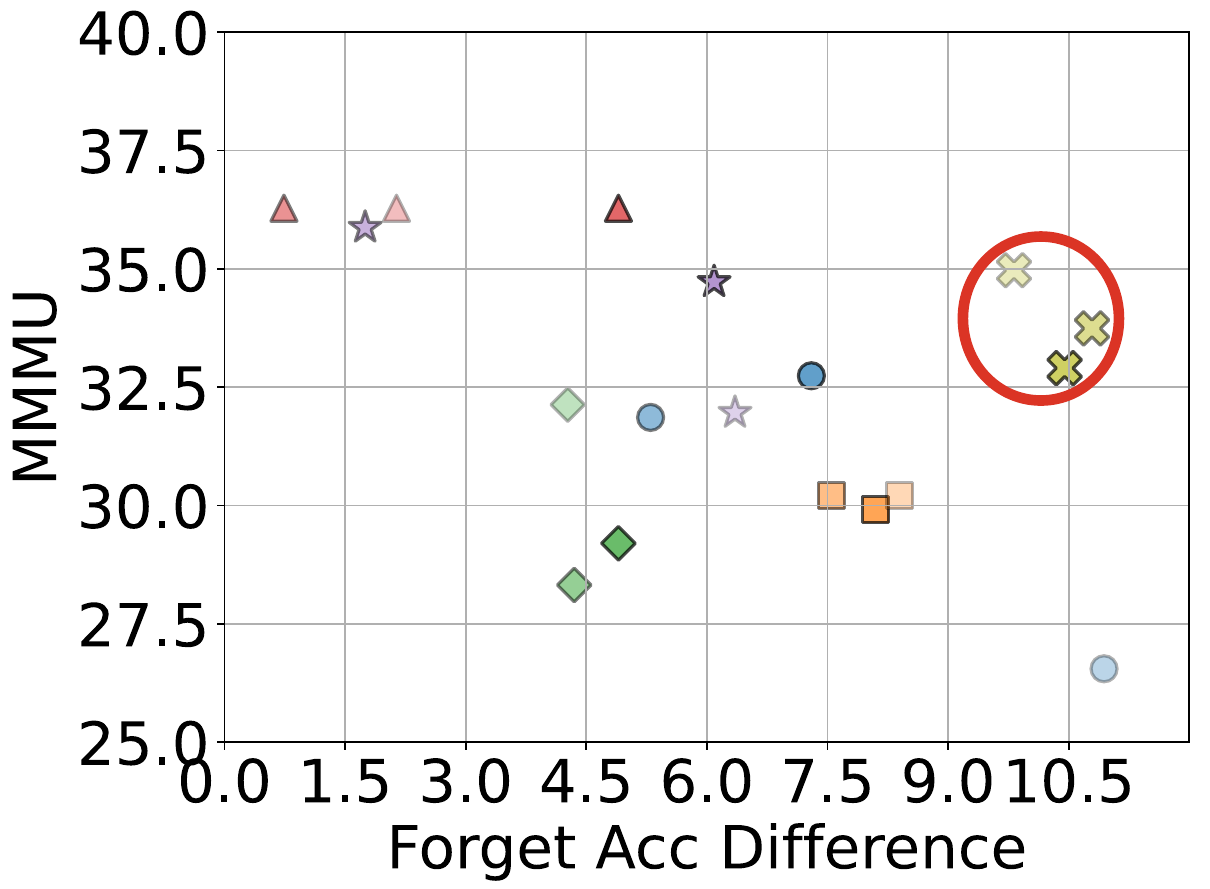}
    \subcaption{Forget Acc vs MMMU}
    \label{fig:llava_forget_mmmu}
\end{subfigure}
\begin{subfigure}{0.244\textwidth}
    \includegraphics[width=\textwidth]{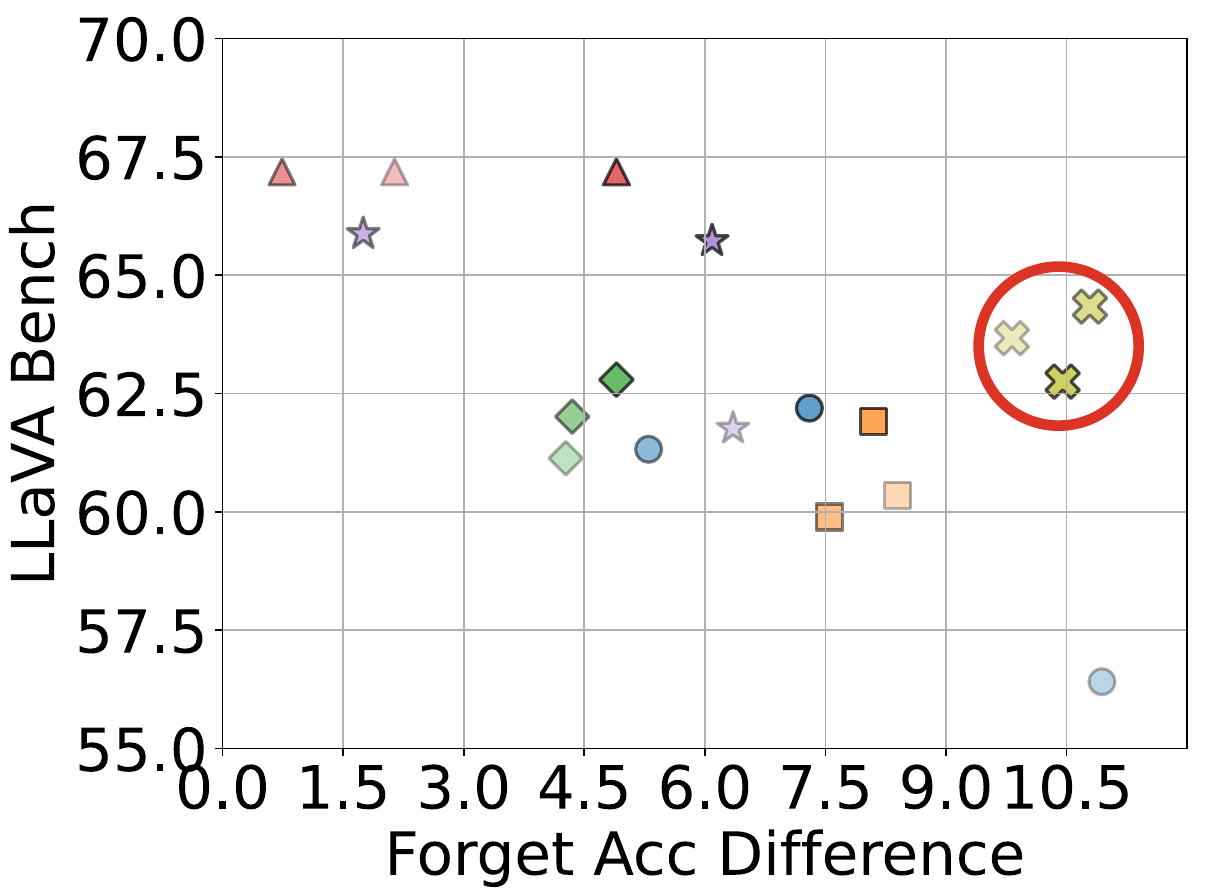}
    \subcaption{Forget Acc vs LLaVABench}
    \label{fig:llava_forget_llavaB}
\end{subfigure}
\vspace{-0.1in}
\caption{
The overall trade-off between unlearning effectiveness and model utility across all baselines using different forget data, with LLaVA as the base model. The $x$-axis shows the difference in forget classification accuracy relative to the vanilla model, while the $y$-axis reflects model utility from various perspectives. From left to right, these perspectives include retain accuracy, real celebrity accuracy, MMMU, and LLaVA-Bench performance, respectively.}
\vspace{-0.20in}
\label{fig:llava_class_tradeoff}
\end{figure*}

Lastly, balancing unlearning and model utility remains a critical challenge in the field of unlearning. Hence, \textbf{can \method achieve a good balance between unlearning the target knowledge and preserving the model's utility?}
Similar to MLLMU-Bench, we decompose "model utility" into three perspectives: retain accuracy, neighboring concepts (Real Celebrity Set), and general model abilities, including reasoning and helpfulness, which are evaluated using MMMU \cite{yue2024mmmu} and LLaVA-Bench \cite{liu2024visual}. The results are shown in Figure \ref{fig:llava_class_tradeoff} from left to right. 

From the figures, we observe that \method maintains a robust balance between unlearning performance and model utility across various aspects. The better an algorithm balances these two aspects, the closer it will appear to the top-right in the figure, indicating a larger difference in forget accuracy and higher retain accuracy. 
For example, in Figures \ref{fig:llava_forget_retain} and \ref{fig:llava_forget_real}, \method achieves a comparable reduction in Forget Set accuracy to GA-based approaches while maintaining high accuracy on the Retain and Real Celebrity sets. 
Similarly, when evaluating model reasoning abilities using MMMU and Llava-Bench (i.e. Figures \ref{fig:llava_forget_mmmu} and \ref{fig:llava_forget_llavaB}), \method performs comparably to prompting techniques while significantly surpassing them in forget set accuracy.
Thus, \method effectively balances unlearning and model utility across multiple dimensions. Further analysis and additional ablation studies can be found in Appendix \ref{appendix:unlearn_vs_utility} and \ref{appendix:ablations_importance_functions}.

\vspace{-0.05in}
\section{Related Work}
\paragraph{MU for Generative Models.}
As LLMs and MLLMs memorize large amounts of sensitive knowledge during pre-training and fine-tuning, privacy concerns have grown with the rise of generative models \cite{liu2024shield, nasr2023scalable, liu2024machine, zhang2023counterfactual}. 
Machine Unlearning (MU) offers an efficient solution to selectively erase unwanted information while preserving overall model performance. \citet{yao2023large} formalized unlearning objectives for LLMs, introducing a gradient-ascent-based approach to remove harmful knowledge. 
To address catastrophic forgetting, task vector-based approaches have been proposed \cite{ilharco2022editing, liu2024towards, dou2024avoiding}. In response to the \textit{Right to Be Forgotten} \cite{dang2021right, bourtoule2021machine}, benchmarks like TOFU \cite{maini2024tofu} and MLLMU-Bench \cite{liu2024protecting} were developed using synthetic data, highlighting the need for privacy-preserving methods. However, existing unlearning algorithms are not explicitly designed for MLLMs to achieve comprehensive unlearning across modalities.


\paragraph{Model Pruning.}
Model pruning has proven to be an effective approach for removing redundant weights to enhance the performance and efficiency of a model. 
For example, \citet{conmy2023towards} proposes a weight pruning-based technique to identify sub-circuits that contribute most to a specific dataset. Additionally, pruning can be used to preserve key model capabilities while reducing computational costs. 
For instance, \citet{michel2019sixteen} introduces a method to prune unused attention heads without impacting overall performance. 
\citet{pochinkov2024dissecting} shows that pruning can be used to unlearn specific behaviors of transformer models through a selective neuron approach. 
Additionally, it empirically demonstrates the effectiveness of neuron pruning over weight pruning. However, without a modality-specific pruning strategy, achieving thorough unlearning to remove target knowledge across different modalities remains challenging.

\section{Conclusion}
In this work, we address the challenge of imbalanced unlearning in MLLMs, which arises due to distinct knowledge distributions and activation patterns across vision and language pathways. To tackle this, we propose \method, a modality-aware neuron pruning framework that ensures balanced unlearning across modalities while preserving model utility. Our approach first applies four importance functions to analyze neuron activations in MLP layers, then employs a scoring function to identify and prune neurons most associated with the targeted forget knowledge. Our results across multiple MLLMs demonstrate the efficacy of \method in achieving comprehensive unlearning while maintaining the model utility.

\section{Limitations}
\paragraph{Adaptations to other applications} Our method is primarily designed to remove sensitive profiles in MLLMU-Bench, where all profiles are fictitious and fine-tuned on the vanilla model. However, it would be valuable to explore how this pruning approach could be extended to unlearn other behaviors of MLLM models, such as harmful generations and copyright infringements. Additionally, while \method is specifically designed for MLLMs, its adaptation and performance on unimodal unlearning benchmarks, such as TOFU \cite{maini2024tofu} and WMDP \cite{li2024wmdp}, remain unexplored, which we leave for future work. We hope this study serves as a foundation to inspire future research toward developing \textbf{a model-agnostic unlearning framework}.
\paragraph{Robustness of Machine Unlearning} Although factuality score is used as one of the evaluation metrics, ROUGE score remains an important measure in our unlearning setting. However, as highlighted by recent work \cite{ippolito2022preventing}, it may create a false sense of privacy. Additionally, the robustness of \method against various attacks requires further validation and exploration, which is crucial as emphasized in prior studies \cite{lucki2024adversarial, cooper2024machine}.
\paragraph{Potential Instability} Furthermore, as discussed in Section \ref{sec: discussion}, variations in the pruning ratio significantly affect both unlearning performance and model utility. While \method demonstrates superior performance across tasks, it has yet to achieve an optimal balance between unlearning effectiveness and model utility. Thus, we position \method as \textbf{a preliminary study} showcasing the benefits of a modality-aware design for MLLM unlearning, laying the foundation for more robust and stable approaches in future research.

\section{Acknowledgment}
This work was partially supported by NSF IIS-2119531, IIS-2137396, IIS-2142827, IIS-2234058, and ONR N00014-22-1-2507.

\bibliography{ref}

\begin{thebibliography}{66}
\providecommand{\natexlab}[1]{#1}

\bibitem[{Bai et~al.(2022)Bai, Jones, Ndousse, Askell, Chen, DasSarma, Drain, Fort, Ganguli, Henighan et~al.}]{bai2022training}
Yuntao Bai, Andy Jones, Kamal Ndousse, Amanda Askell, Anna Chen, Nova DasSarma, Dawn Drain, Stanislav Fort, Deep Ganguli, Tom Henighan, et~al. 2022.
\newblock Training a helpful and harmless assistant with reinforcement learning from human feedback.
\newblock \emph{arXiv preprint arXiv:2204.05862}.

\bibitem[{Bourtoule et~al.(2021)Bourtoule, Chandrasekaran, Choquette-Choo, Jia, Travers, Zhang, Lie, and Papernot}]{bourtoule2021machine}
Lucas Bourtoule, Varun Chandrasekaran, Christopher~A Choquette-Choo, Hengrui Jia, Adelin Travers, Baiwu Zhang, David Lie, and Nicolas Papernot. 2021.
\newblock Machine unlearning.
\newblock In \emph{2021 IEEE Symposium on Security and Privacy (SP)}.

\bibitem[{Brown et~al.(2020)Brown, Mann, Ryder, Subbiah, Kaplan, Dhariwal, Neelakantan, Shyam, Sastry, Askell et~al.}]{brown2020language}
Tom Brown, Benjamin Mann, Nick Ryder, Melanie Subbiah, Jared~D Kaplan, Prafulla Dhariwal, Arvind Neelakantan, Pranav Shyam, Girish Sastry, Amanda Askell, et~al. 2020.
\newblock Language models are few-shot learners.
\newblock \emph{Neurips}.

\bibitem[{Carlini et~al.(2021)Carlini, Tramer, Wallace, Jagielski, Herbert-Voss, Lee, Roberts, Brown, Song, Erlingsson et~al.}]{carlini2021extracting}
Nicholas Carlini, Florian Tramer, Eric Wallace, Matthew Jagielski, Ariel Herbert-Voss, Katherine Lee, Adam Roberts, Tom Brown, Dawn Song, Ulfar Erlingsson, et~al. 2021.
\newblock Extracting training data from large language models.
\newblock In \emph{USENIX Security 21}.

\bibitem[{Chen et~al.(2024)Chen, Li, Dong, Zhang, Zang, Chen, Duan, Wang, Qiao, Lin et~al.}]{chen2024we}
Lin Chen, Jinsong Li, Xiaoyi Dong, Pan Zhang, Yuhang Zang, Zehui Chen, Haodong Duan, Jiaqi Wang, Yu~Qiao, Dahua Lin, et~al. 2024.
\newblock Are we on the right way for evaluating large vision-language models?
\newblock \emph{arXiv preprint arXiv:2403.20330}.

\bibitem[{Chowdhery et~al.(2023)Chowdhery, Narang, Devlin, Bosma, Mishra, Roberts, Barham, Chung, Sutton, Gehrmann et~al.}]{chowdhery2023palm}
Aakanksha Chowdhery, Sharan Narang, Jacob Devlin, Maarten Bosma, Gaurav Mishra, Adam Roberts, Paul Barham, Hyung~Won Chung, Charles Sutton, Sebastian Gehrmann, et~al. 2023.
\newblock Palm: Scaling language modeling with pathways.
\newblock \emph{JMLR}.

\bibitem[{Conmy et~al.(2023)Conmy, Mavor-Parker, Lynch, Heimersheim, and Garriga-Alonso}]{conmy2023towards}
Arthur Conmy, Augustine Mavor-Parker, Aengus Lynch, Stefan Heimersheim, and Adri{\`a} Garriga-Alonso. 2023.
\newblock Towards automated circuit discovery for mechanistic interpretability.
\newblock \emph{Neurips}.

\bibitem[{Cooper et~al.(2024)Cooper, Choquette-Choo, Bogen, Jagielski, Filippova, Liu, Chouldechova, Hayes, Huang, Mireshghallah et~al.}]{cooper2024machine}
A~Feder Cooper, Christopher~A Choquette-Choo, Miranda Bogen, Matthew Jagielski, Katja Filippova, Ken~Ziyu Liu, Alexandra Chouldechova, Jamie Hayes, Yangsibo Huang, Niloofar Mireshghallah, et~al. 2024.
\newblock Machine unlearning doesn't do what you think: Lessons for generative ai policy, research, and practice.
\newblock \emph{arXiv preprint arXiv:2412.06966}.

\bibitem[{Dang(2021)}]{dang2021right}
Quang-Vinh Dang. 2021.
\newblock Right to be forgotten in the age of machine learning.
\newblock In \emph{Advances in Digital Science: ICADS 2021}.

\bibitem[{Dou et~al.(2024)Dou, Liu, Lyu, Ding, and Wong}]{dou2024avoiding}
Guangyao Dou, Zheyuan Liu, Qing Lyu, Kaize Ding, and Eric Wong. 2024.
\newblock Avoiding copyright infringement via machine unlearning.
\newblock \emph{arXiv preprint arXiv:2406.10952}.

\bibitem[{Duarte et~al.(2024)Duarte, Zhao, Oliveira, and Li}]{duarte2024cop}
Andr{\'e}~V Duarte, Xuandong Zhao, Arlindo~L Oliveira, and Lei Li. 2024.
\newblock De-cop: Detecting copyrighted content in language models training data.
\newblock \emph{arXiv preprint arXiv:2402.09910}.

\bibitem[{Fu et~al.(2024)Fu, Yu, Li, Qian, Zhang, Yuan, Shi, Yakunin, and Lin}]{fu2024amoeballm}
Yonggan Fu, Zhongzhi Yu, Junwei Li, Jiayi Qian, Yongan Zhang, Xiangchi Yuan, Dachuan Shi, Roman Yakunin, and Yingyan~Celine Lin. 2024.
\newblock Amoeballm: Constructing any-shape large language models for efficient and instant deployment.
\newblock \emph{arXiv preprint arXiv:2411.10606}.

\bibitem[{Ghiasi et~al.(2022)Ghiasi, Kazemi, Borgnia, Reich, Shu, Goldblum, Wilson, and Goldstein}]{ghiasi2022vision}
Amin Ghiasi, Hamid Kazemi, Eitan Borgnia, Steven Reich, Manli Shu, Micah Goldblum, Andrew~Gordon Wilson, and Tom Goldstein. 2022.
\newblock What do vision transformers learn? a visual exploration.
\newblock \emph{arXiv preprint arXiv:2212.06727}.

\bibitem[{Huang et~al.(2024)Huang, Wang, Zhao, and Liu}]{huang2024commonsense}
Xiusheng Huang, Yequan Wang, Jun Zhao, and Kang Liu. 2024.
\newblock Commonsense knowledge editing based on free-text in llms.
\newblock \emph{arXiv preprint arXiv:2410.23844}.

\bibitem[{Ilharco et~al.(2022)Ilharco, Ribeiro, Wortsman, Gururangan, Schmidt, Hajishirzi, and Farhadi}]{ilharco2022editing}
Gabriel Ilharco, Marco~Tulio Ribeiro, Mitchell Wortsman, Suchin Gururangan, Ludwig Schmidt, Hannaneh Hajishirzi, and Ali Farhadi. 2022.
\newblock Editing models with task arithmetic.
\newblock \emph{arXiv preprint arXiv:2212.04089}.

\bibitem[{Ippolito et~al.(2022)Ippolito, Tram{\`e}r, Nasr, Zhang, Jagielski, Lee, Choquette-Choo, and Carlini}]{ippolito2022preventing}
Daphne Ippolito, Florian Tram{\`e}r, Milad Nasr, Chiyuan Zhang, Matthew Jagielski, Katherine Lee, Christopher~A Choquette-Choo, and Nicholas Carlini. 2022.
\newblock Preventing verbatim memorization in language models gives a false sense of privacy.
\newblock \emph{arXiv preprint arXiv:2210.17546}.

\bibitem[{Joshi et~al.(2024)Joshi, Saha, Shukla, Vema, Jhamtani, Gaur, and Modi}]{joshi2024towards}
Abhinav Joshi, Shaswati Saha, Divyaksh Shukla, Sriram Vema, Harsh Jhamtani, Manas Gaur, and Ashutosh Modi. 2024.
\newblock Towards robust evaluation of unlearning in llms via data transformations.
\newblock \emph{arXiv preprint arXiv:2411.15477}.

\bibitem[{Lauren{\c{c}}on et~al.(2024)Lauren{\c{c}}on, Tronchon, Cord, and Sanh}]{laurenccon2024matters}
Hugo Lauren{\c{c}}on, L{\'e}o Tronchon, Matthieu Cord, and Victor Sanh. 2024.
\newblock What matters when building vision-language models?
\newblock \emph{arXiv preprint arXiv:2405.02246}.

\bibitem[{Li et~al.(2024)Li, Pan, Gopal, Yue, Berrios, Gatti, Li, Dombrowski, Goel, Phan et~al.}]{li2024wmdp}
Nathaniel Li, Alexander Pan, Anjali Gopal, Summer Yue, Daniel Berrios, Alice Gatti, Justin~D Li, Ann-Kathrin Dombrowski, Shashwat Goel, Long Phan, et~al. 2024.
\newblock The wmdp benchmark: Measuring and reducing malicious use with unlearning.
\newblock \emph{arXiv preprint arXiv:2403.03218}.

\bibitem[{Lin(2004)}]{lin2004rouge}
Chin-Yew Lin. 2004.
\newblock Rouge: A package for automatic evaluation of summaries.
\newblock In \emph{Text summarization branches out}.

\bibitem[{Liu et~al.(2022)Liu, Liu, and Stone}]{liu2022continual}
Bo~Liu, Qiang Liu, and Peter Stone. 2022.
\newblock Continual learning and private unlearning.
\newblock In \emph{CoLLAs}.

\bibitem[{Liu et~al.(2024{\natexlab{a}})Liu, Li, Li, and Lee}]{liu2024improved}
Haotian Liu, Chunyuan Li, Yuheng Li, and Yong~Jae Lee. 2024{\natexlab{a}}.
\newblock Improved baselines with visual instruction tuning.
\newblock In \emph{CVPR}.

\bibitem[{Liu et~al.(2024{\natexlab{b}})Liu, Li, Wu, and Lee}]{liu2024visual}
Haotian Liu, Chunyuan Li, Qingyang Wu, and Yong~Jae Lee. 2024{\natexlab{b}}.
\newblock Visual instruction tuning.
\newblock \emph{Neurips}.

\bibitem[{Liu et~al.(2024{\natexlab{c}})Liu, Yao, Jia, Casper, Baracaldo, Hase, Yao, Liu, Xu, Li et~al.}]{liu2024rethinking}
Sijia Liu, Yuanshun Yao, Jinghan Jia, Stephen Casper, Nathalie Baracaldo, Peter Hase, Yuguang Yao, Chris~Yuhao Liu, Xiaojun Xu, Hang Li, et~al. 2024{\natexlab{c}}.
\newblock Rethinking machine unlearning for large language models.
\newblock \emph{arXiv preprint arXiv:2402.08787}.

\bibitem[{Liu et~al.(2024{\natexlab{d}})Liu, Sun, Xu, Wu, Wang, Wang, and Gao}]{liu2024shield}
Xiaoze Liu, Ting Sun, Tianyang Xu, Feijie Wu, Cunxiang Wang, Xiaoqian Wang, and Jing Gao. 2024{\natexlab{d}}.
\newblock Shield: Evaluation and defense strategies for copyright compliance in llm text generation.
\newblock \emph{arXiv preprint arXiv:2406.12975}.

\bibitem[{Liu et~al.(2024{\natexlab{e}})Liu, Dou, Jia, Tan, Zeng, Yuan, and Jiang}]{liu2024protecting}
Zheyuan Liu, Guangyao Dou, Mengzhao Jia, Zhaoxuan Tan, Qingkai Zeng, Yongle Yuan, and Meng Jiang. 2024{\natexlab{e}}.
\newblock Protecting privacy in multimodal large language models with mllmu-bench.
\newblock \emph{arXiv preprint arXiv:2410.22108}.

\bibitem[{Liu et~al.(2024{\natexlab{f}})Liu, Dou, Tan, Tian, and Jiang}]{liu2024machine}
Zheyuan Liu, Guangyao Dou, Zhaoxuan Tan, Yijun Tian, and Meng Jiang. 2024{\natexlab{f}}.
\newblock Machine unlearning in generative ai: A survey.
\newblock \emph{arXiv preprint arXiv:2407.20516}.

\bibitem[{Liu et~al.(2024{\natexlab{g}})Liu, Dou, Tan, Tian, and Jiang}]{liu2024towards}
Zheyuan Liu, Guangyao Dou, Zhaoxuan Tan, Yijun Tian, and Meng Jiang. 2024{\natexlab{g}}.
\newblock Towards safer large language models through machine unlearning.
\newblock \emph{arXiv preprint arXiv:2402.10058}.

\bibitem[{Liu et~al.(2023)Liu, Wang, Dao, Zhou, Yuan, Song, Shrivastava, Zhang, Tian, Re et~al.}]{liu2023deja}
Zichang Liu, Jue Wang, Tri Dao, Tianyi Zhou, Binhang Yuan, Zhao Song, Anshumali Shrivastava, Ce~Zhang, Yuandong Tian, Christopher Re, et~al. 2023.
\newblock Deja vu: Contextual sparsity for efficient llms at inference time.
\newblock In \emph{ICML}.

\bibitem[{{\L}ucki et~al.(2024){\L}ucki, Wei, Huang, Henderson, Tram{\`e}r, and Rando}]{lucki2024adversarial}
Jakub {\L}ucki, Boyi Wei, Yangsibo Huang, Peter Henderson, Florian Tram{\`e}r, and Javier Rando. 2024.
\newblock An adversarial perspective on machine unlearning for ai safety.
\newblock \emph{arXiv preprint arXiv:2409.18025}.

\bibitem[{Maini et~al.(2024)Maini, Feng, Schwarzschild, Lipton, and Kolter}]{maini2024tofu}
Pratyush Maini, Zhili Feng, Avi Schwarzschild, Zachary~C Lipton, and J~Zico Kolter. 2024.
\newblock Tofu: A task of fictitious unlearning for llms.
\newblock \emph{arXiv preprint arXiv:2401.06121}.

\bibitem[{Meng et~al.(2022{\natexlab{a}})Meng, Bau, Andonian, and Belinkov}]{meng2022locating}
Kevin Meng, David Bau, Alex Andonian, and Yonatan Belinkov. 2022{\natexlab{a}}.
\newblock Locating and editing factual associations in gpt.
\newblock \emph{Neurips}.

\bibitem[{Meng et~al.(2022{\natexlab{b}})Meng, Sharma, Andonian, Belinkov, and Bau}]{meng2022mass}
Kevin Meng, Arnab~Sen Sharma, Alex Andonian, Yonatan Belinkov, and David Bau. 2022{\natexlab{b}}.
\newblock Mass-editing memory in a transformer.
\newblock \emph{arXiv preprint arXiv:2210.07229}.

\bibitem[{Michel et~al.(2019)Michel, Levy, and Neubig}]{michel2019sixteen}
Paul Michel, Omer Levy, and Graham Neubig. 2019.
\newblock Are sixteen heads really better than one?
\newblock \emph{Neurips}.

\bibitem[{Nasr et~al.(2023)Nasr, Carlini, Hayase, Jagielski, Cooper, Ippolito, Choquette-Choo, Wallace, Tram{\`e}r, and Lee}]{nasr2023scalable}
Milad Nasr, Nicholas Carlini, Jonathan Hayase, Matthew Jagielski, A~Feder Cooper, Daphne Ippolito, Christopher~A Choquette-Choo, Eric Wallace, Florian Tram{\`e}r, and Katherine Lee. 2023.
\newblock Scalable extraction of training data from (production) language models.
\newblock \emph{arXiv preprint arXiv:2311.17035}.

\bibitem[{Nguyen et~al.(2020)Nguyen, Low, and Jaillet}]{nguyen2020variational}
Quoc~Phong Nguyen, Bryan Kian~Hsiang Low, and Patrick Jaillet. 2020.
\newblock Variational bayesian unlearning.
\newblock \emph{Neurips}.

\bibitem[{Nguyen et~al.(2022)Nguyen, Huynh, Ren, Nguyen, Liew, Yin, and Nguyen}]{nguyen2022survey}
Thanh~Tam Nguyen, Thanh~Trung Huynh, Zhao Ren, Phi~Le Nguyen, Alan Wee-Chung Liew, Hongzhi Yin, and Quoc Viet~Hung Nguyen. 2022.
\newblock A survey of machine unlearning.
\newblock \emph{arXiv preprint arXiv:2209.02299}.

\bibitem[{Ni et~al.(2025)Ni, Liu, Wang, Lei, Zhao, Cheng, Zeng, Dong, Xia, Kenthapadi et~al.}]{ni2025towards}
Bo~Ni, Zheyuan Liu, Leyao Wang, Yongjia Lei, Yuying Zhao, Xueqi Cheng, Qingkai Zeng, Luna Dong, Yinglong Xia, Krishnaram Kenthapadi, et~al. 2025.
\newblock Towards trustworthy retrieval augmented generation for large language models: A survey.
\newblock \emph{arXiv preprint arXiv:2502.06872}.

\bibitem[{Ouyang et~al.(2022)Ouyang, Wu, Jiang, Almeida, Wainwright, Mishkin, Zhang, Agarwal, Slama, Ray et~al.}]{ouyang2022training}
Long Ouyang, Jeffrey Wu, Xu~Jiang, Diogo Almeida, Carroll Wainwright, Pamela Mishkin, Chong Zhang, Sandhini Agarwal, Katarina Slama, Alex Ray, et~al. 2022.
\newblock Training language models to follow instructions with human feedback.
\newblock \emph{Neurips}.

\bibitem[{Papantoniou et~al.(2024)Papantoniou, Lattas, Moschoglou, Deng, Kainz, and Zafeiriou}]{papantoniou2024arc2face}
Foivos~Paraperas Papantoniou, Alexandros Lattas, Stylianos Moschoglou, Jiankang Deng, Bernhard Kainz, and Stefanos Zafeiriou. 2024.
\newblock Arc2face: A foundation model for id-consistent human faces.
\newblock In \emph{ECCV}.

\bibitem[{Pochinkov and Schoots(2024)}]{pochinkov2024dissecting}
Nicholas Pochinkov and Nandi Schoots. 2024.
\newblock Dissecting language models: Machine unlearning via selective pruning.
\newblock \emph{arXiv preprint arXiv:2403.01267}.

\bibitem[{Qian et~al.(2024)Qian, Ye, Fauconnier, Grasch, Yang, and Gan}]{qian2024mia}
Yusu Qian, Hanrong Ye, Jean-Philippe Fauconnier, Peter Grasch, Yinfei Yang, and Zhe Gan. 2024.
\newblock Mia-bench: Towards better instruction following evaluation of multimodal llms.
\newblock \emph{arXiv preprint arXiv:2407.01509}.

\bibitem[{Qin et~al.(2023)Qin, Zhang, Zhang, Chen, Yasunaga, and Yang}]{qin2023chatgpt}
Chengwei Qin, Aston Zhang, Zhuosheng Zhang, Jiaao Chen, Michihiro Yasunaga, and Diyi Yang. 2023.
\newblock Is chatgpt a general-purpose natural language processing task solver?
\newblock \emph{arXiv preprint arXiv:2302.06476}.

\bibitem[{Rafailov et~al.(2024)Rafailov, Sharma, Mitchell, Manning, Ermon, and Finn}]{rafailov2024direct}
Rafael Rafailov, Archit Sharma, Eric Mitchell, Christopher~D Manning, Stefano Ermon, and Chelsea Finn. 2024.
\newblock Direct preference optimization: Your language model is secretly a reward model.
\newblock \emph{Neurips}.

\bibitem[{Sun et~al.(2023)Sun, Shen, Cao, Liu, Li, Shen, Gan, Gui, Wang, Yang et~al.}]{sun2023aligning}
Zhiqing Sun, Sheng Shen, Shengcao Cao, Haotian Liu, Chunyuan Li, Yikang Shen, Chuang Gan, Liang-Yan Gui, Yu-Xiong Wang, Yiming Yang, et~al. 2023.
\newblock Aligning large multimodal models with factually augmented rlhf.
\newblock \emph{arXiv preprint arXiv:2309.14525}.

\bibitem[{Tan et~al.(2024)Tan, Liu, and Jiang}]{tan2024personalized}
Zhaoxuan Tan, Zheyuan Liu, and Meng Jiang. 2024.
\newblock Personalized pieces: Efficient personalized large language models through collaborative efforts.
\newblock \emph{arXiv preprint arXiv:2406.10471}.

\bibitem[{Thudi et~al.(2022)Thudi, Deza, Chandrasekaran, and Papernot}]{thudi2022unrolling}
Anvith Thudi, Gabriel Deza, Varun Chandrasekaran, and Nicolas Papernot. 2022.
\newblock Unrolling sgd: Understanding factors influencing machine unlearning.
\newblock In \emph{EuroS\&P}.

\bibitem[{Touvron et~al.(2023)Touvron, Martin, Stone, Albert, Almahairi, Babaei, Bashlykov, Batra, Bhargava, Bhosale et~al.}]{touvron2023llama}
Hugo Touvron, Louis Martin, Kevin Stone, Peter Albert, Amjad Almahairi, Yasmine Babaei, Nikolay Bashlykov, Soumya Batra, Prajjwal Bhargava, Shruti Bhosale, et~al. 2023.
\newblock Llama 2: Open foundation and fine-tuned chat models.
\newblock \emph{arXiv preprint arXiv:2307.09288}.

\bibitem[{Varley(2023)}]{varley2023information}
Thomas~F Varley. 2023.
\newblock Information theory for complex systems scientists.
\newblock \emph{arXiv preprint arXiv:2304.12482}.

\bibitem[{Wang et~al.(2024{\natexlab{a}})Wang, Wu, He, Chen, and McAuley}]{wang2024large}
Yu~Wang, Ruihan Wu, Zexue He, Xiusi Chen, and Julian McAuley. 2024{\natexlab{a}}.
\newblock Large scale knowledge washing.
\newblock \emph{arXiv preprint arXiv:2405.16720}.

\bibitem[{Wang et~al.(2024{\natexlab{b}})Wang, Liu, Zhang, Ma, Zhang, and Ye}]{wang2024can}
Zehong Wang, Sidney Liu, Zheyuan Zhang, Tianyi Ma, Chuxu Zhang, and Yanfang Ye. 2024{\natexlab{b}}.
\newblock Can llms convert graphs to text-attributed graphs?
\newblock \emph{arXiv preprint arXiv:2412.10136}.

\bibitem[{Xie et~al.(2017)Xie, Lai, Dai, and Hovy}]{xie2017large}
Qizhe Xie, Guokun Lai, Zihang Dai, and Eduard Hovy. 2017.
\newblock Large-scale cloze test dataset created by teachers.
\newblock \emph{arXiv preprint arXiv:1711.03225}.

\bibitem[{Yao et~al.(2023)Yao, Xu, and Liu}]{yao2023large}
Yuanshun Yao, Xiaojun Xu, and Yang Liu. 2023.
\newblock Large language model unlearning.
\newblock \emph{arXiv preprint arXiv:2310.10683}.

\bibitem[{Ye et~al.(2023)Ye, Xu, Xu, Ye, Yan, Zhou, Wang, Hu, Shi, Shi et~al.}]{ye2023mplug}
Qinghao Ye, Haiyang Xu, Guohai Xu, Jiabo Ye, Ming Yan, Yiyang Zhou, Junyang Wang, Anwen Hu, Pengcheng Shi, Yaya Shi, et~al. 2023.
\newblock mplug-owl: Modularization empowers large language models with multimodality.
\newblock \emph{arXiv preprint arXiv:2304.14178}.

\bibitem[{Ye et~al.(2024)Ye, Xu, Ye, Yan, Hu, Liu, Qian, Zhang, and Huang}]{ye2024mplug}
Qinghao Ye, Haiyang Xu, Jiabo Ye, Ming Yan, Anwen Hu, Haowei Liu, Qi~Qian, Ji~Zhang, and Fei Huang. 2024.
\newblock mplug-owl2: Revolutionizing multi-modal large language model with modality collaboration.
\newblock In \emph{CVPR}.

\bibitem[{Yu et~al.(2024)Yu, Yao, Zhang, He, Han, Cui, Hu, Liu, Zheng, Sun et~al.}]{yu2024rlhf}
Tianyu Yu, Yuan Yao, Haoye Zhang, Taiwen He, Yifeng Han, Ganqu Cui, Jinyi Hu, Zhiyuan Liu, Hai-Tao Zheng, Maosong Sun, et~al. 2024.
\newblock Rlhf-v: Towards trustworthy mllms via behavior alignment from fine-grained correctional human feedback.
\newblock In \emph{CVPR}.

\bibitem[{Yu et~al.(2023)Yu, Yang, Li, Wang, Lin, Liu, Wang, and Wang}]{yu2023mm}
Weihao Yu, Zhengyuan Yang, Linjie Li, Jianfeng Wang, Kevin Lin, Zicheng Liu, Xinchao Wang, and Lijuan Wang. 2023.
\newblock Mm-vet: Evaluating large multimodal models for integrated capabilities.
\newblock \emph{arXiv preprint arXiv:2308.02490}.

\bibitem[{Yue et~al.(2024)Yue, Ni, Zhang, Zheng, Liu, Zhang, Stevens, Jiang, Ren, Sun et~al.}]{yue2024mmmu}
Xiang Yue, Yuansheng Ni, Kai Zhang, Tianyu Zheng, Ruoqi Liu, Ge~Zhang, Samuel Stevens, Dongfu Jiang, Weiming Ren, Yuxuan Sun, et~al. 2024.
\newblock Mmmu: A massive multi-discipline multimodal understanding and reasoning benchmark for expert agi.
\newblock In \emph{CVPR}.

\bibitem[{Zhang et~al.(2023)Zhang, Ippolito, Lee, Jagielski, Tram{\`e}r, and Carlini}]{zhang2023counterfactual}
Chiyuan Zhang, Daphne Ippolito, Katherine Lee, Matthew Jagielski, Florian Tram{\`e}r, and Nicholas Carlini. 2023.
\newblock Counterfactual memorization in neural language models.
\newblock \emph{Neurips}.

\bibitem[{Zhang et~al.(2025{\natexlab{a}})Zhang, Jian, Ouyang, and Vosoughi}]{zhang2025pretrained}
Chunhui Zhang, Yiren Jian, Zhongyu Ouyang, and Soroush Vosoughi. 2025{\natexlab{a}}.
\newblock Pretrained image-text models are secretly video captioners.
\newblock In \emph{Annual Conference of the North American Chapter of the Association for Computational Linguistics}.

\bibitem[{Zhang et~al.(2025{\natexlab{b}})Zhang, Ouyang, Lee, Agarwal, Houlihan, Vosoughi, and Lo}]{zhang2025overcoming}
Chunhui Zhang, Zhongyu Ouyang, Kwonjoon Lee, Nakul Agarwal, Sean~Dae Houlihan, Soroush Vosoughi, and Shao-Yuan Lo. 2025{\natexlab{b}}.
\newblock Overcoming multi-step complexity in theory-of-mind reasoning: A scalable bayesian planner.
\newblock In \emph{Proceedings of the 42nd International Conference on Machine Learning}.
\newblock Spotlight.

\bibitem[{Zhang et~al.(2024{\natexlab{a}})Zhang, Lin, Bai, and Mei}]{zhang2024negative}
Ruiqi Zhang, Licong Lin, Yu~Bai, and Song Mei. 2024{\natexlab{a}}.
\newblock Negative preference optimization: From catastrophic collapse to effective unlearning.
\newblock \emph{arXiv preprint arXiv:2404.05868}.

\bibitem[{Zhang et~al.(2021)Zhang, Lin, Liu, Li, Sun, and Zhou}]{zhang2021moefication}
Zhengyan Zhang, Yankai Lin, Zhiyuan Liu, Peng Li, Maosong Sun, and Jie Zhou. 2021.
\newblock Moefication: Transformer feed-forward layers are mixtures of experts.
\newblock \emph{arXiv preprint arXiv:2110.01786}.

\bibitem[{Zhang et~al.(2024{\natexlab{b}})Zhang, Wang, Ma, Taneja, Nelson, Le, Murugesan, Ju, Chawla, Zhang et~al.}]{zhang2024mopi}
Zheyuan Zhang, Zehong Wang, Tianyi Ma, Varun~Sameer Taneja, Sofia Nelson, Nhi Ha~Lan Le, Keerthiram Murugesan, Mingxuan Ju, Nitesh~V Chawla, Chuxu Zhang, et~al. 2024{\natexlab{b}}.
\newblock Mopi-hfrs: A multi-objective personalized health-aware food recommendation system with llm-enhanced interpretation.
\newblock \emph{arXiv preprint arXiv:2412.08847}.

\bibitem[{Zheng et~al.(2023)Zheng, Chiang, Sheng, Zhuang, Wu, Zhuang, Lin, Li, Li, Xing et~al.}]{zheng2023judging}
Lianmin Zheng, Wei-Lin Chiang, Ying Sheng, Siyuan Zhuang, Zhanghao Wu, Yonghao Zhuang, Zi~Lin, Zhuohan Li, Dacheng Li, Eric Xing, et~al. 2023.
\newblock Judging llm-as-a-judge with mt-bench and chatbot arena.
\newblock \emph{Neurips}.

\bibitem[{Zhu et~al.(2023)Zhu, Chen, Shen, Li, and Elhoseiny}]{zhu2023minigpt}
Deyao Zhu, Jun Chen, Xiaoqian Shen, Xiang Li, and Mohamed Elhoseiny. 2023.
\newblock Minigpt-4: Enhancing vision-language understanding with advanced large language models.
\newblock \emph{arXiv preprint arXiv:2304.10592}.

\end{thebibliography}

\newpage
\appendix

\section{Appendix: Important Function Design}
\label{appendix:neuron_activation_analysis}

\subsection{Neuron Act. Distribution ($I_{\text{abs}}$, $I_{\text{freq}}$)}
\label{appendix:neuron_activation_distribution}

\begin{figure*}
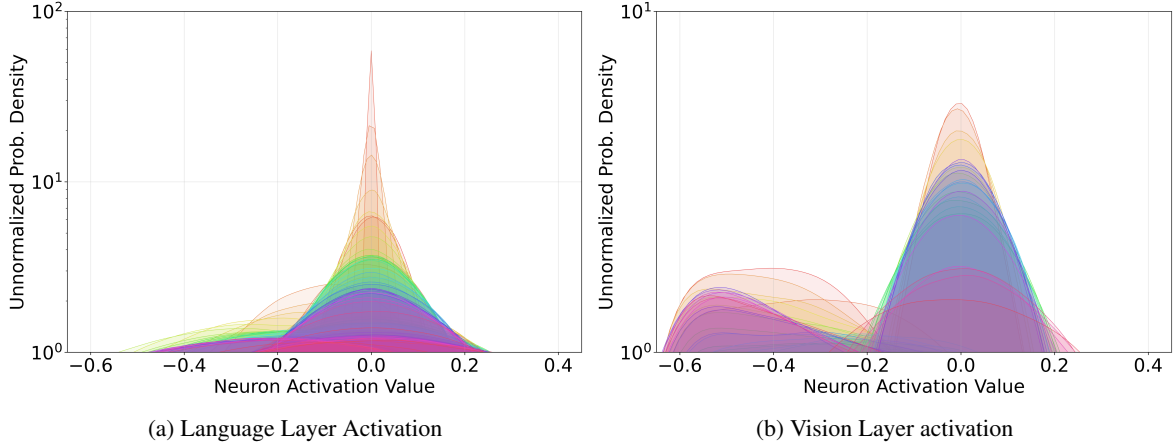

\centering
\begin{subfigure}{\columnwidth}
    \includegraphics[width=\columnwidth]{Figure/neuron_activation/language_layer_activations.pdf}
    \subcaption{Language Layer Activation}
    \label{fig:language_layer_act}
\end{subfigure}    
\begin{subfigure}{\columnwidth}
    \includegraphics[width=\columnwidth]{Figure/neuron_activation/vision_layer_activations.pdf}
    \subcaption{Vision Layer activation}
    \label{fig:vision_layer_act}
\end{subfigure}
\vspace{-0.1in}
\caption{Visualization of neuron activations across language MLP layers and vision MLP layers of MLLM. Figure \ref{fig:language_layer_act} shows neuron activations of language layers, while Figure \ref{fig:vision_layer_act} illustrates neuron activation patterns of vision layers. The $x$ axis represents neuron activation value, the $y$ axis shows the unnormalized probability density.}
\label{fig:neuron_activation_appendix}
\vspace{-0.2in}
\end{figure*}

In Figure \ref{fig:neuron_activation_appendix}, we present examples of neuron activation distributions for both the language and vision modules, respectively. As shown in the figure, the majority (not all) of pre-activation neurons exhibit a default activation of 0.0. This observation further reinforces the motivation behind our first importance function \( I_{\text{abs}} \), which leverages this sparsity pattern to quantify the extent to which activations deviate from zero. By capturing the magnitude of deviation, \( I_{\text{abs}} \) allows us to identify neurons that are more actively engaged in processing modality-specific information, distinguishing them from those that remain inactive across inputs.

Additionally, we observe a significant spike around zero (Figure \ref{fig:neuron_activation_appendix}), which aligns with the findings of \citet{zhang2021moefication}, emphasizing that meaningful nonzero activations occur only in select cases where neurons contribute to specific information processing tasks. This further validates the rationale behind our second importance function \( I_{\text{freq}} \) and underscores the necessity of capturing activation frequency when identifying neurons that are crucial for processing the target dataset.

\subsection{Information Diversity in Neural Act. ($I_{\text{var}}$)}
One key insight from information theory is that systems carrying more meaningful information exhibit diverse activation patterns rather than consistently remaining near zero. This principle is particularly relevant to our design of variance importance ($I_{\text{var}}$), which quantifies the spread of neuron activation values between modalities. Inspired by information theory principles \cite{varley2023information}, $I_{\text{var}}$ is formulated to capture the degree of information differentiation across modalities—a higher variance in activations implies stronger modality-specific processing, while a lower variance suggests redundancy or shared information. This metric allows us to identify neurons that contribute distinctively to multimodal versus unimodal inputs, ensuring that pruning decisions target modality-specific information rather than broadly removing neurons with minimal impact.

By leveraging variance as a measure of information richness, our approach aligns with information theory’s emphasis on quantifying uncertainty and diversity in signal representations, ultimately leading to a more effective and principled method for unlearning within MLLMs.

\subsection{Contextual Sparsity ($I_{\text{rms}}$)}
Recent studies have demonstrated that a substantial portion of neurons and attention heads in LLMs remain inactive or contribute minimally to output generation, highlighting the presence of significant redundancy within model activations. The work of \citet{liu2023deja} formally establishes this by introducing the concept of contextual sparsity, which leverages the observation that only a small, input-dependent subset of parameters is necessary to approximate the full model’s output effectively. Empirical findings in \citet{liu2023deja} reveal that up to 85\% of MLP neurons can be pruned dynamically at inference time without substantial degradation in model performance. These results strongly indicate that a large fraction of parameters within LLMs are redundant across different inputs. Building on these findings, we extend the notion of contextual sparsity to modality-aware unlearning, where redundant neurons may persist across different input types without contributing to modality-specific knowledge. This motivates our design of Root Mean Square Importance ($I_{\text{rms}}$), which quantifies neurons with consistently high yet uninformative activations. By identifying and pruning such neurons, we ensure that unlearning targets modality-relevant parameters while preserving overall model utility.

\section{Appendix: MLLMU-Bench}
\label{appendix:mllmu-bench}
\subsection{Benchmark Overview}

Our experimental results and observations are primarily based on MLLMU-Bench \cite{liu2024protecting}, which aims to advance the understanding of multimodal machine unlearning. We selected MLLMU-Bench for its comprehensive evaluation across various modalities and tasks. Specifically, it includes 500 fictitious profiles and 153 public celebrity profiles, each featuring over 14 customized question-answer pairs, assessed in both multimodal and unimodal settings. From a multimodal perspective, both the image and associated textual information of each individual's profile are provided, whereas the unimodal setting relies solely on textual information. Inspired by \cite{liu2024machine}, the benchmark is divided into four subsets: \textbf{Forget Set}, \textbf{Test Set}, \textbf{Retain Set}, and \textbf{Real Celebrity Set}, designed to evaluate unlearning algorithms in terms of efficacy, generalizability, and model utility. For each of these properties, MLLMU-Bench evaluates model performance on classification, generation, and cloze tasks under the aforementioned multimodal and unimodal settings. Detailed statistics about the benchmark are provided in Table \ref{tab:dataset_statistics}.

\subsection{Unlearning Efficacy}
The \textbf{Forget Set} is designed to evaluate unlearning efficacy of algorithms. Specifically, it is created by randomly selecting 5\%, 10\%, and 15\% of the 500 profiles, with each selected profile serving as an unlearning target. The primary goal of this dataset is to test the algorithm's ability to erase target knowledge while ensuring no residual traces of it remain.

\subsection{Unlearning Generalizability}

The \textbf{Test Set} is designed to evaluate the unlearning generalizability of algorithms. It is derived from the Forget Set by transforming both image and text data. For images, MLLMU-Bench uses Arc2Face \cite{papantoniou2024arc2face} to modify profile images with different poses and angles. For text, it employs GPT-4o to paraphrase questions into varied expressions. These transformations aim to assess whether the model has truly unlearned the target knowledge or still retains its transformed versions.

\subsection{Model Utility}

Lastly, the \textbf{Retain Set} and \textbf{Real Celebrity Set} are designed to evaluate model utility from different perspectives. The Retain Set consists of the remaining 95\%, 90\%, or 85\% of profiles, excluding the Forget Set. After the unlearning process, the model is expected to maintain high-fidelity knowledge of these profiles. The \textbf{Real Celebrity Set} serves as a control set to measure unintended interference with general pre-trained knowledge after unlearning. Like the other sets, it includes both multimodal (image and text) and text-only formats of real public figures.

\begin{table}[!t]
\centering
\small
    \begin{tabular}{lr}
        \toprule
        \textbf{Statistics}                     & \multicolumn{1}{r}{\textbf{Number}} \\ \midrule
        Total Questions                & 20,754                  \\
        \hspace{1em}* Image + Text Questions      & 10,377             \\
        \hspace{1em}* Pure Text Questions      & 10,377          \\
       
        Total Images & 1,153 \\ 
        \midrule

        Forget Percentile & 5\%/10\%/15\% \\\midrule
        
        Multiple-choice Questions      & 11,530            \\
        Free Generation Questions                 & 4,612 \\  
        Fill-in-the-blank Questions & 4,612\\
        \midrule
         
        Total Profiles               & 653               \\
        \hspace{1em}* Fictitious      & 500             \\
        \hspace{1em}* Real Celeb         & 153             \\
        
        Total Countries               & 70               \\
        Total Regions & 240 \\ 
        Total Birth Years & 211 \\ 
        Total Employement & 145 \\ 
        \bottomrule
    \end{tabular}
    \caption{Key statistics of the MLLMU-Bench.}
    \label{tab:dataset_statistics}
     \vspace{-0.25in}
\end{table}

\subsection{Evaluation Metrics}
As mentioned in the previous section, the post-unlearned model is evaluated on classification, generation, and cloze tasks across both multimodal and unimodal settings for each of these properties.

\subsubsection{Classification Task}
The classification task is designed around key attributes of each profile (e.g., education, occupation) by generating multiple-choice questions about personal details. In particular, the model is passed with \( \langle \text{image}, x, y \rangle \), where \( \text{image} \) represents the visual input in the multimodal setting (not applicable in the unimodal setting), \( x \) is the question, and \( y \) is the correct answer. The model then predicts \( \hat{y} \) based on the input $x$, and accuracy is calculated by comparing \( \hat{y} \) with the correct answer $y$.

\subsubsection{Generation Task}
In addition to classification, MLLMU-Bench evaluates the generation capabilities of post-unlearned models using free-generation questions. Each question is tailored to an individual’s profile, with GPT-4o generating answers based on the key attributes extracted from the profile. MLLMU-Bench employs the \textbf{ROUGE-L score} and \textbf{Factuality Score} for evaluation. Specifically, the ROUGE-L score \cite{lin2004rouge} measures the overlap of the longest matching subsequences between generated and reference texts. Next, inspired by prior benchmarks \cite{sun2023aligning, yu2024rlhf, zheng2023judging}, the Factuality Score assesses the factual accuracy and quality of generated responses using GPT-4o as the evaluator. It is rated on a scale of 1 to 10, where 1 represents an inaccurate response, and 10 signifies a fully correct and factually consistent answer.

\subsubsection{Cloze Test}
Lastly, inspired by previous Cloze-style tasks for evaluating models' memorization abilities \cite{xie2017large, duarte2024cop, carlini2021extracting, joshi2024towards}, MLLMU-Bench incorporates a Cloze-style task to assess whether sensitive information remains in the model after unlearning. Specifically, MLLMU-Bench provides only the individual's name as publicly available information, replacing all other key attributes with a \textit{[Blank]}. The model is then prompted to complete the missing information. This task aims to evaluate the model's unlearning capability regarding target knowledge when only partial context about the individual is revealed.

\section{Rationale for Targeting MLP Layers}
\label{appendix:rationale_mlp}

Recent research has demonstrated that MLP layers serve as primary knowledge storage components in transformer architectures. For example, \cite{huang2024commonsense} introduces the concept of "knowledge neurons," highlighting that specific neurons within MLP layers are responsible for encoding and storing information. By manipulating these neurons, it is possible to edit or selectively remove knowledge, offering fine-grained control over the model’s retained information. Beyond individual neurons, broader findings in knowledge editing literature reinforce the significance of MLP layers for model knowledge control. Prior works have shown that knowledge manipulation techniques, including direct parameter modification and knowledge attribution methods, consistently identify MLP layers as the primary repository of factual and task-specific knowledge \cite{wang2024large, meng2022locating, meng2022mass}. Given that vision transformers share a fundamentally similar architecture with language transformers, where MLPs play an analogous role in feature extraction and information processing \cite{ghiasi2022vision}, we extend this insight to the vision tower as well. The effectiveness of targeting MLPs for knowledge controlling is further supported by recent work in LLM pruning \cite{pochinkov2024dissecting}, which demonstrates that modifying MLP layers enables precise control over model knowledge while maintaining core model capabilities.

\section{Appendix: Implementation Details}

\subsection{Baseline Methods}
\label{appendix:baselines}
\subsubsection{Gradient Ascent}
The Gradient Ascent approach \cite{thudi2022unrolling} is a simple yet effective method for unlearning. The primary goal of GA is to increase the loss for samples in the forget set \( \mathcal{D}_f \), thereby minimizing the likelihood of the model retaining specific information about these profiles. In particular, for each sample \( x \in \mathcal{D}_f \), GA aims to maximize the loss, driving the model away from its original predictions. The objective is to maximize the average loss across the $\mathcal{D}_f$:
\[
\mathcal{L}(\mathcal{D}_f, w) = \frac{1}{|\mathcal{D}_f|} \sum_{x \in \mathcal{D}_f} \ell(x, w),
\]
where \( \ell(x, w) \) denotes the loss for a sample \( x \) with model parameters \( w \). This process encourages the model to unlearn the associations it formed during fine-tuning with respect to the forget set.

\subsubsection{Gradient Difference}
Gradient Difference \cite{liu2022continual} builds upon Gradient Ascent by balancing the unlearning of the forget set with the preservation of performance on the retain set \( \mathcal{D}_r \). The objective is to increase the loss on \( \mathcal{D}_f \) while minimizing the impact on \( \mathcal{D}_r \). This method ensures that the model forgets the targeted data without negatively affecting unrelated knowledge. The overall loss function is defined as:
\[
\mathcal{L}_{\text{diff}} = -\mathcal{L}(\mathcal{D}_f, w) + \mathcal{L}(\mathcal{D}_r, w),
\]
where \( \mathcal{L}(\mathcal{D}_r, w) \) is the loss computed on the retain set and $w$ indicates the model parameters. By optimizing this combined loss, the model selectively forgets the specified profiles while retaining performance on the rest of the dataset.

\subsubsection{KL Minimization}
The KL Minimization method \cite{nguyen2020variational} aims to align the model’s predictions on the retain set with those of the original fine-tuned model while encouraging divergence on the forget set. Specifically, it minimizes the Kullback-Leibler (KL) divergence between the outputs of the current model and the original model for samples in \( \mathcal{D}_r \), ensuring that important knowledge is retained. Simultaneously, the conventional loss is maximized on \( \mathcal{D}_f \). Formally, the objective is:
\[
\mathcal{L}_{\text{KL}} = -\mathcal{L}(\mathcal{D}_f, w) + \frac{1}{|\mathcal{D}_r|} \sum_{s \in \mathcal{D}_r} \text{KL}(M_{\text{o}} \| M_{\text{c}})(s)
\]
where \( M_{\text{o}} \) and \( M_{\text{c}} \) represent the \textit{original} and \textit{current} models, respectively. This method ensures that unlearning is targeted while the model’s behavior on the retain set remains unchanged.

\subsubsection{Generic Prevention using prompt}
To demonstrate the applicability of system prompts in unlearning scenarios, we append a system prompt to the unlearned model during evaluation as follows:
\begin{quote}
"You are a helpful, respectful, and honest assistant. When generating your response, please do not generate any personal-related information."
\end{quote}
This provides a concise instruction that supplements the default system prompt, explicitly instructing the model not to generate any privacy-related content.

\subsubsection{Negative Preference Optimization}
Negative Preference Optimization (NPO) technique aims to address the issue of catastrophic collapse that often associated with gradient ascent methods. NPO \cite{zhang2024negative} is inspired by preference-based learning \cite{rafailov2024direct, ouyang2022training, bai2022training}, where it operates within the preference optimization framework, targeting negative samples from the \( \mathcal{D}_f \). In particular, the NPO loss function is defined as follows:
\[
    \mathcal{L}_{\text{NPO}} = \frac{2}{\beta} \mathbb{E}_{(x, y) \in D_{\text{f}}} \left[ \log \left(1 + \left(\frac{\pi_\theta(y|x)}{\pi_{\text{ref}}(y|x)}\right)^\beta \right) \right]
\]
where \( \pi_\theta(y|x) \) represents the prediction probability of the current model for token \( y \) given the input \( x \), and \( \pi_{\text{ref}}(y|x) \) is the prediction probability from the reference model trained on the entire dataset. The parameter \( \beta \) controls the smoothness of the optimization, and as \( \beta \to 0 \), the NPO loss converges to the standard gradient ascent loss. By minimizing this loss, NPO decreases the model's dependence on the \( \mathcal{D}_f \), thereby promoting a more stable unlearning process while preventing the rapid degradation commonly observed with gradient ascent methods. In our experiments, we set \( \beta = 0.9 \), following the default setting from the original paper and MLLMU-Bench. Then, we define \( \pi_{\text{ref}} \) by fine-tuning the pre-trained model solely on the \( \mathcal{D}_r \).

\subsection{Hyperparameters Settings}
\label{appendix:hyperparameter_settings}
Here, we present the hyperparameter settings for \method using LLaVA and Idefics2 as the base models. Since the pruning process does not involve gradient updates, the primary tunable parameter is the batch size, which we set to 4. All experiments are conducted on NVIDIA A6000 GPUs (48 GB).

\section{Appendix: Additional Experiments}
\label{appendix:additional_experiments}

\subsection{Main Experiments (Idefics2)}
\label{appendix:main_experiments-idefics2}
In this section, we present additional experiments on MLLMU-Bench using Idefics2 as the base model, with results shown in Table \ref{tab:main-table-idefics}. The trends align with Table \ref{tab:main-table-llava} for LLaVA, where \method outperforms baselines across all datasets and tasks, consistently ranking first or runner-up.

\begin{table*}[t!]
    \centering
\scalebox{0.51}{
\begin{tabular}{l|cccc|cccc|cccc|cccc}
        \toprule
        \multirow{3}{*}{\textbf{Models}} 
        & \multicolumn{4}{c|}{\textbf{Forget Set}} 
        & \multicolumn{4}{c|}{\textbf{Test Set}} 
        & \multicolumn{4}{c|}{\textbf{Retain Set}} 
        & \multicolumn{4}{c}{\textbf{Real Celebrity}} \\
        \cline{2-17}
        & \begin{tabular}[c]{@{}c@{}}Class.\\ Acc (\textcolor{blue}{$\downarrow$})\end{tabular} 
        & \begin{tabular}[c]{@{}c@{}}Rouge\\ Score (\textcolor{red}{$\downarrow$})\end{tabular} 
        & \begin{tabular}[c]{@{}c@{}}Fact.\\ Score (\textcolor{red}{$\downarrow$})\end{tabular} 
        & \begin{tabular}[c]{@{}c@{}}Cloze\\ Acc (\textcolor{teal}{$\downarrow$})\end{tabular} 
        & \begin{tabular}[c]{@{}c@{}}Class.\\ Acc (\textcolor{blue}{$\downarrow$})\end{tabular} 
        & \begin{tabular}[c]{@{}c@{}}Rouge\\ Score (\textcolor{red}{$\downarrow$})\end{tabular} 
        & \begin{tabular}[c]{@{}c@{}}Fact.\\ Score (\textcolor{red}{$\downarrow$})\end{tabular} 
        & \begin{tabular}[c]{@{}c@{}}Cloze\\ Acc (\textcolor{teal}{$\downarrow$})\end{tabular} 
        & \begin{tabular}[c]{@{}c@{}}Class.\\ Acc (\textcolor{blue}{$\uparrow$})\end{tabular} 
        & \begin{tabular}[c]{@{}c@{}}Rouge\\ Score (\textcolor{red}{$\uparrow$})\end{tabular} 
        & \begin{tabular}[c]{@{}c@{}}Fact.\\ Score (\textcolor{red}{$\uparrow$})\end{tabular} 
        & \begin{tabular}[c]{@{}c@{}}Cloze\\ Acc (\textcolor{teal}{$\uparrow$})\end{tabular} 
        & \begin{tabular}[c]{@{}c@{}}Class.\\ Acc (\textcolor{blue}{$\uparrow$})\end{tabular} 
        & \begin{tabular}[c]{@{}c@{}}Rouge\\ Score (\textcolor{red}{$\uparrow$})\end{tabular} 
        & \begin{tabular}[c]{@{}c@{}}Fact.\\ Score (\textcolor{red}{$\uparrow$})\end{tabular} 
        & \begin{tabular}[c]{@{}c@{}}Cloze\\ Acc (\textcolor{teal}{$\uparrow$})\end{tabular} \\
        \midrule

        \multicolumn{17}{c}{\textbf{Idefics-2-8B (5\% Forget)}} \\
        \midrule
        Vanilla & 53.80\% & 0.630 & 6.22 & 44.75\% & 47.86\% & 0.434 & 5.00 & 24.97\% & 46.11\% & 0.644 & 6.51 & 42.35\% & 52.75\% & 0.459 & 5.75 & 20.05\% \\
        
        GA & \textbf{36.27\%} & \textbf{0.405} & \textbf{2.90} & \textbf{30.07\%} & \underline{38.40\%} & \underline{0.374} & \textbf{3.42} & \underline{21.44\%} & 39.09\% & 0.410 & 3.81 & 28.01\% & 41.27\% & 0.202 & 2.62 & 15.07\% \\
        
        Grad. Diff. & 40.38\% & 0.426 & 3.96 & 32.24\% & 41.41\% & 0.408 & \underline{3.73} & 22.66\% & 40.07\% & 0.408 & 4.05 & 33.19\% & 43.52\% & 0.363 & 3.91 & 16.37\% \\

        KL Minimization & 39.69\% & 0.459 & 3.39 & 36.79\% & 45.20\% & 0.419 & 4.24 & 23.32\% & 38.83\% & 0.393 & 3.76 & 39.82\% & 45.64\% & 0.360 & 3.27 & 17.74\% \\
        
        Prompting & 45.45\% & 0.492 & 3.91 & 42.61\% & 44.87\% & 0.423 & 4.39 & 23.88\% & \textbf{44.99\%} & \textbf{0.601} & \textbf{5.02} &\textbf{42.05\%} & \textbf{52.00\%} & \textbf{0.427} & \textbf{4.88} & \textbf{19.95\%} \\
        
        NPO & 43.29\% & 0.501 & 4.87 & 39.77\% & 41.98\% & 0.391 & 4.47 & 22.75\% & 41.19\% & 0.484 & 4.57 & 39.99\% & 50.05\% & 0.384 & 4.05 & 18.17\% \\

        \rowcolor{gray!12}\method & \underline{37.13\%} &  \underline{0.413} & \underline{3.20} & \underline{32.11\%} & \textbf{35.55\%} & \textbf{0.361} & 3.91 & \textbf{20.97\%} & \underline{42.71\%} & \underline{0.538} & \underline{4.60} &\underline{40.01\%} & \underline{50.09\%} & \underline{0.399} & \underline{4.11} & \underline{18.80\%}  \\

        \midrule

        \multicolumn{17}{c}{\textbf{Idefics-2-8B (10\% Forget)}} \\
        \midrule
        Vanilla & 54.48\% & 0.645 & 6.27 & 46.55\% & 48.09\% & 0.492 & 5.36 & 27.81\% & 47.52\% & 0.643 & 6.63 & 43.37\% & 52.75\% & 0.459 & 5.75 & 20.05\% \\
        
        GA & 37.81\% & \textbf{0.459} & \textbf{3.09} & \underline{31.05\%} & \textbf{38.17\%} & \textbf{0.313} & \textbf{3.64} & \textbf{20.43\%} & 38.15\% & 0.494 & 4.56 & 33.58\% & 42.16\% & 0.250 & 2.75 & 15.88\% \\
        
        Grad. Diff. & \textbf{36.60\%} & 0.471 & \underline{3.33} & 35.57\% & 40.22\% & 0.414 & \underline{3.68} & 24.65\% & 36.82\% & 0.461 & 4.34 & 35.80\% & 41.52\% & 0.386 & 3.62 & 17.72\% \\
        
        KL Minimization & 41.28\% & 0.524 & 3.71 & 43.34\% & 42.74\% & 0.491 & 3.75 & 25.00\% & 38.10\% & 0.499 & 4.33 & 39.53\% & 43.64\% & 0.395 & 3.42 & 18.58\% \\
        
        Prompting & 46.40\% & 0.504 & 3.55 & 45.27\% & 45.10\% & 0.422 & 4.09 & 26.31\% & \textbf{44.31\%} & \textbf{0.634} & \textbf{5.06} & \textbf{43.27\%} & \textbf{52.00\%} & \textbf{0.458} & \textbf{4.90} & \textbf{20.05\%} \\
        
        NPO & 42.91\% & 0.521 & 4.12 & 41.44\% & 41.09\% & 0.399 & 3.77 & 23.11\% & 42.39\% & \underline{0.541} & 4.82 & 40.02\% & 48.76\% & 0.421 & 3.91 & 17.39\% \\

        \rowcolor{gray!12}\method & \underline{37.01\%} & \underline{0.467} & 3.45 & \textbf{29.98\%} & \underline{39.75\%} & \underline{0.355} & \textbf{3.51} & \underline{21.88\%} & \underline{43.07\%} & 0.539 & \underline{4.94} & \underline{41.13\%} & \underline{49.31\%} & \underline{0.437} & \underline{3.98} & \underline{19.10\%} \\

        \midrule
        \multicolumn{17}{c}{\textbf{Idefics-2-8B (15\% Forget)}} \\
        \midrule
        Vanilla & 54.67\% & 0.630 & 6.42 & 46.33\% & 47.99\% & 0.436 & 5.30 & 27.77\% & 46.86\% & 0.645 & 6.48 & 42.81\% & 52.75\% & 0.459 & 5.75 & 20.05\% \\
        
        GA & 37.87\% & \textbf{0.335} & 3.23 & \underline{31.11\%} & 37.90\% & 0.342 & 3.20 & \textbf{15.67\%} & 38.66\% & 0.444 & 3.06 & 28.95\% & 43.56\% & 0.341 & 2.42 & 13.92\% \\
        
        Grad. Diff. & \textbf{35.33\%} & \underline{0.340} & \textbf{3.01} & 33.50\% & \underline{36.41\%} & \textbf{0.310} & \textbf{2.99} & 18.59\% & 36.07\% & 0.370 & 3.19 & 35.00\% & 45.52\% & 0.408 & 3.03 & 15.88\% \\
        
        KL Minimization & 41.09\% & 0.521 & 4.03 & 42.76\% & 44.81\% & 0.428 & 3.94 & 23.67\% & 39.54\% & 0.491 & 3.35 & 40.80\% & 47.64\% & 0.419 & 3.79 & 17.72\% \\
        
        Prompting & 45.73\% & 0.482 & 3.88 & 45.23\% & 45.66\% & 0.409 & 3.72 & 26.16\% & 43.01\% & \textbf{0.606} & \underline{5.03} & \textbf{42.27\%} & \textbf{52.00\%} & \textbf{0.459} & \textbf{4.88} & \textbf{19.93\%} \\
        
        NPO & 41.44\% & 0.447 & 3.97 & 40.06\% & 38.75\% & 0.389 & 3.49 & 22.10\% & \underline{43.23\%} & \underline{0.597} & \textbf{5.17} & 40.19\% & 48.99\% & 0.424 & 4.07 & 18.88\% \\

        \rowcolor{gray!12}\method & \underline{36.49\%} & 0.366 & \underline{3.17} & \textbf{30.67\%} & \textbf{35.32\%} & \underline{0.333} & \underline{3.10} & \underline{16.66\%} & \textbf{43.95\%} & 0.566 & 4.98 & \underline{41.61\%} & \underline{50.81\%} & \underline{0.430} & \underline{4.22} & \underline{19.26\%} \\
                
        \bottomrule
    \end{tabular}}
    \vspace{-0.1in}
    \caption{Overall average performance of baseline methods and \method on Idefics2, combining multimodal and unimodal evaluations across three forget setups. \textbf{Bold} indicates the best performance, and \underline{underline} denotes the runner-up. Each method is evaluated on four datasets from MLLMU-Bench, assessed by classification accuracy, ROUGE-L score, factuality score and cloze accuracy. We abbreviate the Factuality Score as Fact. Score due to space limits. \textcolor{blue}{$\sbullet[.75]$}, \textcolor{red}{$\sbullet[.75]$}, and \textcolor{teal}{$\sbullet[.75]$} represent classification, generation and cloze evaluations, respectively. $\downarrow$ indicates that lower values are better, while $\uparrow$ indicates that higher values are better.}
    \label{tab:main-table-idefics}
    \vspace{-0.1in}
\end{table*}

\subsection{Pruning Ratio Analysis:}
\label{appendix:prune_ratio}
In this section, we present additional analyses on the influence of different pruning ratios on unlearning effectiveness and model utility, as shown in Tables \ref{tab:prune-ratio-5-15}. As observed in these tables, regardless of the split ratio for the forget set, the trend remains consistent with the findings in Table \ref{tab:prune-ratio-10}. Specifically, as the pruning ratio increases, unlearning performance improves, but model utility deteriorates. These experimental results further validate the phenomenon that larger pruning ratios can disrupt the balance between effective unlearning and model utility.
\begin{table*}[t!]
    \centering
\scalebox{0.51}{
\begin{tabular}{l|cccc|cccc|cccc|cccc}
        \toprule
        \multirow{3}{*}{\textbf{Models}} 
        & \multicolumn{4}{c|}{\textbf{Forget Set}} 
        & \multicolumn{4}{c|}{\textbf{Test Set}} 
        & \multicolumn{4}{c|}{\textbf{Retain Set}} 
        & \multicolumn{4}{c}{\textbf{Real Celebrity}} \\
        \cline{2-17}
        & \begin{tabular}[c]{@{}c@{}}Class.\\ Acc (\textcolor{blue}{$\downarrow$})\end{tabular} 
        & \begin{tabular}[c]{@{}c@{}}Rouge\\ Score (\textcolor{red}{$\downarrow$})\end{tabular} 
        & \begin{tabular}[c]{@{}c@{}}Fact.\\ Score (\textcolor{red}{$\downarrow$})\end{tabular} 
        & \begin{tabular}[c]{@{}c@{}}Cloze\\ Acc (\textcolor{teal}{$\downarrow$})\end{tabular} 
        & \begin{tabular}[c]{@{}c@{}}Class.\\ Acc (\textcolor{blue}{$\downarrow$})\end{tabular} 
        & \begin{tabular}[c]{@{}c@{}}Rouge\\ Score (\textcolor{red}{$\downarrow$})\end{tabular} 
        & \begin{tabular}[c]{@{}c@{}}Fact.\\ Score (\textcolor{red}{$\downarrow$})\end{tabular} 
        & \begin{tabular}[c]{@{}c@{}}Cloze\\ Acc (\textcolor{teal}{$\downarrow$})\end{tabular} 
        & \begin{tabular}[c]{@{}c@{}}Class.\\ Acc (\textcolor{blue}{$\uparrow$})\end{tabular} 
        & \begin{tabular}[c]{@{}c@{}}Rouge\\ Score (\textcolor{red}{$\uparrow$})\end{tabular} 
        & \begin{tabular}[c]{@{}c@{}}Fact.\\ Score (\textcolor{red}{$\uparrow$})\end{tabular} 
        & \begin{tabular}[c]{@{}c@{}}Cloze\\ Acc (\textcolor{teal}{$\uparrow$})\end{tabular} 
        & \begin{tabular}[c]{@{}c@{}}Class.\\ Acc (\textcolor{blue}{$\uparrow$})\end{tabular} 
        & \begin{tabular}[c]{@{}c@{}}Rouge\\ Score (\textcolor{red}{$\uparrow$})\end{tabular} 
        & \begin{tabular}[c]{@{}c@{}}Fact.\\ Score (\textcolor{red}{$\uparrow$})\end{tabular} 
        & \begin{tabular}[c]{@{}c@{}}Cloze\\ Acc (\textcolor{teal}{$\uparrow$})\end{tabular} \\
        \midrule
        \multicolumn{17}{c}{\textbf{LLaVA-1.5-7B (5\% Forget)}} \\
        \midrule

        Vanilla & 51.70\% & 0.645 & 6.78 & 25.81\% & 47.86\% & 0.539 & 4.89 & 23.01\% & 46.11\% & 0.632 & 6.41 & 27.83\% & 51.80\% & 0.479 & 5.47 & 17.35\% \\
        
        \method (2\%) & 41.25\% & 0.491 & 3.27 & 17.08\% & 41.67\% & 0.334 & 3.81 & 15.78\% & 43.38\% & 0.542 & 4.45 & 24.08\% & 49.57\% & 0.448 & 4.67 & 16.01\% \\
        
        \method (5\%) & 36.25\% & 0.477 & 3.16 & 13.54\% & 39.17\% & 0.291 & 3.56 & 15.71\% & 38.25\% & 0.514 & 4.14 & 10.02\% & 49.09\% & 0.436 & 3.51 & 11.44\% \\
        
        \method (10\%) & 32.40\% & 0.473 & 3.01 & 11.25\% & 33.40\% & 0.262 & 3.35 & 13.25\% & 37.34\% & 0.476 & 3.97 & 11.62\% & 48.77\% & 0.409 & 3.47 & 11.11\% \\
                
        \midrule

        \multicolumn{17}{c}{\textbf{Idefics2-8B (5\% Forget)}} \\
        \midrule

        Vanilla & 53.80\% & 0.630 & 6.22 & 44.75\% & 47.86\% & 0.434 & 5.00 & 24.97\% & 46.11\% & 0.644 & 6.51 & 42.35\% & 52.75\% & 0.459 & 5.75 & 20.05\% \\
        
        \method (2\%) & 37.13\% & 0.413 & 3.20 & 32.11\% & 35.55\% & 0.361 & 3.91 & 20.97\% & 42.71\% & 0.538 & 4.60 & 40.01\% & 50.09\% & 0.399 & 4.11 & 18.80\%  \\
        
        \method (5\%) & 36.01\% & 0.339 & 3.13 & 30.18\% & 34.97\% & 0.343  &3.84 & 15.98\% & 41.54\% & 0.505 & 4.47 & 38.72\% & 48.61\% & 0.376 & 3.96 & 17.27\%\\
        
        \method (10\%) & 31.87\% & 0.307 & 3.07 & 28.04\% & 33.77\% & 0.321 &3.53  & 15.05\% & 39.80\% & 0.481 & 4.39 & 36.20\% & 46.66\% & 0.352 & 3.77 & 16.93\%\\

        \midrule
        \multicolumn{17}{c}{\textbf{LLaVA-1.5-7B (15\% Forget)}} \\
        \midrule

        Vanilla & 51.87\% & 0.575 & 6.34 & 26.62\% & 47.53\% & 0.502 & 4.08 &  25.33\% & 48.06\% & 0.585 & 5.46 & 28.51\% & 51.80\% & 0.479 & 5.47 & 17.35\% \\
        
        \method (2\%) & 42.05\% & 0.481 & 3.73 & 17.91\% & 41.75\% & 0.360 & 3.52 & 17.01\% & 46.86\% & 0.557 & 5.19 & 24.62\% & 50.42\% & 0.448 & 4.05 & 16.77\% \\
        
        \method (5\%) & 41.94\% & 0.477 & 3.24 & 17.26\% & 41.35\% & 0.357 & 3.43 & 16.89\% & 43.68\% & 0.509 & 5.19 & 17.27\% & 49.77\% & 0.451 & 3.50 & 12.41\% \\
        
        \method (10\%) & 35.14\% & 0.468 & 3.11 & 12.97\% & 38.81\% & 0.327 & 3.40 & 14.53\% & 41.76\% & 0.493 & 5.02 & 10.75\% & 48.01\% & 0.410 & 3.41 & 10.78\% \\
                
        \midrule

        \multicolumn{17}{c}{\textbf{Idefics2-8B (15\% Forget)}} \\
        \midrule

        Vanilla & 54.67\% & 0.630 & 6.42 & 46.33\% & 47.99\% & 0.436 & 5.30 & 27.77\% & 46.86\% & 0.645 & 6.48 & 42.81\% & 52.75\% & 0.459 & 5.75 & 20.05\% \\
        
        \method (2\%) & 36.49\% & 0.366 & 3.17 & 30.67\% &35.32\% & 0.333 & 3.10 & 16.66\% & 43.95\% & 0.566 & 4.98 & 41.61\% & 50.81\% & 0.430 & 4.22 & 19.26\% \\

        \method (5\%) & 33.82\% & 0.328 & 3.15 & 28.88\% & 34.44\% & 0.329 & 3.07 & 15.59\% & 41.11\% & 0.528 & 4.54 & 39.63\% & 49.93\% & 0.414 & 4.19 & 18.77\% \\
        
        \method (10\%) & 30.39\% & 0.301 & 3.13 & 25.93\% & 32.46\% & 0.315 & 3.01 & 15.20\% & 39.41\% & 0.509 & 4.44  & 38.01\% & 48.02\% & 0.403 & 4.07 & 18.01\% \\

        \bottomrule
    \end{tabular}}
    \vspace{-0.1in}
    \caption{Overall results of \method with varying pruning ratios on two base MLLM models under a 5\% and 15\% forget data setup. For each MLLM, the pruning ratio is iteratively increased from 2\% to 10\%. }
    \label{tab:prune-ratio-5-15}
    \vspace{-0.2in}
\end{table*}

\subsection{Unlearning across modalities }
\label{appendix:unlearn_across_modality}
\begin{figure*}[!t]
\centering
\begin{subfigure}[b]{\textwidth}
    \centering
    \includegraphics[width=0.65\textwidth]{Figure/llava_modality_analyze/llava_legend.pdf}
\end{subfigure}
\begin{subfigure}{0.244\textwidth}
    \includegraphics[width=\textwidth]{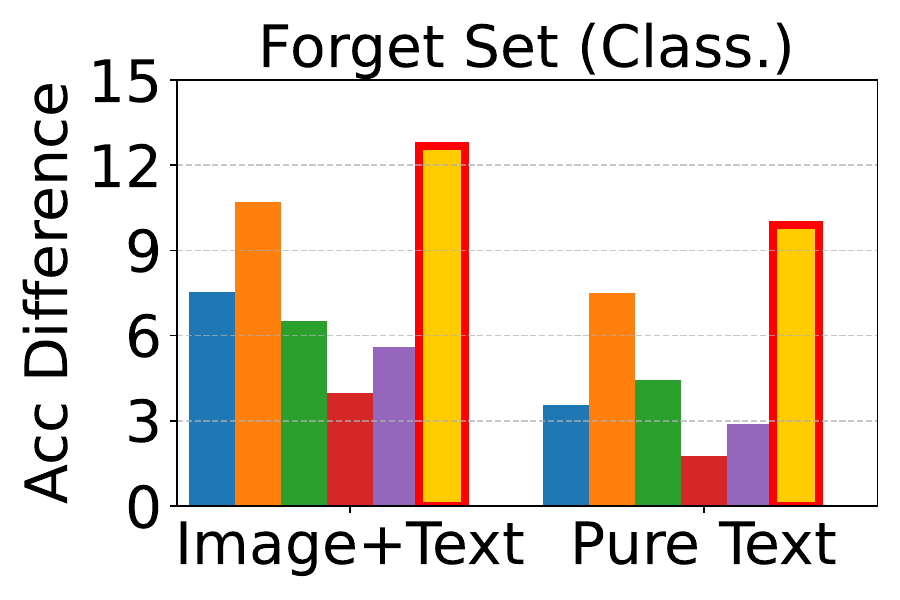}
    \subcaption{Forget Set (Classification)}
    \label{fig:llava_10_class_forget}
\end{subfigure}    
\begin{subfigure}{0.244\textwidth}
    \includegraphics[width=\textwidth]{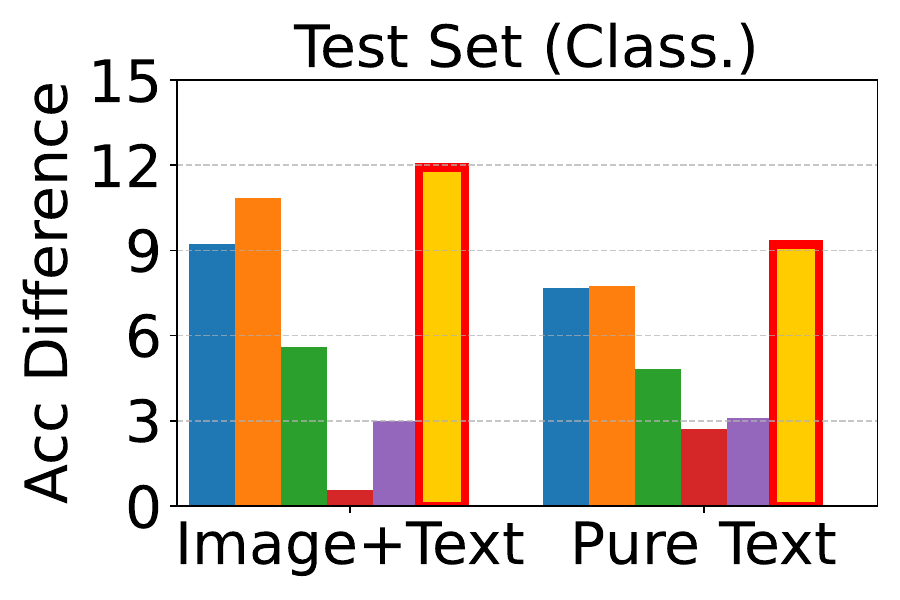}
    \subcaption{Test Set (Classification)}
    \label{fig:llava_10_class_test}
\end{subfigure}
\begin{subfigure}{0.244\textwidth}
    \includegraphics[width=\textwidth]{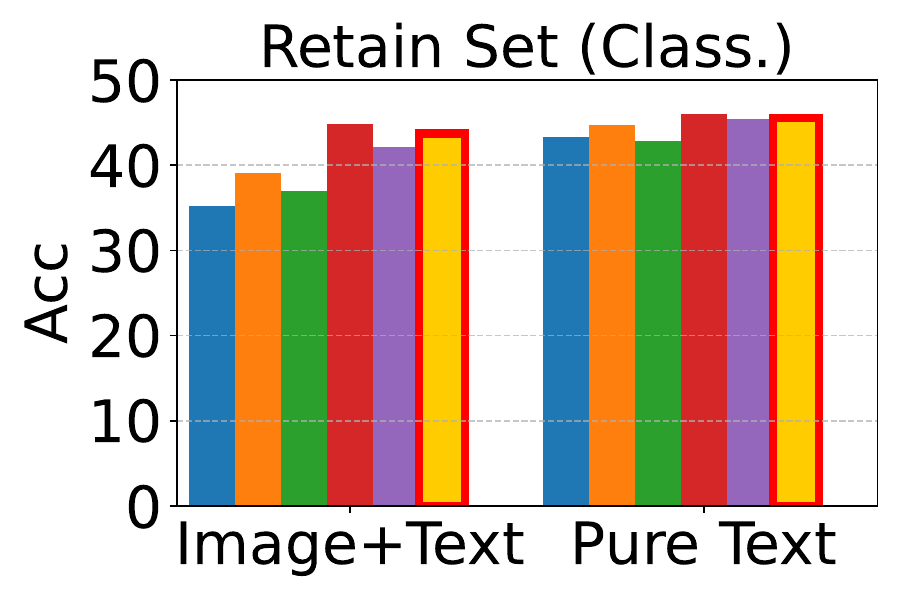}
    \subcaption{Retain Set (Classification)}
    \label{fig:llava_10_class_retain}
\end{subfigure}    
\begin{subfigure}{0.244\textwidth}
    \includegraphics[width=\textwidth]{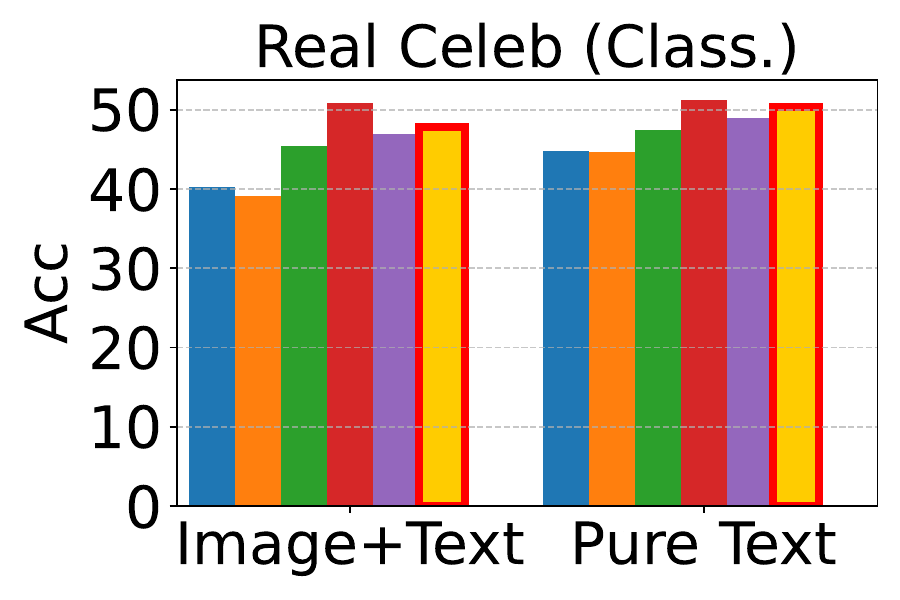} 
    \subcaption{Real Celeb (Classification)}
    \label{fig:llava_10_class_real}
\end{subfigure}


\begin{subfigure}{0.244\textwidth}
    \includegraphics[width=\textwidth]{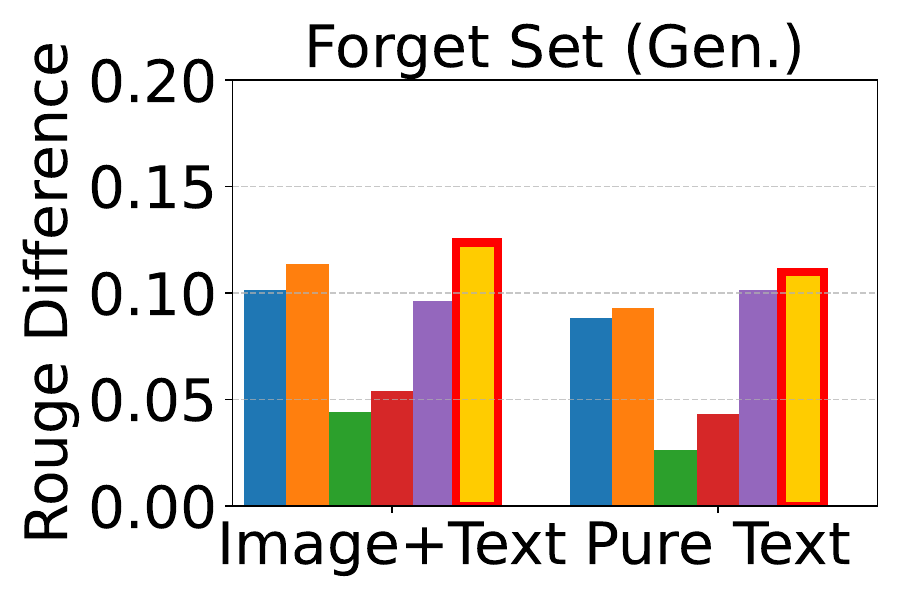}
    \subcaption{Forget Set (Generation)}
    \label{fig:llava_10_gen_forget}
\end{subfigure}
\begin{subfigure}{0.244\textwidth}
    \includegraphics[width=\textwidth]{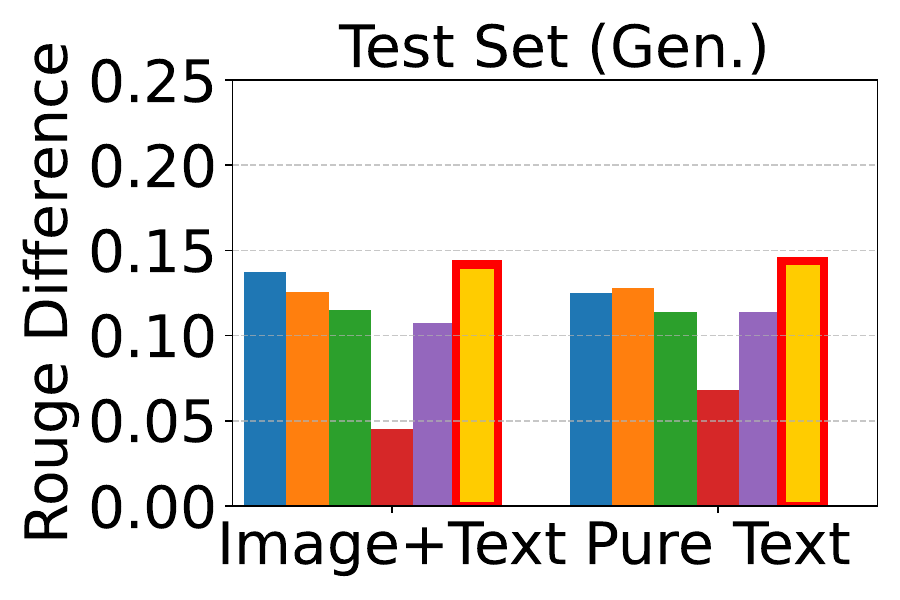}
    \subcaption{Test Set (Generation)}
    \label{fig:llava_10_gen_test}
\end{subfigure}
\begin{subfigure}{0.244\textwidth}
    \includegraphics[width=\textwidth]{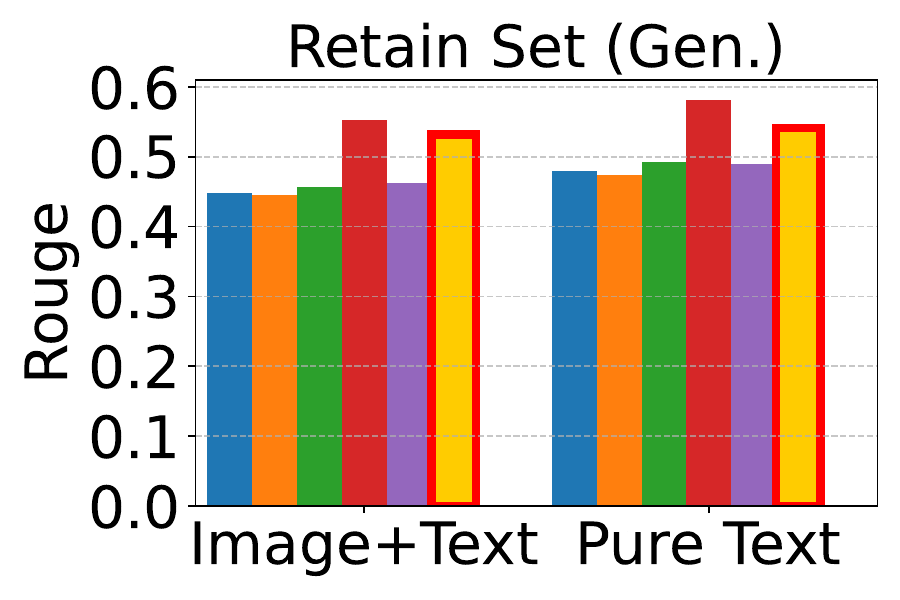}
    \subcaption{Retain Set (Generation)}
    \label{fig:llava_10_gen_retain}
\end{subfigure}
\begin{subfigure}{0.244\textwidth}
    \includegraphics[width=\textwidth]{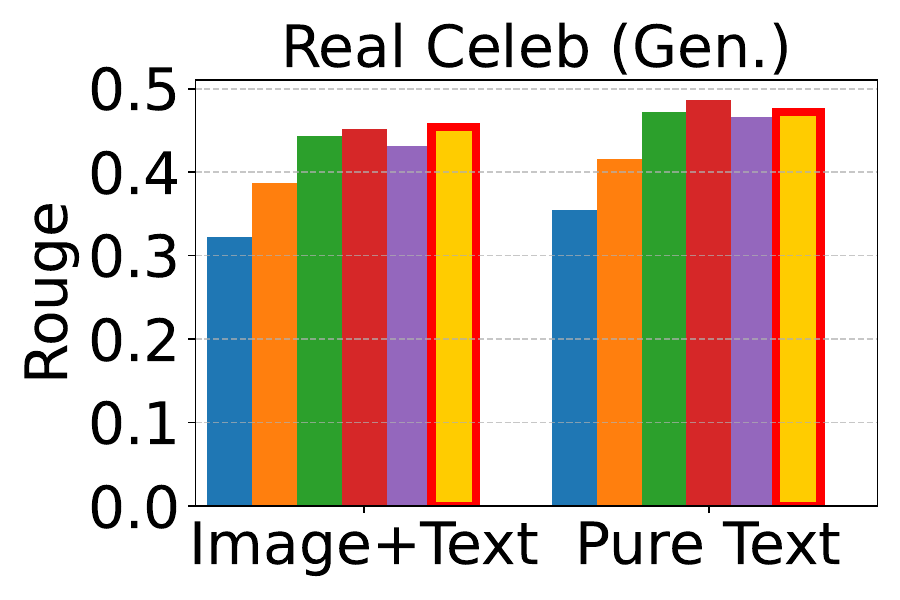}
    \subcaption{Real Celeb (Generation)}
    \label{fig:llava_10_gen_real}
\end{subfigure}

\begin{subfigure}{0.244\textwidth}
    \includegraphics[width=\textwidth]{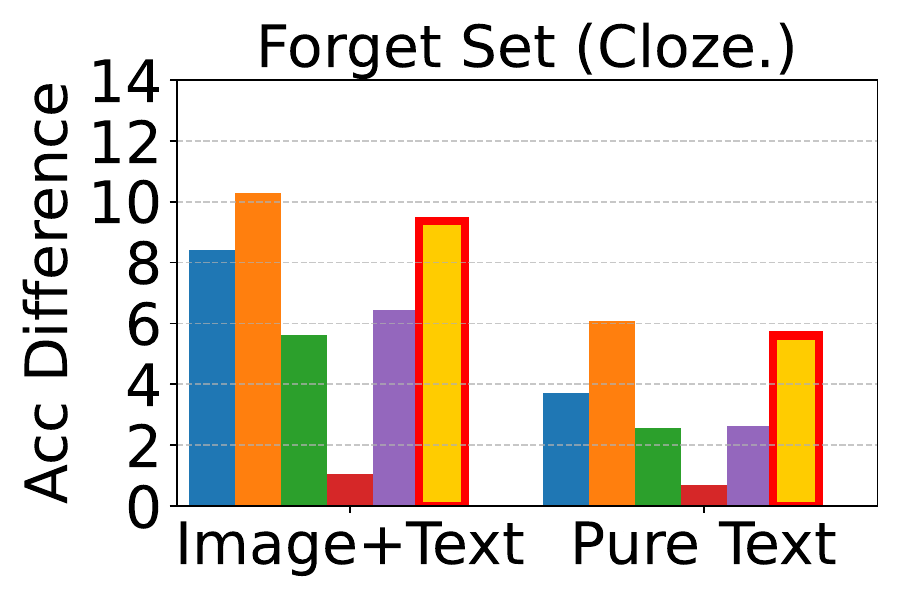}
    \subcaption{Forget Set (Cloze)}
    \label{fig:llava_10_cloze_forget}
\end{subfigure}
\begin{subfigure}{0.244\textwidth}
    \includegraphics[width=\textwidth]{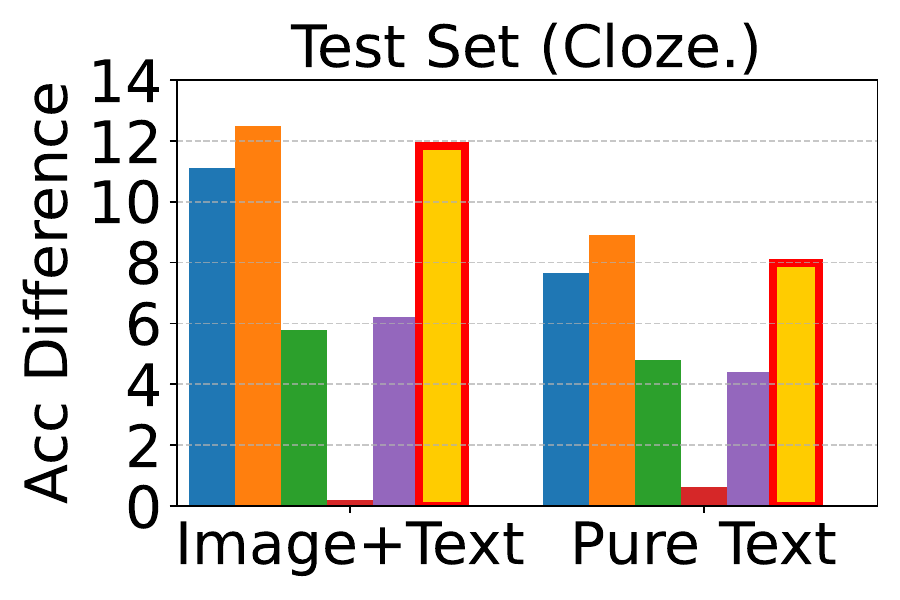}
    \subcaption{Test Set (Cloze)}
    \label{fig:llava_10_cloze_test}
\end{subfigure}
\begin{subfigure}{0.244\textwidth}
    \includegraphics[width=\textwidth]{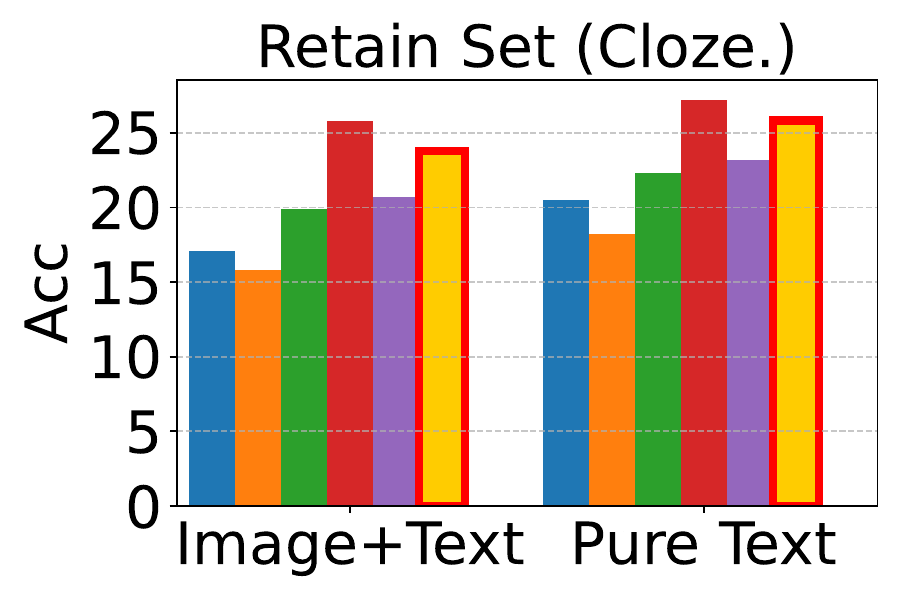}
    \subcaption{Retain Set (Cloze)}
    \label{fig:llava_10_cloze_retain}
\end{subfigure}
\begin{subfigure}{0.244\textwidth}
    \includegraphics[width=\textwidth]{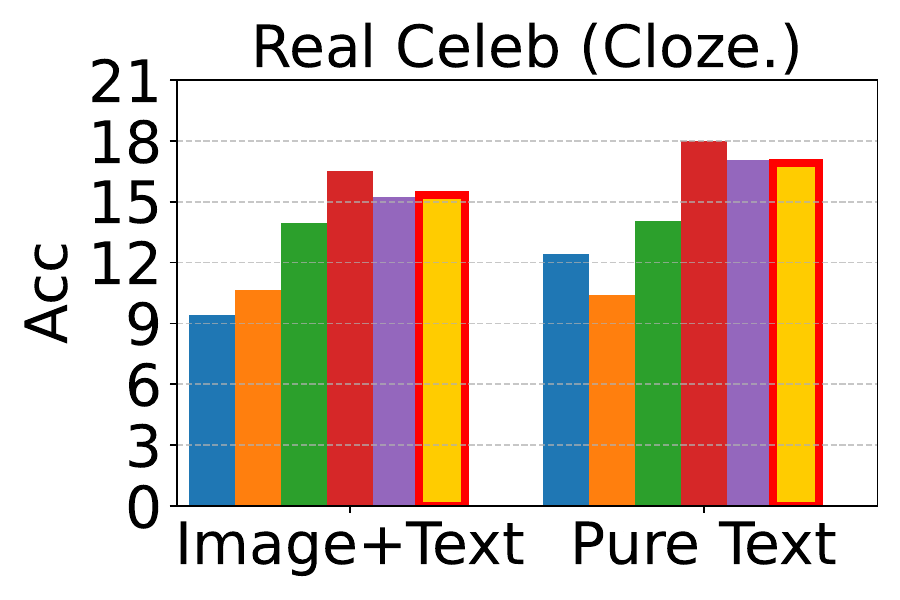}
    \subcaption{Real Celeb (Cloze)}
    \label{fig:llava_10_cloze_real}
\end{subfigure}
\vspace{-0.1in}
\caption{
Classification, generation, and cloze performance of \method and baselines in multimodal and unimodal setups with 10\% forget data, using LLaVA as the base model. In subplots (a), (b), (e), (f), (i), and (j), the $y$-axis represents the change in classification accuracy, ROUGE-L score, and cloze accuracy relative to the vanilla model, evaluated on the Forget and Test sets. In the remaining subplots, the $y$-axis indicates classification accuracy, ROUGE-L score, and cloze accuracy, respectively. The $x$-axis represents performance across different modalities.}
\vspace{-0.19in}
\label{fig:llava_10_compare}
\end{figure*}

\begin{figure*}[!t]
\centering
\begin{subfigure}[b]{\textwidth}
    \centering
    \includegraphics[width=0.65\textwidth]{Figure/llava_modality_analyze/llava_legend.pdf}
\end{subfigure}
\begin{subfigure}{0.244\textwidth}
    \includegraphics[width=\textwidth]{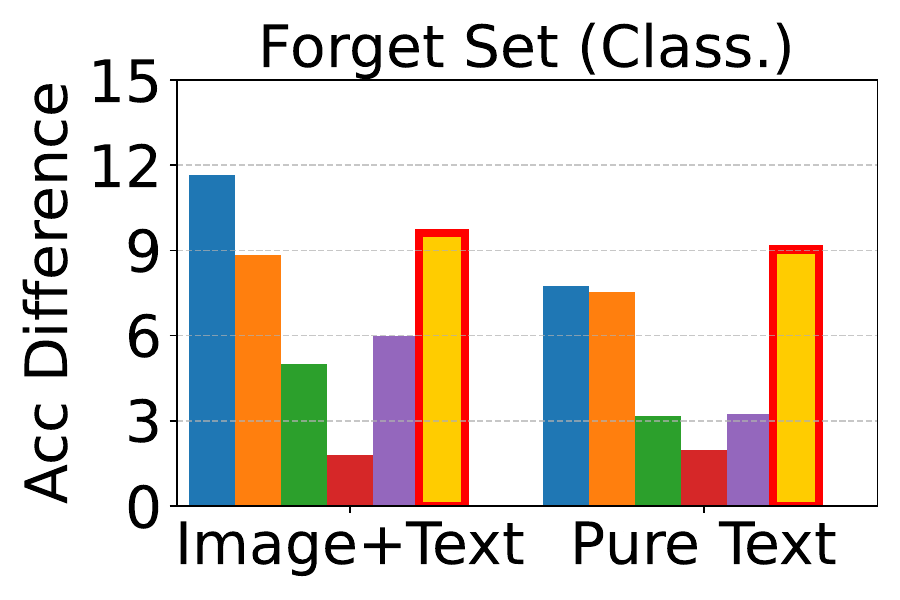}
    \subcaption{Forget Set (Classification)}
    \label{fig:llava_15_class_forget}
\end{subfigure}    
\begin{subfigure}{0.244\textwidth}
    \includegraphics[width=\textwidth]{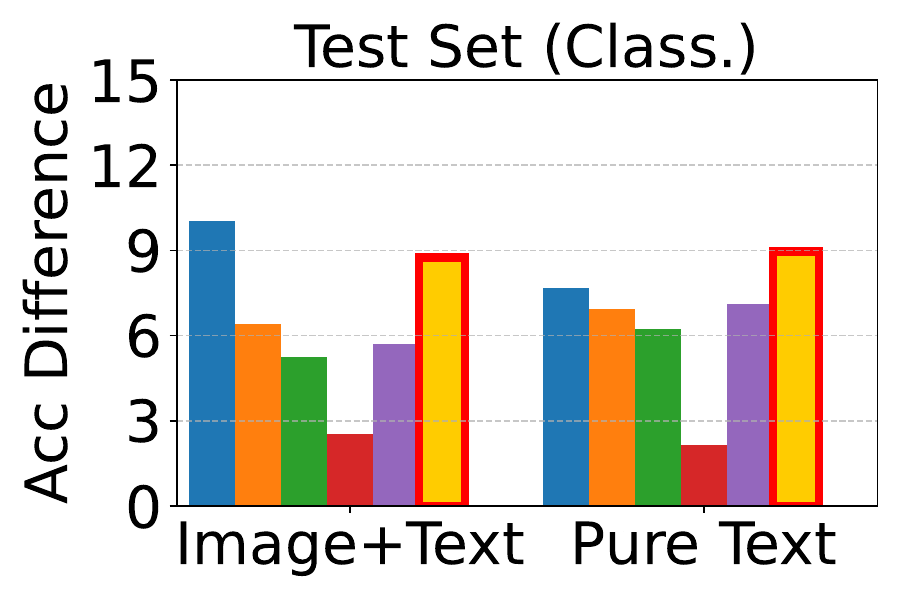}
    \subcaption{Test Set (Classification)}
    \label{fig:llava_15_class_test}
\end{subfigure}
\begin{subfigure}{0.244\textwidth}
    \includegraphics[width=\textwidth]{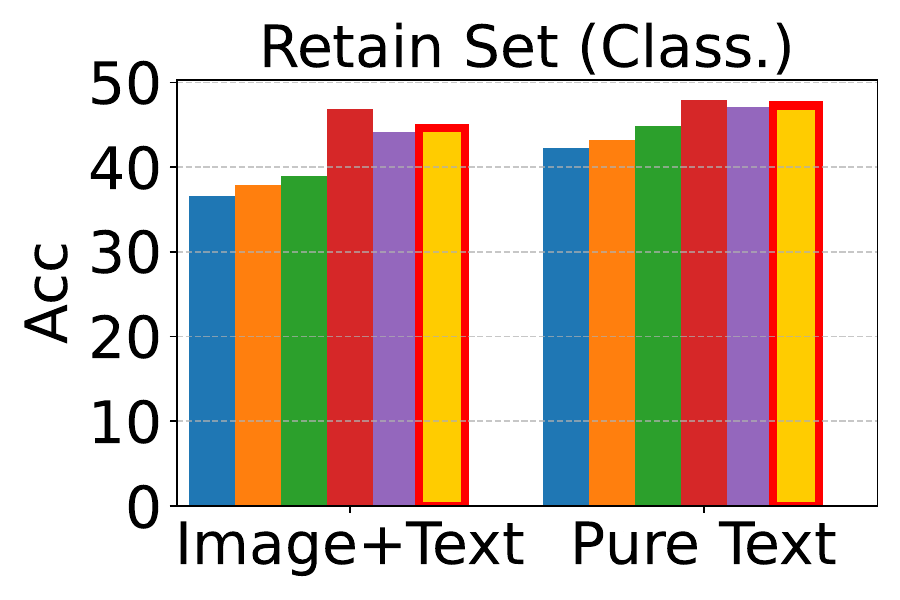}
    \subcaption{Retain Set (Classification)}
    \label{fig:llava_15_class_retain}
\end{subfigure}    
\begin{subfigure}{0.244\textwidth}
    \includegraphics[width=\textwidth]{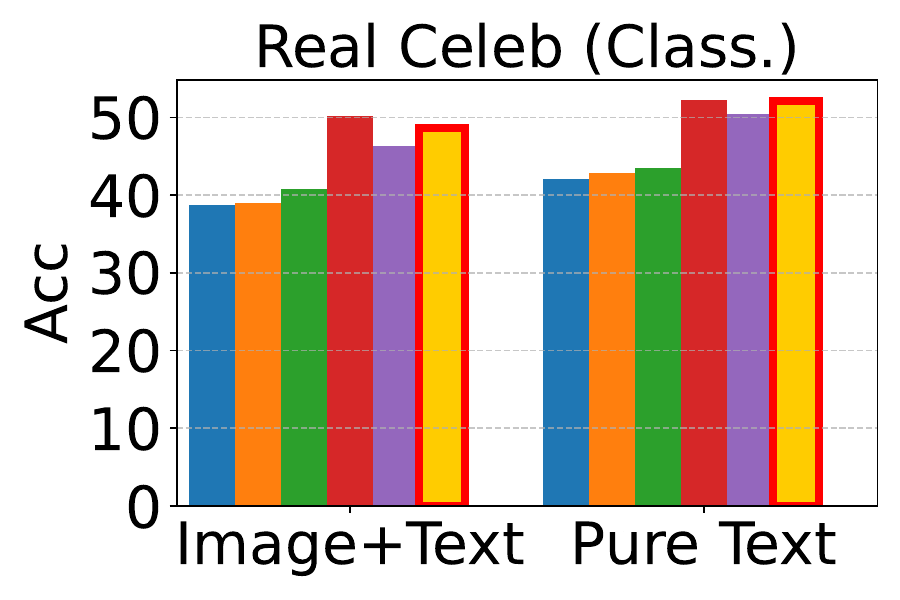} 
    \subcaption{Real Celeb (Classification)}
    \label{fig:llava_15_class_real}
\end{subfigure}


\begin{subfigure}{0.244\textwidth}
    \includegraphics[width=\textwidth]{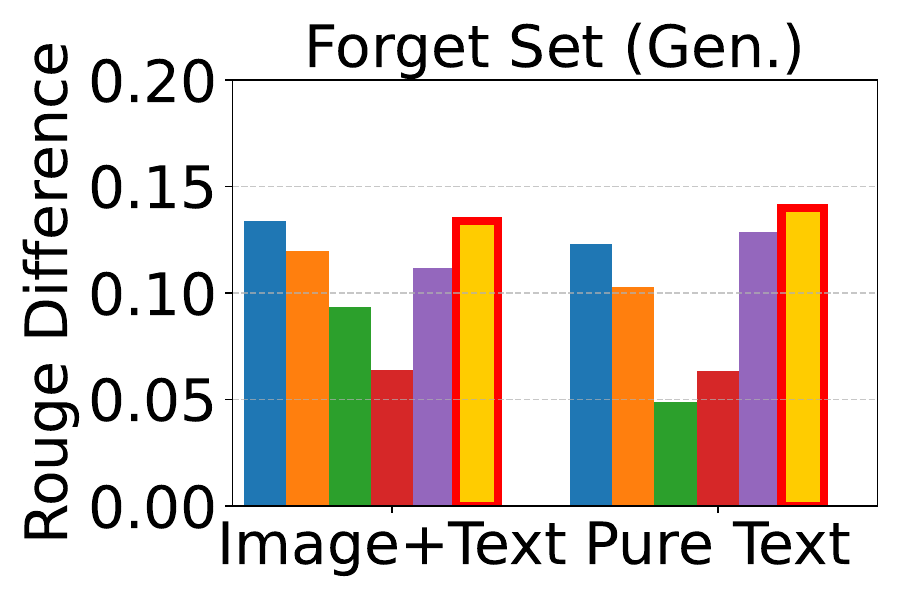}
    \subcaption{Forget Set (Generation)}
    \label{fig:llava_15_gen_forget}
\end{subfigure}
\begin{subfigure}{0.244\textwidth}
    \includegraphics[width=\textwidth]{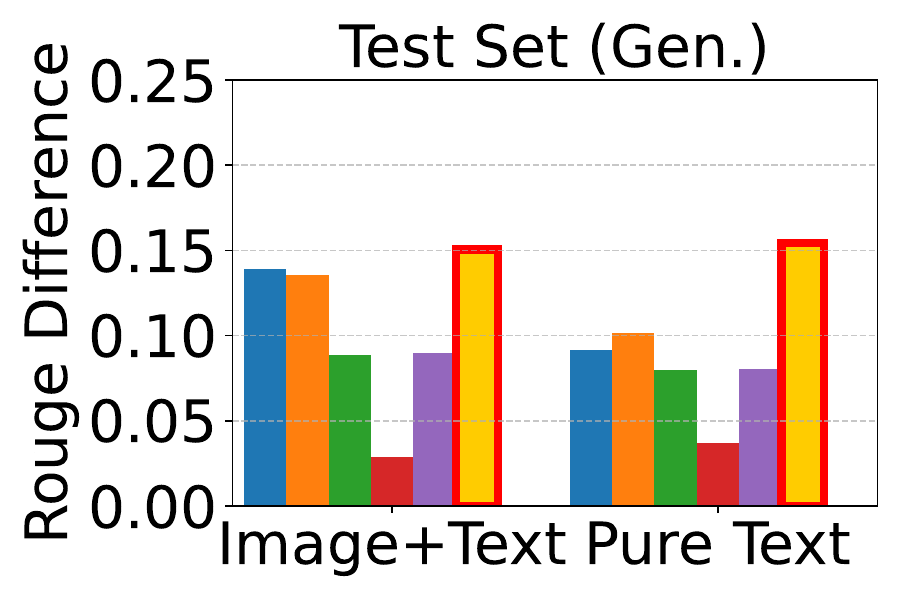}
    \subcaption{Test Set (Generation)}
    \label{fig:llava_15_gen_test}
\end{subfigure}
\begin{subfigure}{0.244\textwidth}
    \includegraphics[width=\textwidth]{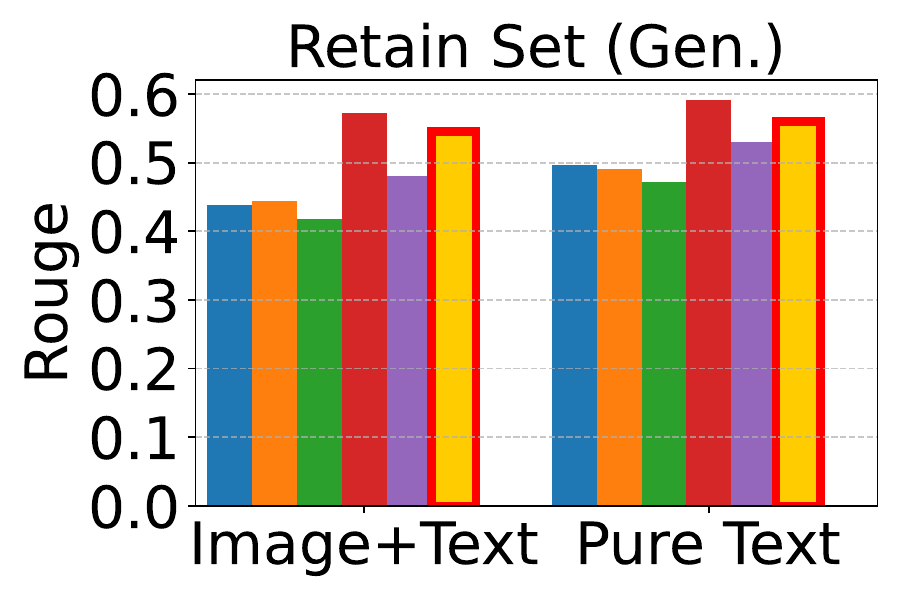}
    \subcaption{Retain Set (Generation)}
    \label{fig:llava_15_gen_retain}
\end{subfigure}
\begin{subfigure}{0.244\textwidth}
    \includegraphics[width=\textwidth]{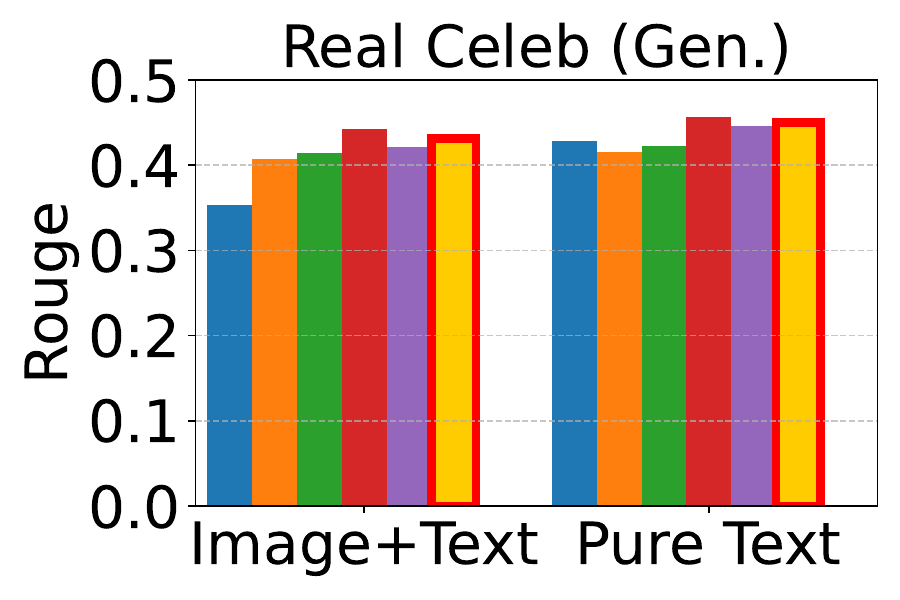}
    \subcaption{Real Celeb (Generation)}
    \label{fig:llava_15_gen_real}
\end{subfigure}

\begin{subfigure}{0.244\textwidth}
    \includegraphics[width=\textwidth]{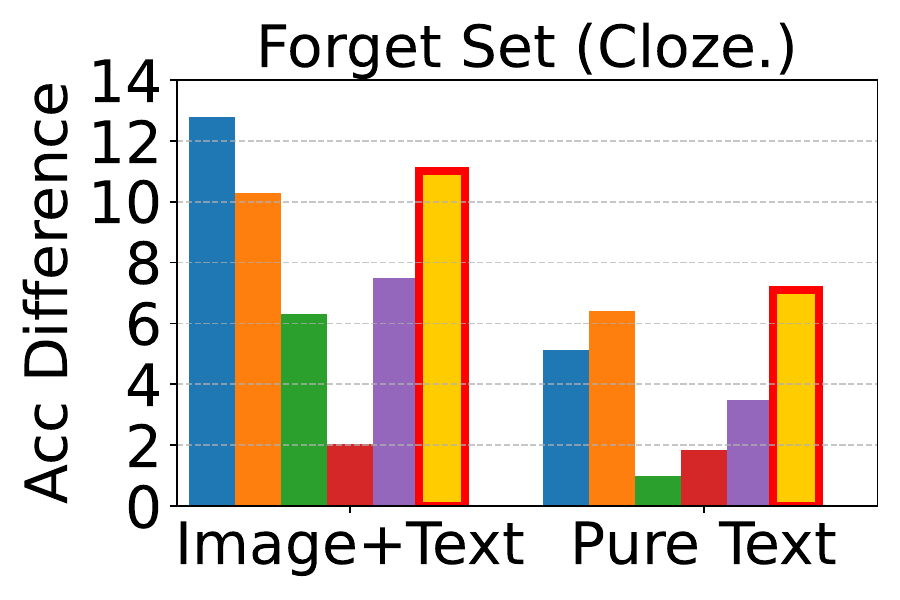}
    \subcaption{Forget Set (Cloze)}
    \label{fig:llava_15_cloze_forget}
\end{subfigure}
\begin{subfigure}{0.244\textwidth}
    \includegraphics[width=\textwidth]{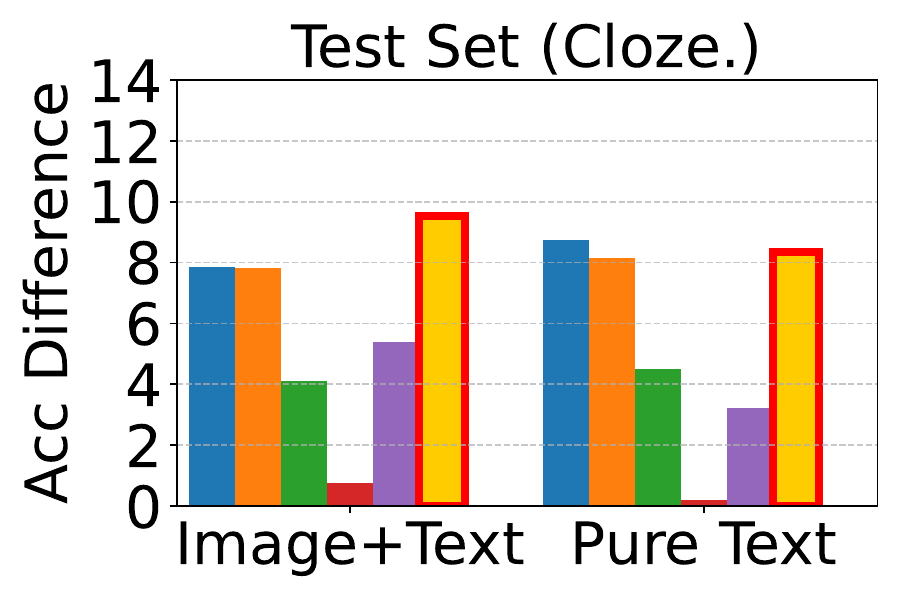}
    \subcaption{Test Set (Cloze)}
    \label{fig:llava_15_cloze_test}
\end{subfigure}
\begin{subfigure}{0.244\textwidth}
    \includegraphics[width=\textwidth]{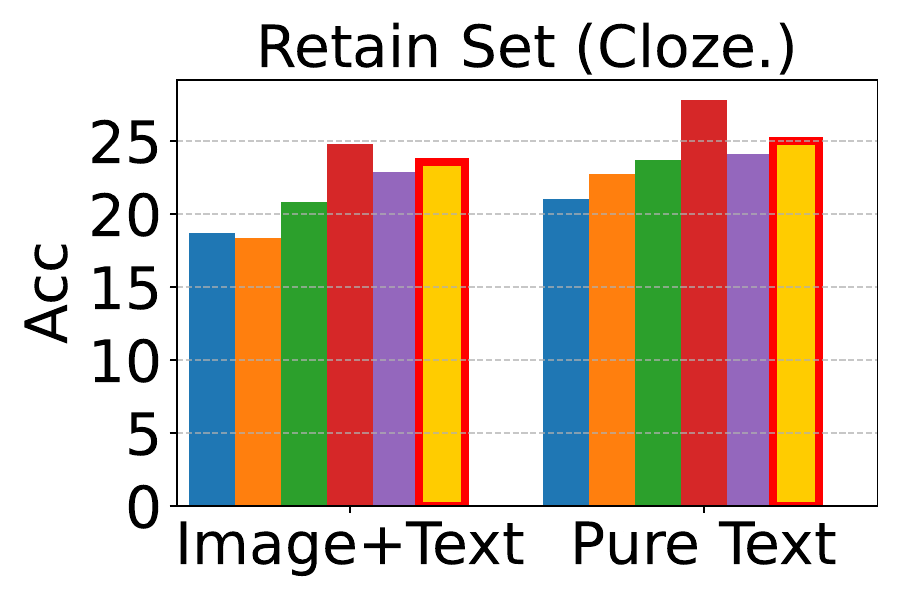}
    \subcaption{Retain Set (Cloze)}
    \label{fig:llava_15_cloze_retain}
\end{subfigure}
\begin{subfigure}{0.244\textwidth}
    \includegraphics[width=\textwidth]{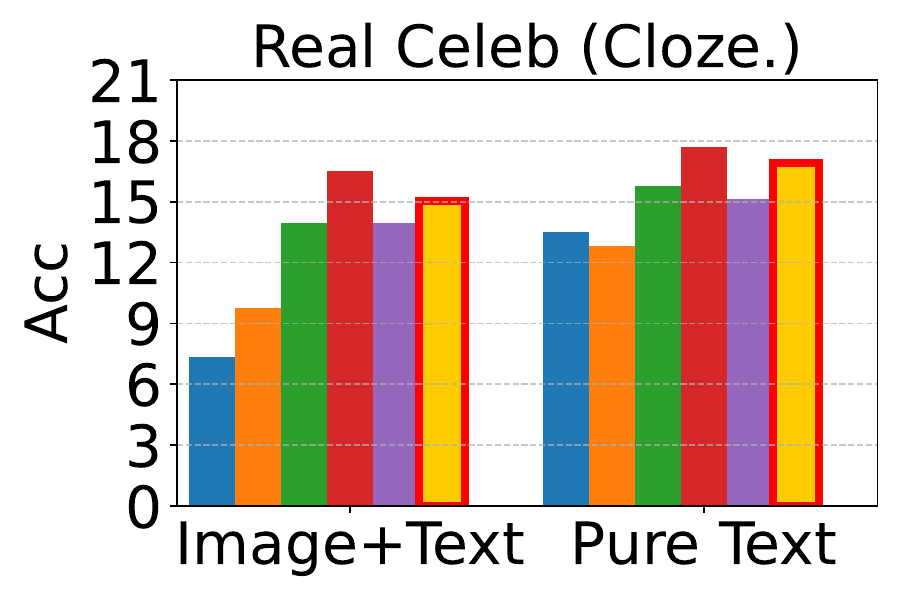}
    \subcaption{Real Celeb (Cloze)}
    \label{fig:llava_15_cloze_real}
\end{subfigure}
\vspace{-0.1in}
\caption{
Classification, generation, and cloze performance of \method and baselines in multimodal and unimodal setups with 15\% forget data, using LLaVA as the base model. In subplots (a), (b), (e), (f), (i), and (j), the $y$-axis represents the change in classification accuracy, ROUGE-L score, and cloze accuracy relative to the vanilla model, evaluated on the Forget and Test sets. In the remaining subplots, the $y$-axis indicates classification accuracy, ROUGE-L score, and cloze accuracy, respectively. The $x$-axis represents performance across different modalities.}
\vspace{-0.2in}
\label{fig:llava_15_compare}
\end{figure*}

Here, we present additional experiments evaluating the unlearning effectiveness of all tested algorithms using forget split ratios of 10\% (Figure \ref{fig:llava_10_compare}) and 15\% (Figure \ref{fig:llava_15_compare}), with Llava as the base model. These experiments aim to demonstrate that \method effectively addresses the unique challenge of incomplete unlearning across different input types in the context of MLLM unlearning. As shown in the figures, we observe a trend similar to that in Figure \ref{fig:llava_5_compare}. In particular, while some algorithms (e.g. GA based algorithms) perform well in multimodal evaluation, they often exhibit shortcomings in unimodal evaluation due to the absence of a curated modality-specific design. This further underscores the importance of modality-aware methodologies in MLLM unlearning.

\begin{figure*}[!t]
\centering
\begin{subfigure}[b]{\textwidth}
    \centering
    \includegraphics[width=0.7\textwidth]{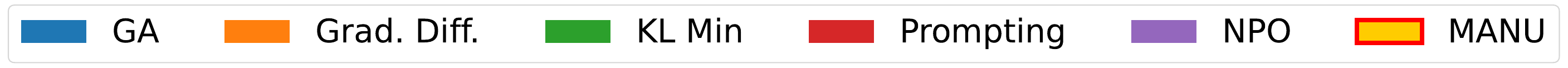}
\end{subfigure}
\begin{subfigure}{0.244\textwidth}
    \includegraphics[width=\textwidth]{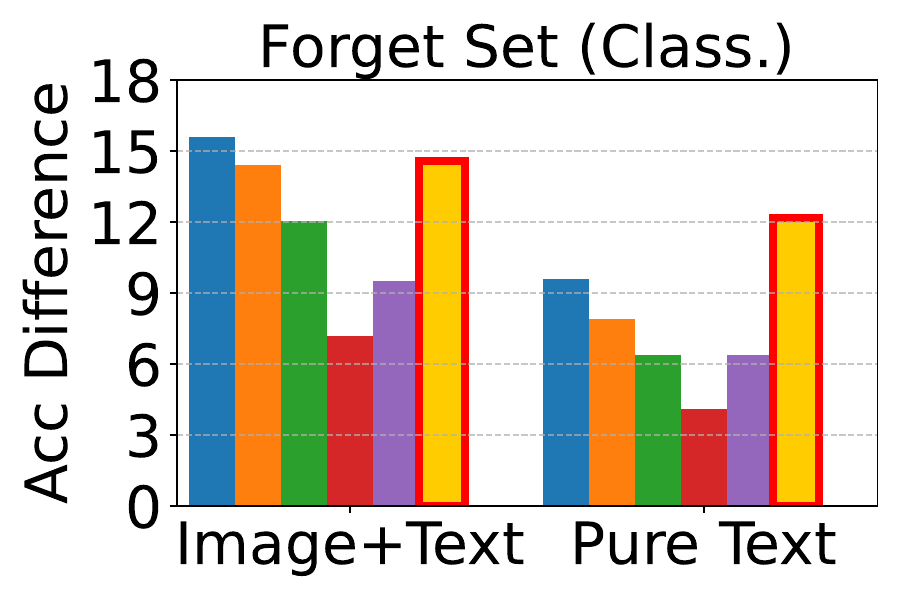}
    \subcaption{Forget Set (Classification)}
    \label{fig:idefics_5_class_forget}
\end{subfigure}    
\begin{subfigure}{0.244\textwidth}
    \includegraphics[width=\textwidth]{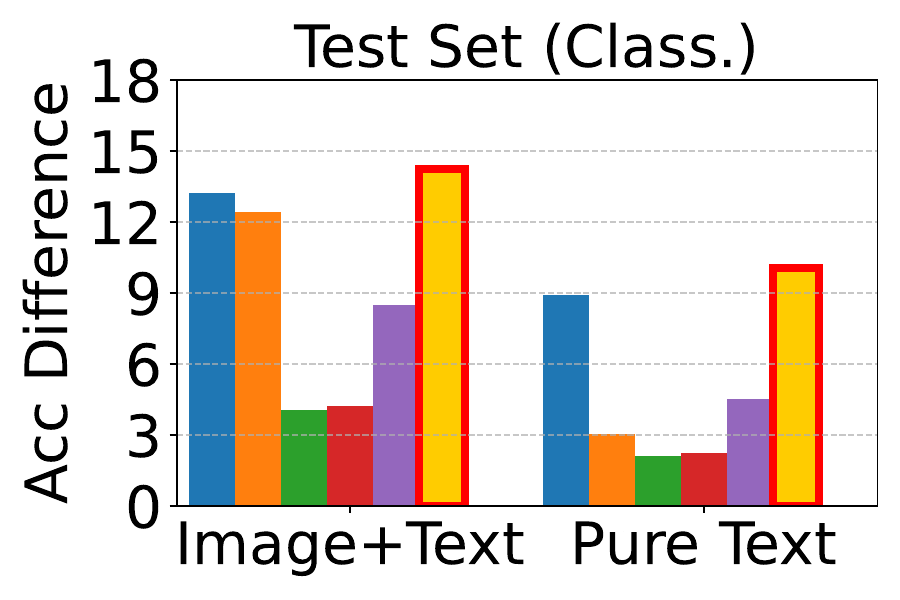}
    \subcaption{Test Set (Classification)}
    \label{fig:idefics_5_class_test}
\end{subfigure}
\begin{subfigure}{0.244\textwidth}
    \includegraphics[width=\textwidth]{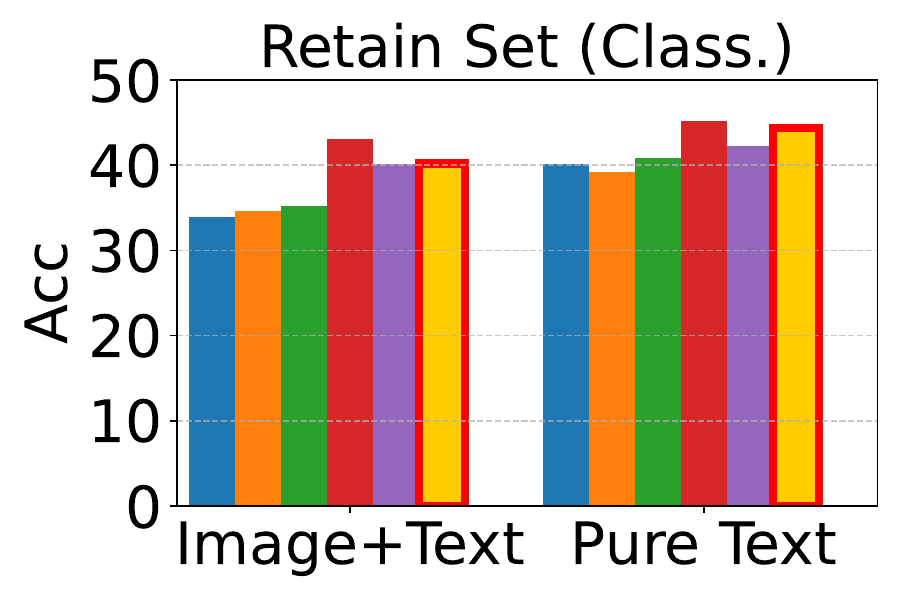}
    \subcaption{Retain Set (Classification)}
    \label{fig:idefics_5_class_retain}
\end{subfigure}    
\begin{subfigure}{0.244\textwidth}
    \includegraphics[width=\textwidth]{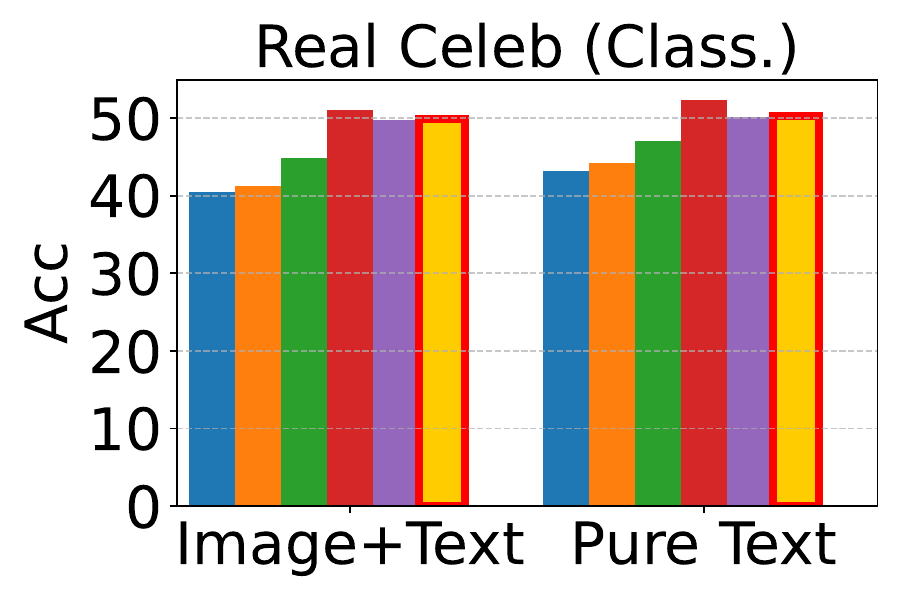} 
    \subcaption{Real Celeb (Classification)}
    \label{fig:idefics_5_class_real}
\end{subfigure}


\begin{subfigure}{0.244\textwidth}
    \includegraphics[width=\textwidth]{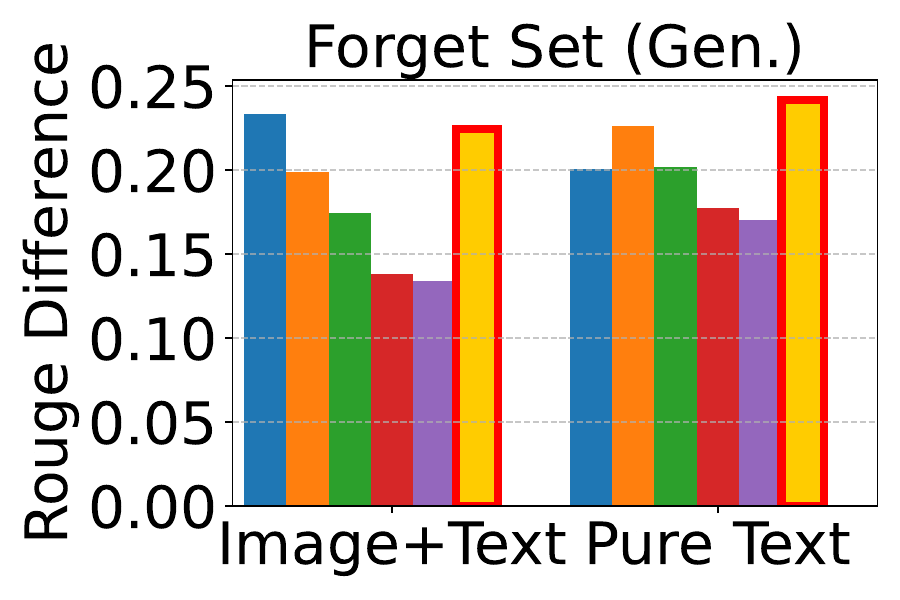}
    \subcaption{Forget Set (Generation)}
    \label{fig:idefics_5_gen_forget}
\end{subfigure}
\begin{subfigure}{0.244\textwidth}
    \includegraphics[width=\textwidth]{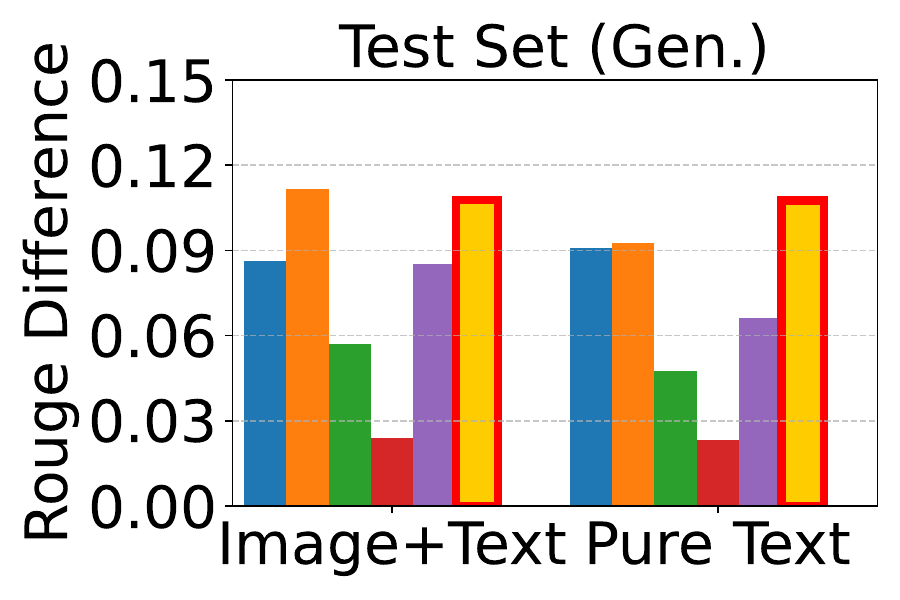}
    \subcaption{Test Set (Generation)}
    \label{fig:idefics_5_gen_test}
\end{subfigure}
\begin{subfigure}{0.244\textwidth}
    \includegraphics[width=\textwidth]{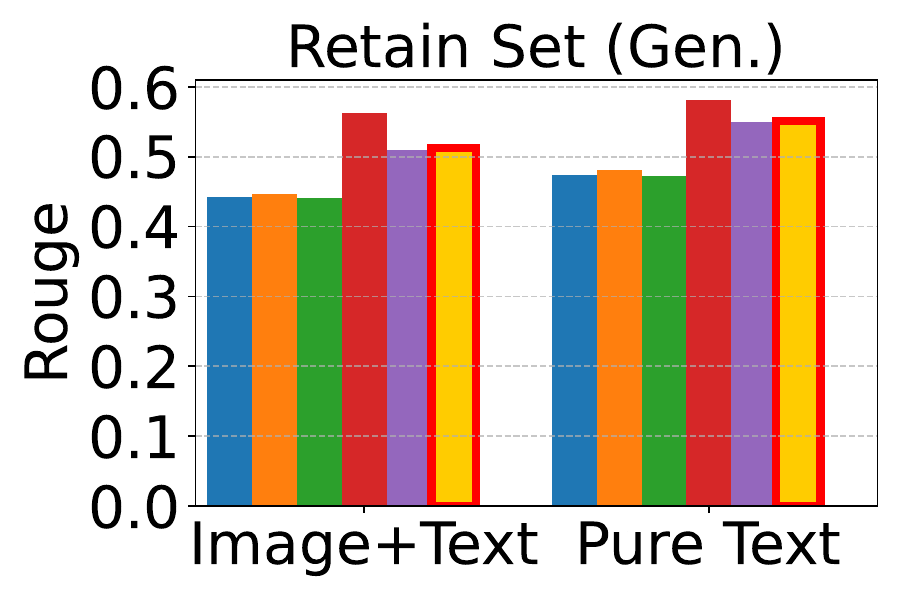}
    \subcaption{Retain Set (Generation)}
    \label{fig:idefics_5_gen_retain}
\end{subfigure}
\begin{subfigure}{0.244\textwidth}
    \includegraphics[width=\textwidth]{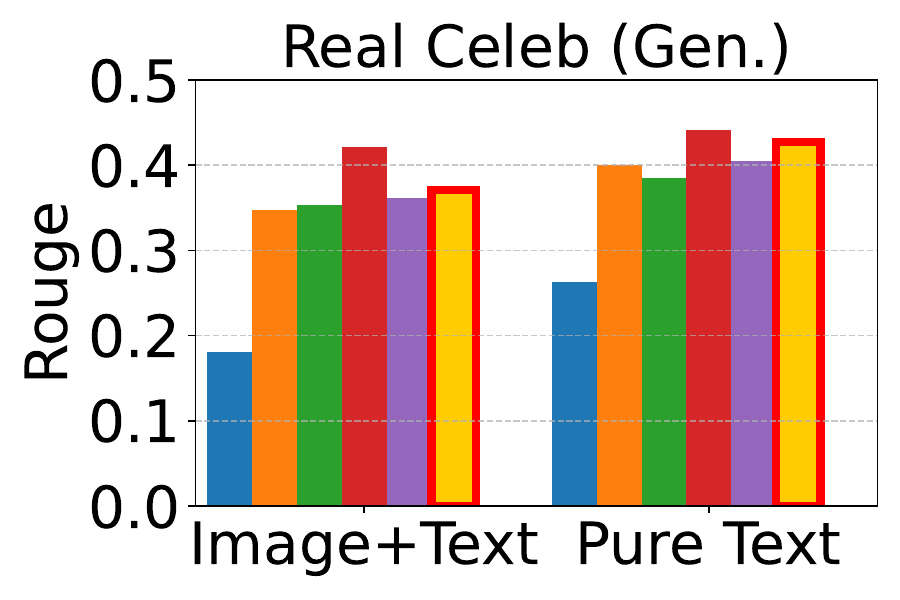}
    \subcaption{Real Celeb (Generation)}
    \label{fig:idefics_5_gen_real}
\end{subfigure}

\begin{subfigure}{0.244\textwidth}
    \includegraphics[width=\textwidth]{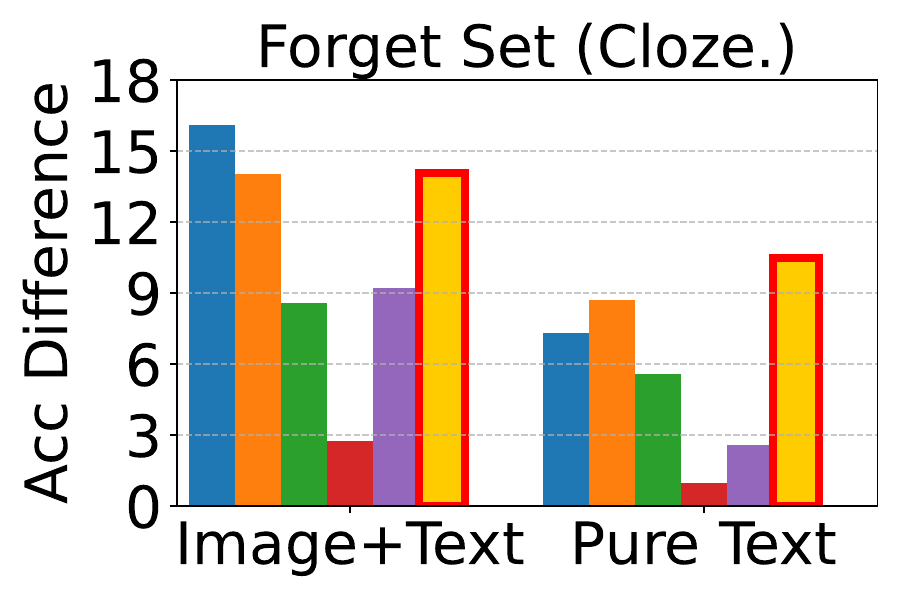}
    \subcaption{Forget Set (Cloze)}
    \label{fig:idefics_5_cloze_forget}
\end{subfigure}
\begin{subfigure}{0.244\textwidth}
    \includegraphics[width=\textwidth]{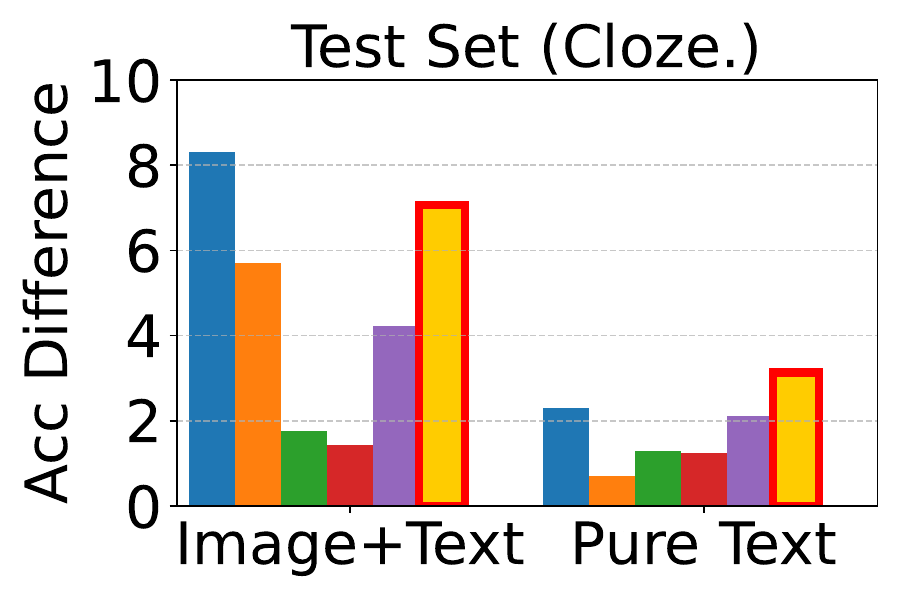}
    \subcaption{Test Set (Cloze)}
    \label{fig:idefics_5_cloze_test}
\end{subfigure}
\begin{subfigure}{0.244\textwidth}
    \includegraphics[width=\textwidth]{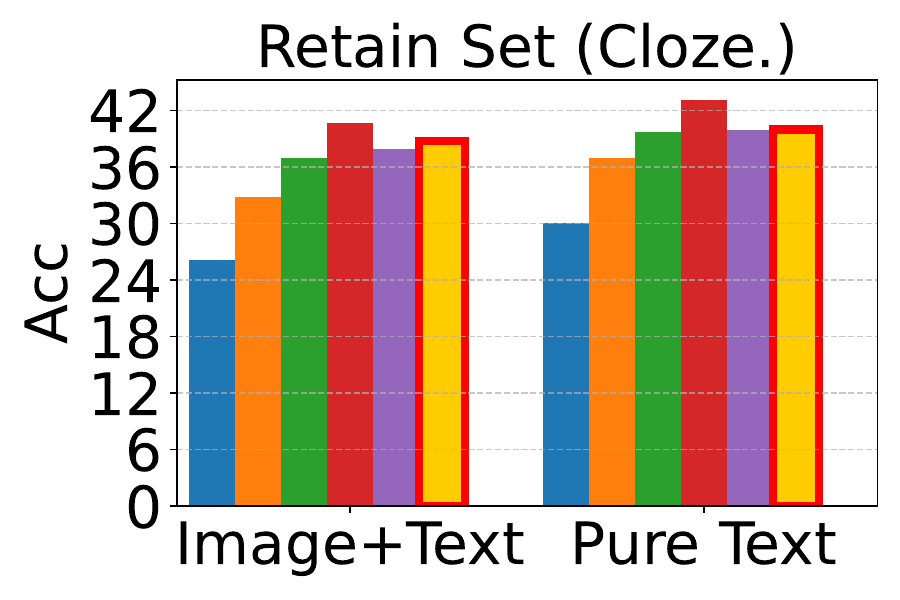}
    \subcaption{Retain Set (Cloze)}
    \label{fig:idefics_5_cloze_retain}
\end{subfigure}
\begin{subfigure}{0.244\textwidth}
    \includegraphics[width=\textwidth]{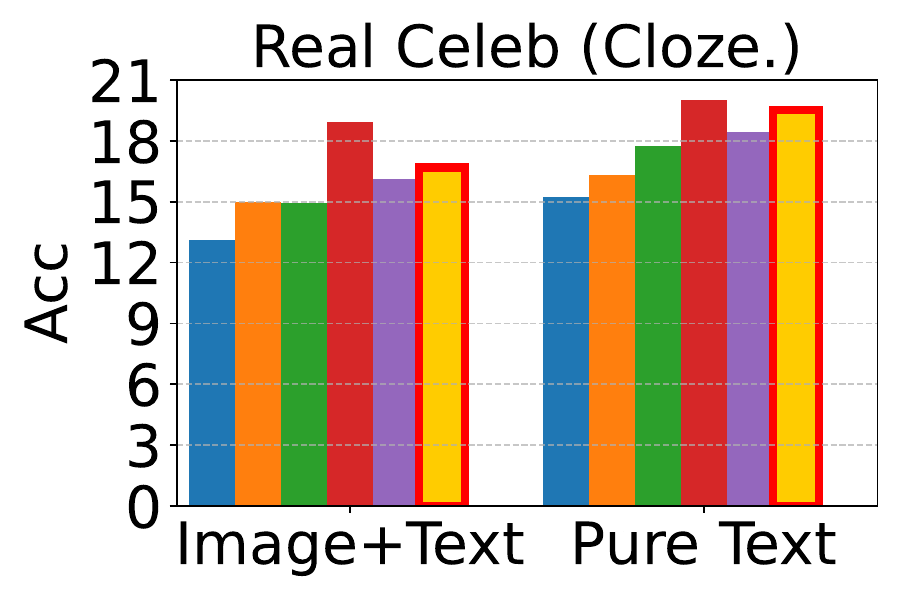}
    \subcaption{Real Celeb (Cloze)}
    \label{fig:idefics_5_cloze_real}
\end{subfigure}
\vspace{-0.1in}
\caption{
Classification, generation, and cloze performance of \method and baselines in multimodal and unimodal setups with 5\% forget data, using Idefics2 as the base model. In subplots (a), (b), (e), (f), (i), and (j), the $y$-axis represents the change in classification accuracy, ROUGE-L score, and cloze accuracy relative to the vanilla model, evaluated on the Forget and Test sets. In the remaining subplots, the $y$-axis indicates classification accuracy, ROUGE-L score, and cloze accuracy, respectively. The $x$-axis represents performance across different modalities.}
\vspace{-0.19in}
\label{fig:idefics_5_compare}
\end{figure*}

\begin{figure*}[!t]
\centering
\begin{subfigure}[b]{\textwidth}
    \centering
    \includegraphics[width=0.65\textwidth]{Figure/idefics2_modality_analyze/idefics_legend.pdf}
\end{subfigure}
\begin{subfigure}{0.244\textwidth}
    \includegraphics[width=\textwidth]{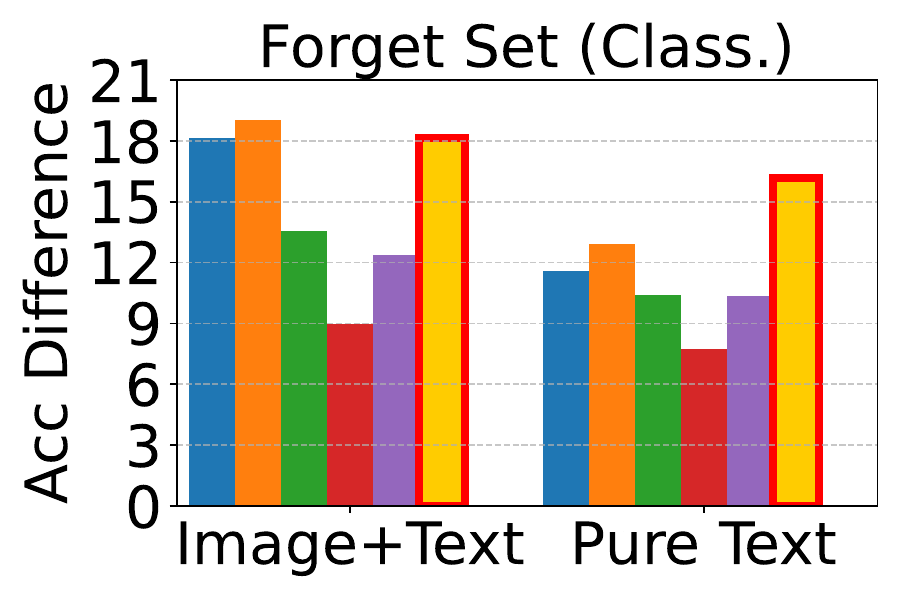}
    \subcaption{Forget Set (Classification)}
    \label{fig:idefics_10_class_forget}
\end{subfigure}    
\begin{subfigure}{0.244\textwidth}
    \includegraphics[width=\textwidth]{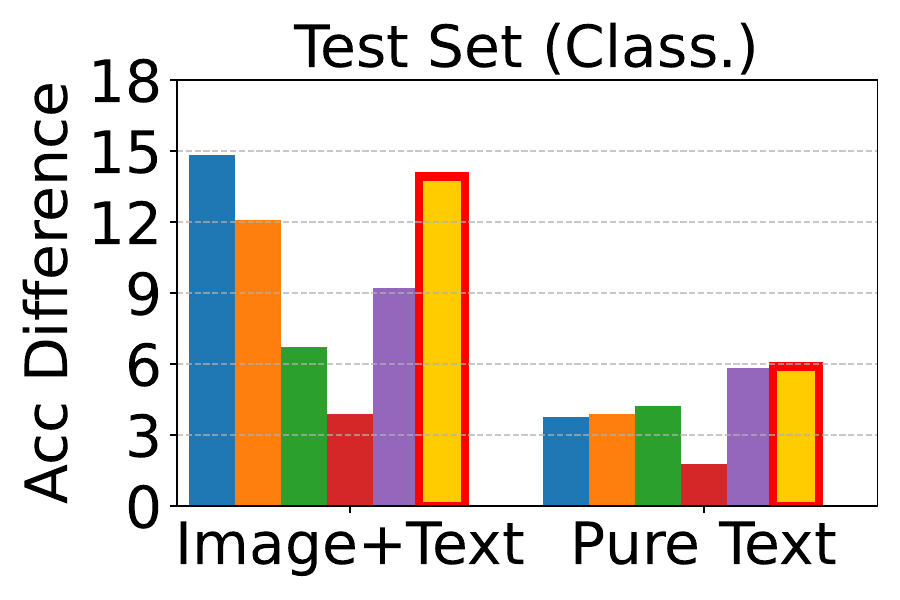}
    \subcaption{Test Set (Classification)}
    \label{fig:idefics_10_class_test}
\end{subfigure}
\begin{subfigure}{0.244\textwidth}
    \includegraphics[width=\textwidth]{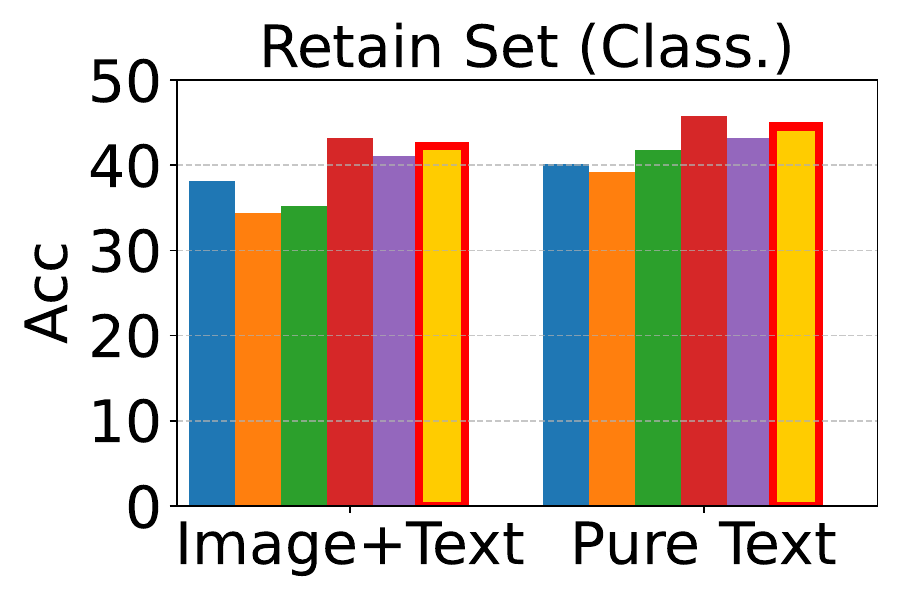}
    \subcaption{Retain Set (Classification)}
    \label{fig:idefics_10_class_retain}
\end{subfigure}    
\begin{subfigure}{0.244\textwidth}
    \includegraphics[width=\textwidth]{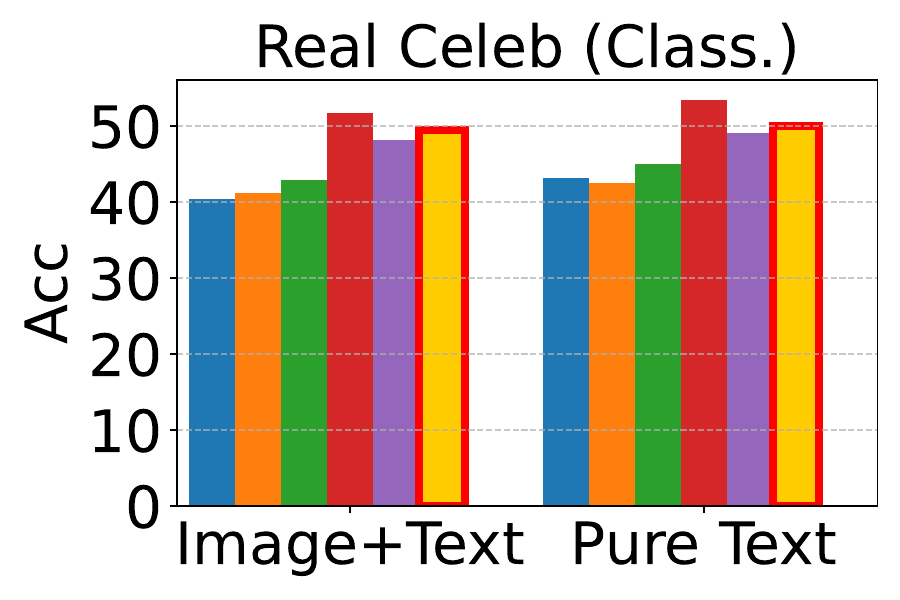} 
    \subcaption{Real Celeb (Classification)}
    \label{fig:idefics_10_class_real}
\end{subfigure}


\begin{subfigure}{0.244\textwidth}
    \includegraphics[width=\textwidth]{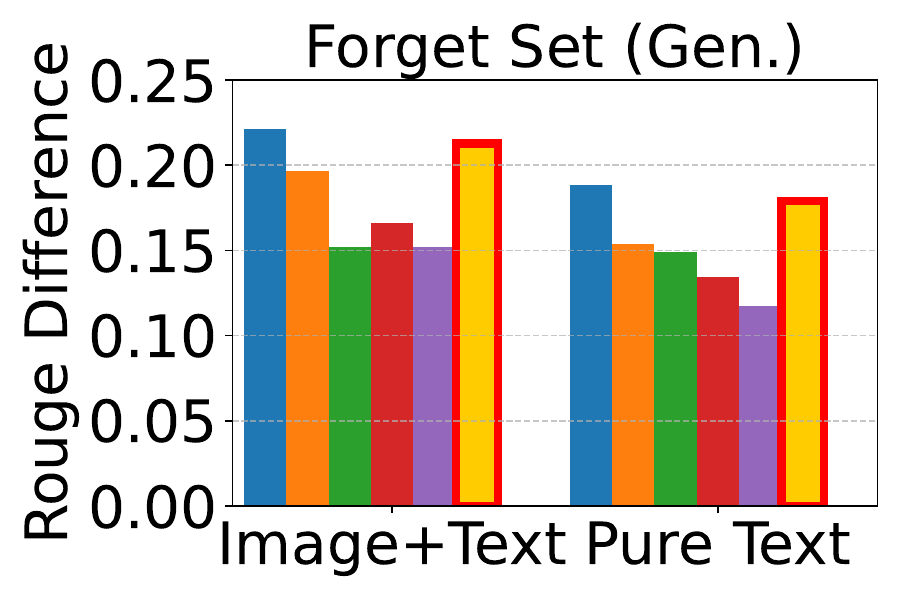}
    \subcaption{Forget Set (Generation)}
    \label{fig:idefics_10_gen_forget}
\end{subfigure}
\begin{subfigure}{0.244\textwidth}
    \includegraphics[width=\textwidth]{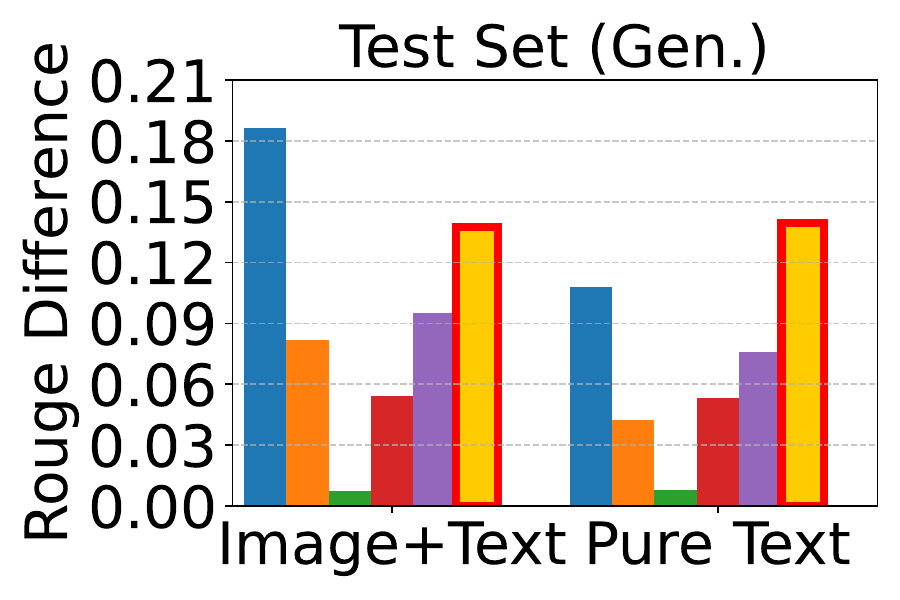}
    \subcaption{Test Set (Generation)}
    \label{fig:idefics_10_gen_test}
\end{subfigure}
\begin{subfigure}{0.244\textwidth}
    \includegraphics[width=\textwidth]{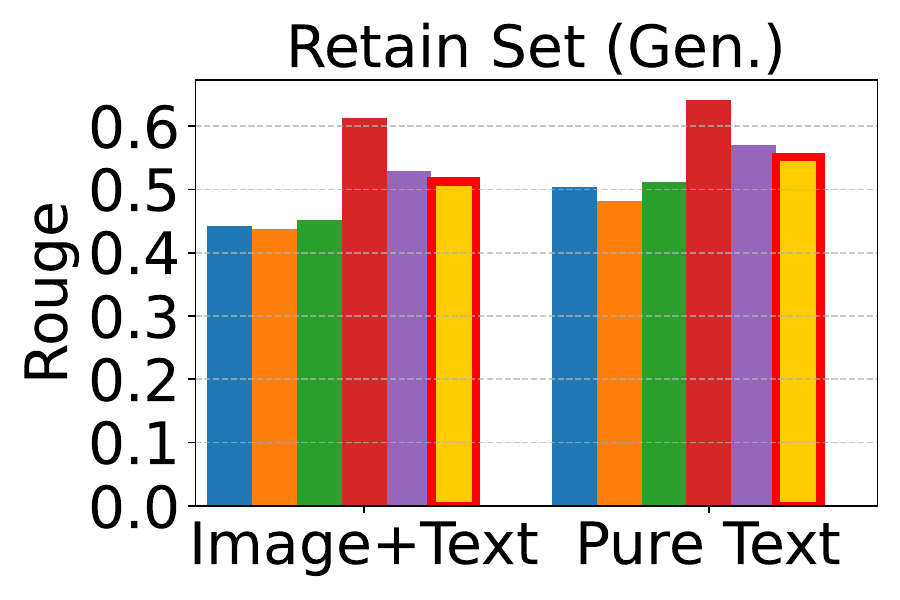}
    \subcaption{Retain Set (Generation)}
    \label{fig:idefics_10_gen_retain}
\end{subfigure}
\begin{subfigure}{0.244\textwidth}
    \includegraphics[width=\textwidth]{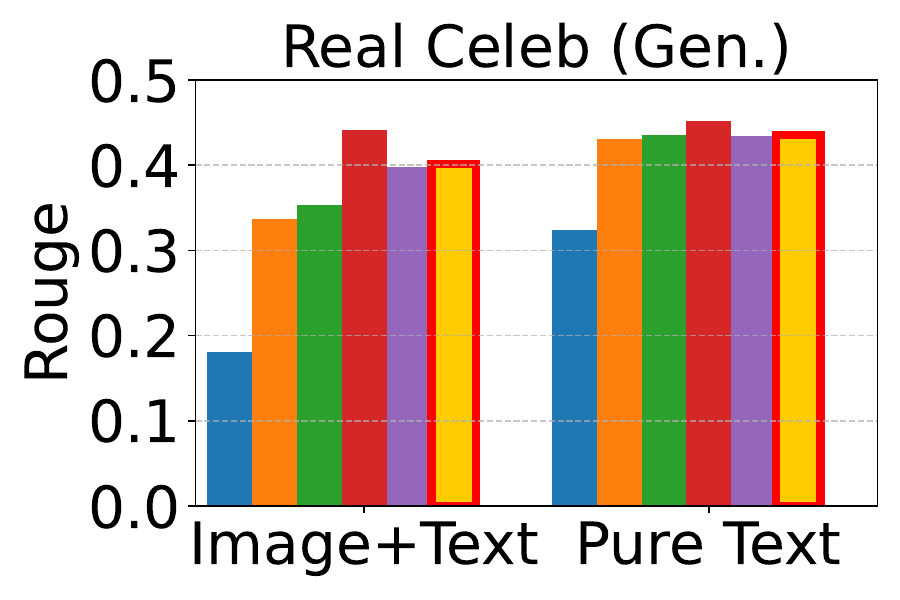}
    \subcaption{Real Celeb (Generation)}
    \label{fig:idefics_10_gen_real}
\end{subfigure}

\begin{subfigure}{0.244\textwidth}
    \includegraphics[width=\textwidth]{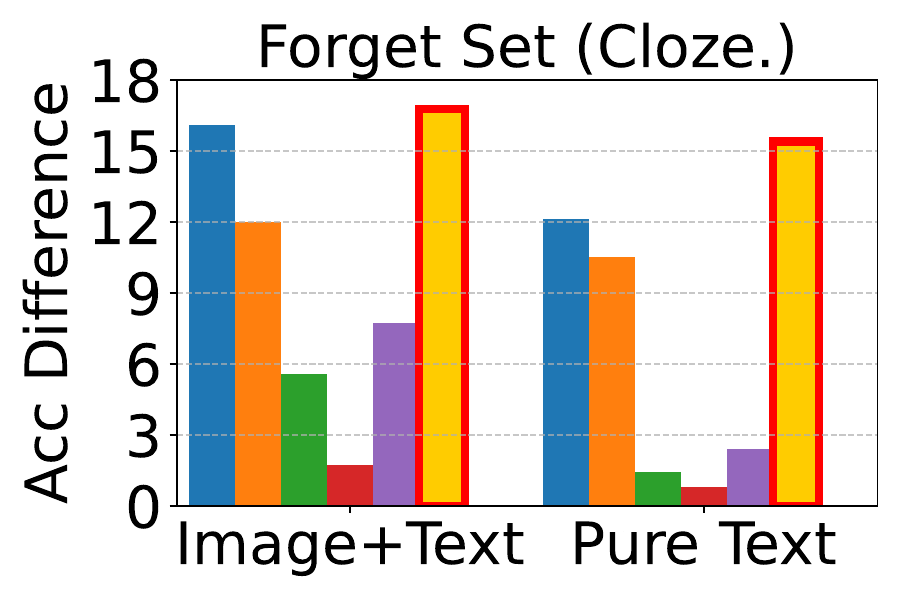}
    \subcaption{Forget Set (Cloze)}
    \label{fig:idefics_10_cloze_forget}
\end{subfigure}
\begin{subfigure}{0.244\textwidth}
    \includegraphics[width=\textwidth]{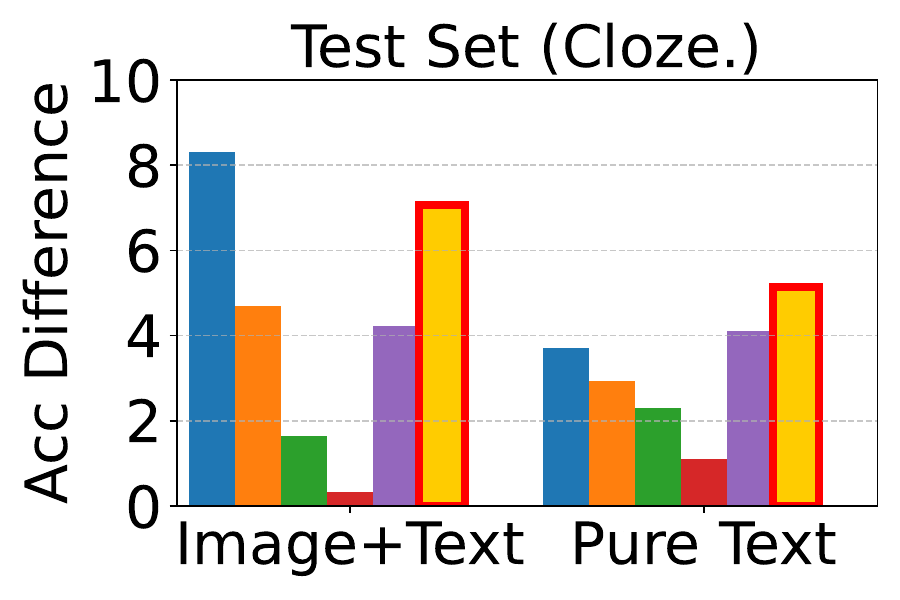}
    \subcaption{Test Set (Cloze)}
    \label{fig:idefics_10_cloze_test}
\end{subfigure}
\begin{subfigure}{0.244\textwidth}
    \includegraphics[width=\textwidth]{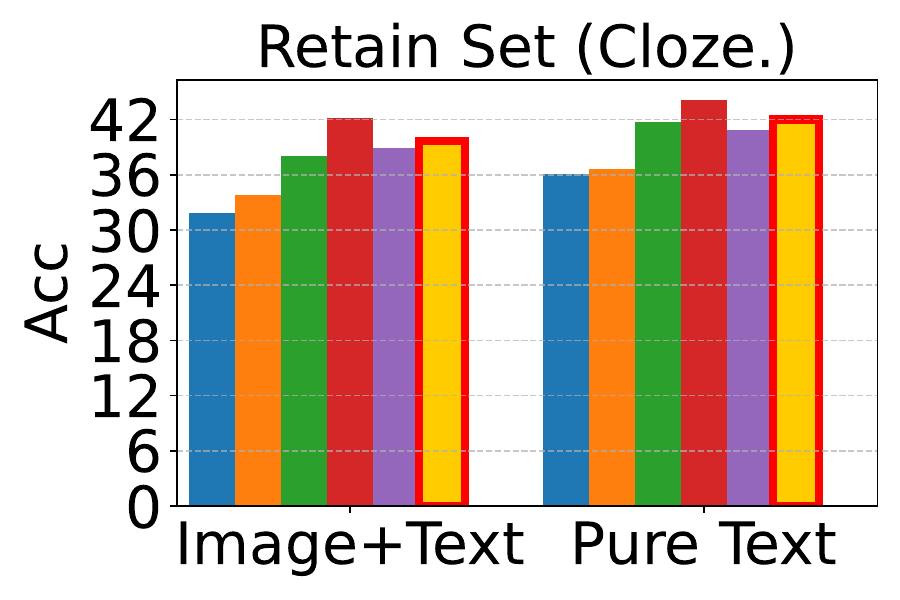}
    \subcaption{Retain Set (Cloze)}
    \label{fig:idefics_10_cloze_retain}
\end{subfigure}
\begin{subfigure}{0.244\textwidth}
    \includegraphics[width=\textwidth]{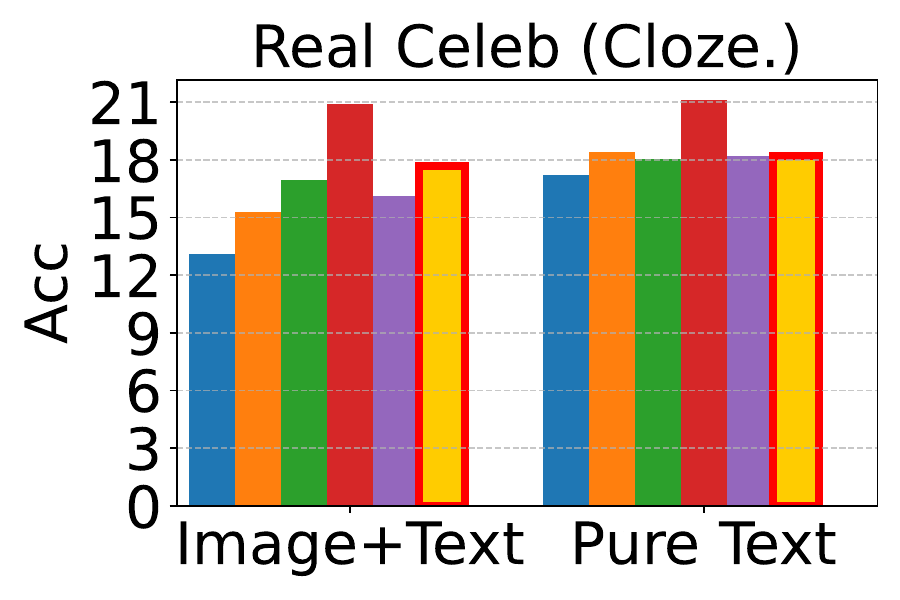}
    \subcaption{Real Celeb (Cloze)}
    \label{fig:idefics_10_cloze_real}
\end{subfigure}
\vspace{-0.1in}
\caption{
Classification, generation, and cloze performance of \method and baselines in multimodal and unimodal setups with 10\% forget data, using Idefics2 as the base model. In subplots (a), (b), (e), (f), (i), and (j), the $y$-axis represents the change in classification accuracy, ROUGE-L score, and cloze accuracy relative to the vanilla model, evaluated on the Forget and Test sets. In the remaining subplots, the $y$-axis indicates classification accuracy, ROUGE-L score, and cloze accuracy, respectively. The $x$-axis represents performance across different modalities.}
\vspace{-0.2in}
\label{fig:idefics_10_compare}
\end{figure*}

\begin{figure*}[!t]
\centering
\begin{subfigure}[b]{\textwidth}
    \centering
    \includegraphics[width=0.7\textwidth]{Figure/idefics2_modality_analyze/idefics_legend.pdf}
\end{subfigure}
\begin{subfigure}{0.244\textwidth}
    \includegraphics[width=\textwidth]{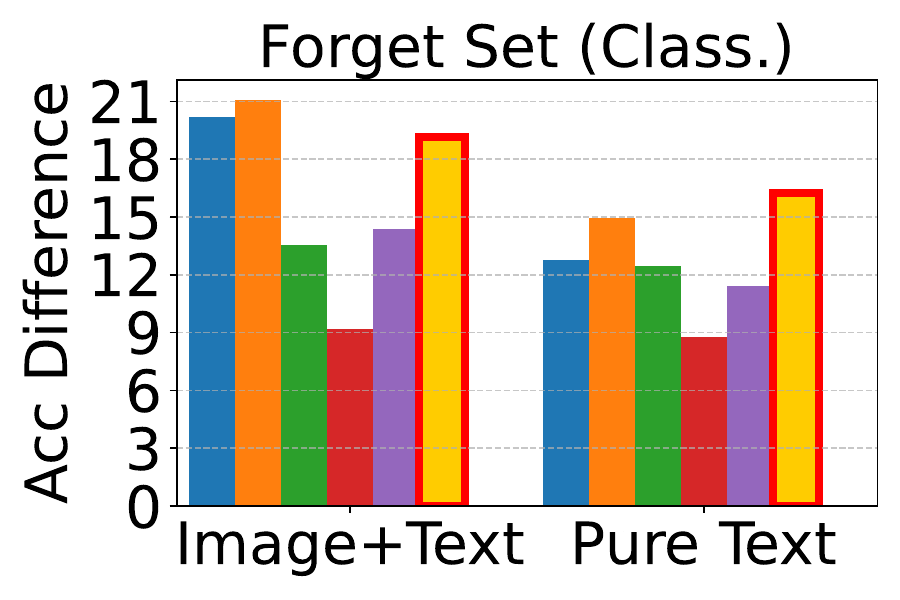}
    \subcaption{Forget Set (Classification)}
    \label{fig:idefics_15_class_forget}
\end{subfigure}    
\begin{subfigure}{0.244\textwidth}
    \includegraphics[width=\textwidth]{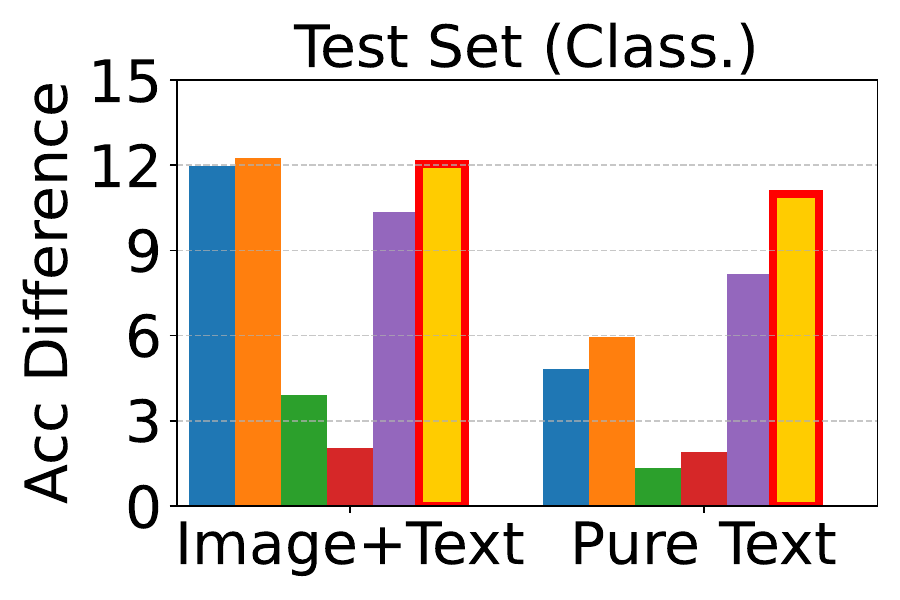}
    \subcaption{Test Set (Classification)}
    \label{fig:idefics_15_class_test}
\end{subfigure}
\begin{subfigure}{0.244\textwidth}
    \includegraphics[width=\textwidth]{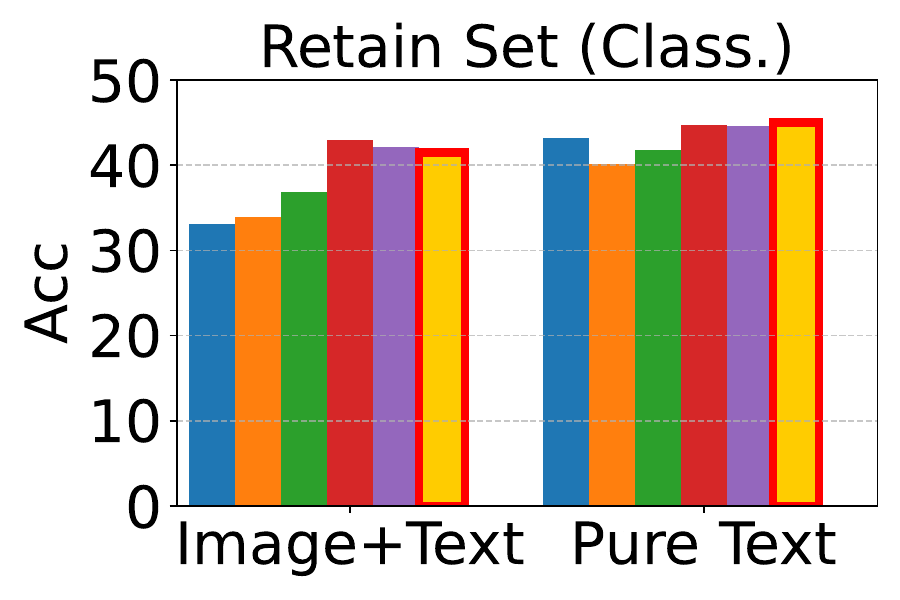}
    \subcaption{Retain Set (Classification)}
    \label{fig:idefics_15_class_retain}
\end{subfigure}    
\begin{subfigure}{0.244\textwidth}
    \includegraphics[width=\textwidth]{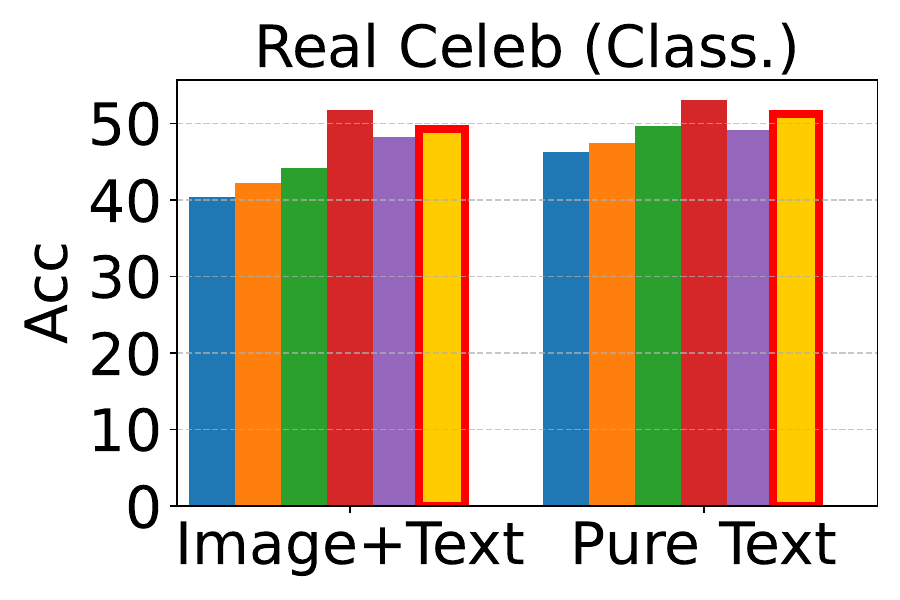} 
    \subcaption{Real Celeb (Classification)}
    \label{fig:idefics_15_class_real}
\end{subfigure}


\begin{subfigure}{0.244\textwidth}
    \includegraphics[width=\textwidth]{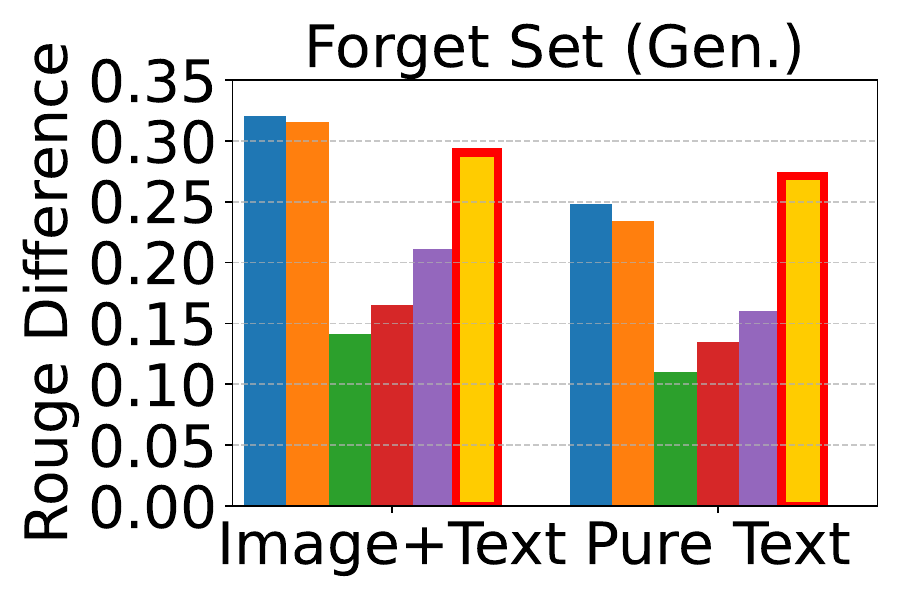}
    \subcaption{Forget Set (Generation)}
    \label{fig:idefics_15_gen_forget}
\end{subfigure}
\begin{subfigure}{0.244\textwidth}
    \includegraphics[width=\textwidth]{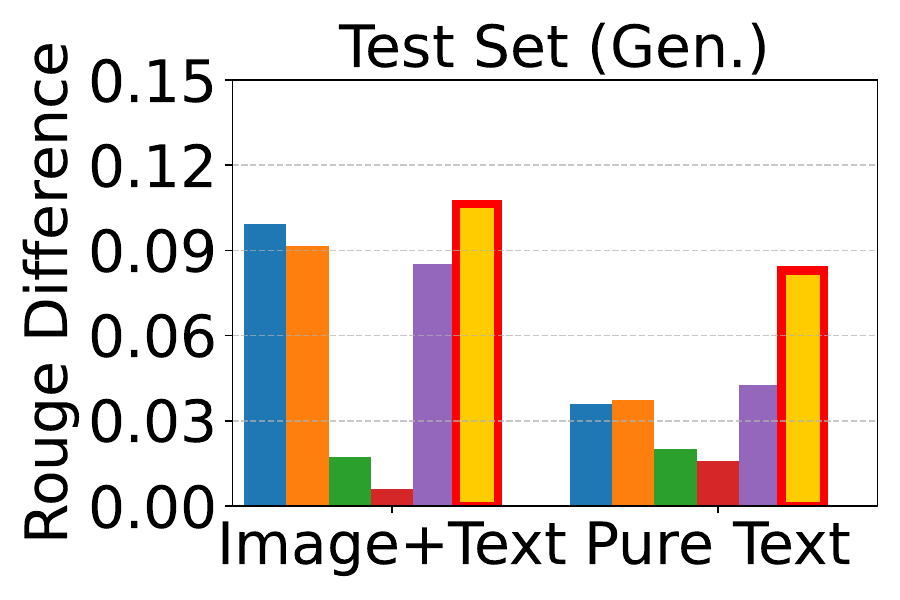}
    \subcaption{Test Set (Generation)}
    \label{fig:idefics_15_gen_test}
\end{subfigure}
\begin{subfigure}{0.244\textwidth}
    \includegraphics[width=\textwidth]{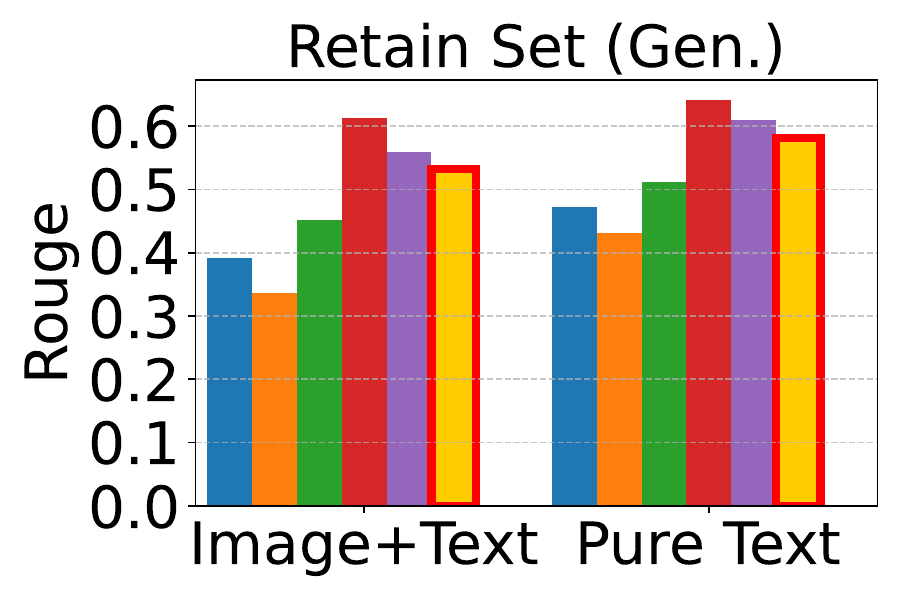}
    \subcaption{Retain Set (Generation)}
    \label{fig:idefics_15_gen_retain}
\end{subfigure}
\begin{subfigure}{0.244\textwidth}
    \includegraphics[width=\textwidth]{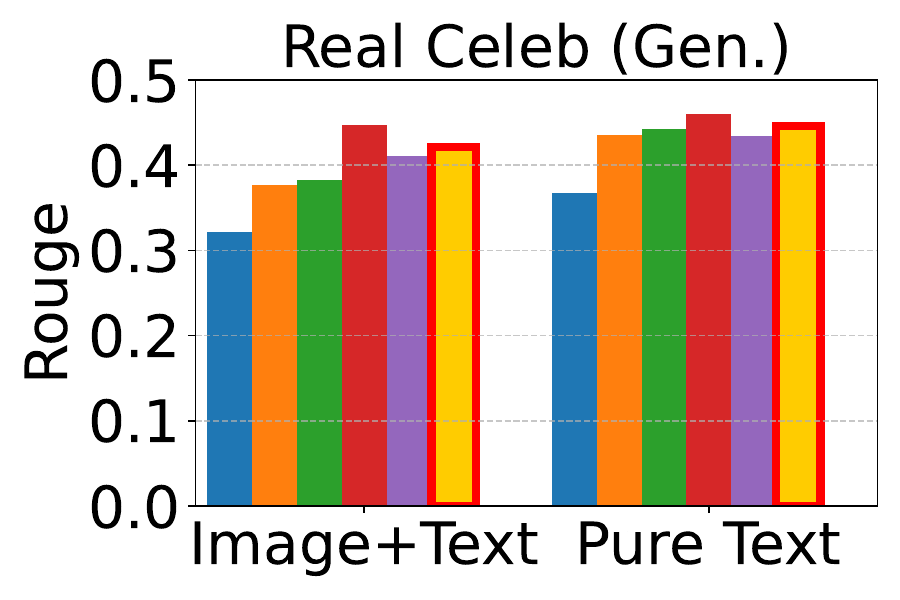}
    \subcaption{Real Celeb (Generation)}
    \label{fig:idefics_15_gen_real}
\end{subfigure}

\begin{subfigure}{0.244\textwidth}
    \includegraphics[width=\textwidth]{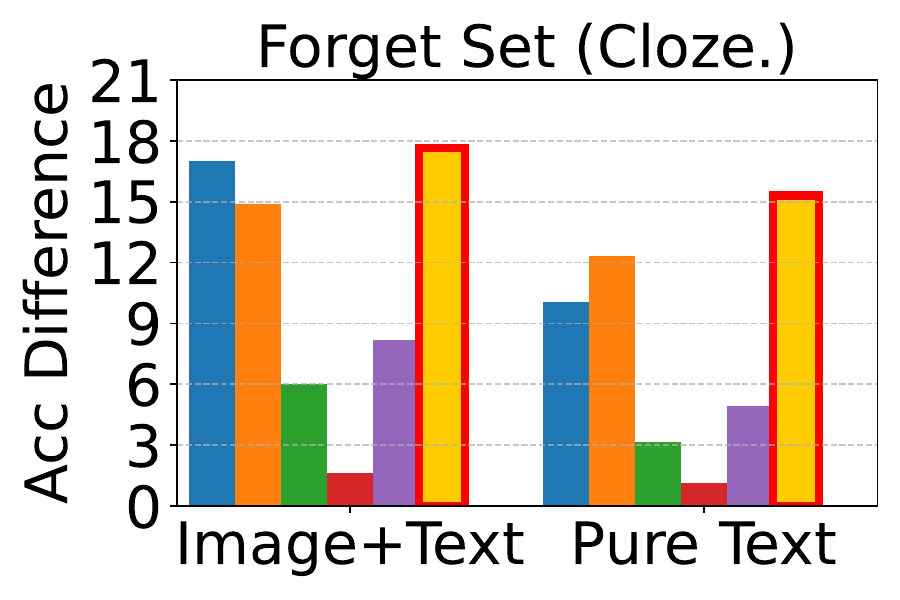}
    \subcaption{Forget Set (Cloze)}
    \label{fig:idefics_15_cloze_forget}
\end{subfigure}
\begin{subfigure}{0.244\textwidth}
    \includegraphics[width=\textwidth]{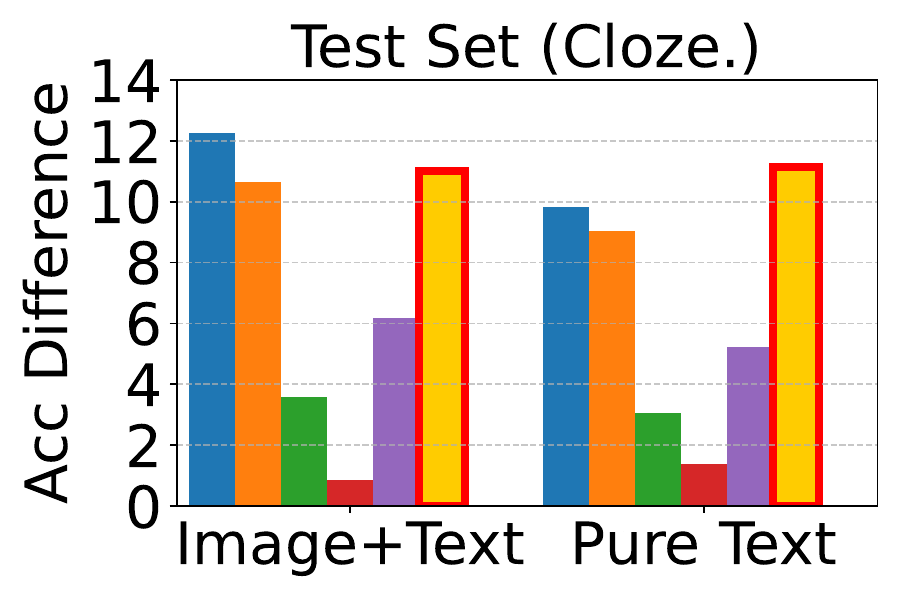}
    \subcaption{Test Set (Cloze)}
    \label{fig:idefics_15_cloze_test}
\end{subfigure}
\begin{subfigure}{0.244\textwidth}
    \includegraphics[width=\textwidth]{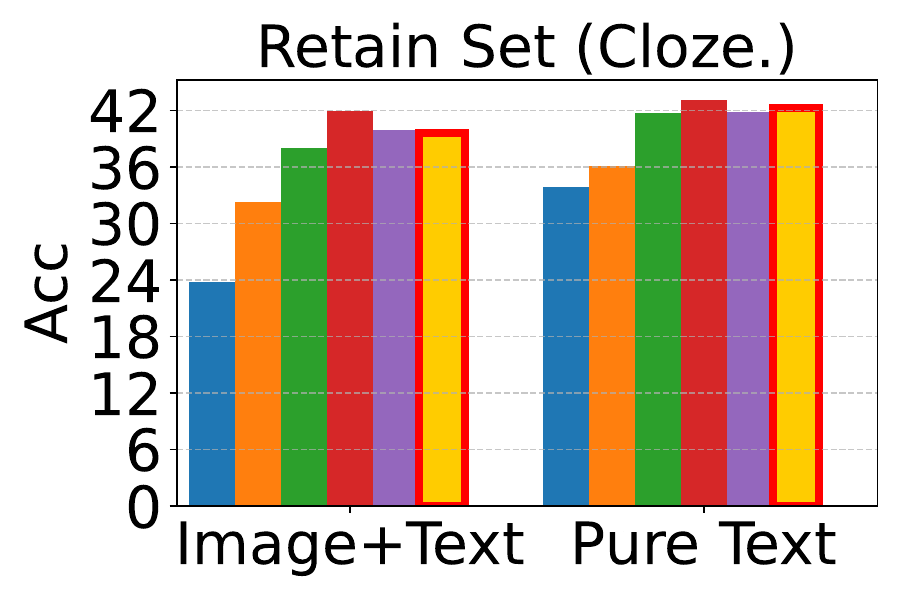}
    \subcaption{Retain Set (Cloze)}
    \label{fig:idefics_15_cloze_retain}
\end{subfigure}
\begin{subfigure}{0.244\textwidth}
    \includegraphics[width=\textwidth]{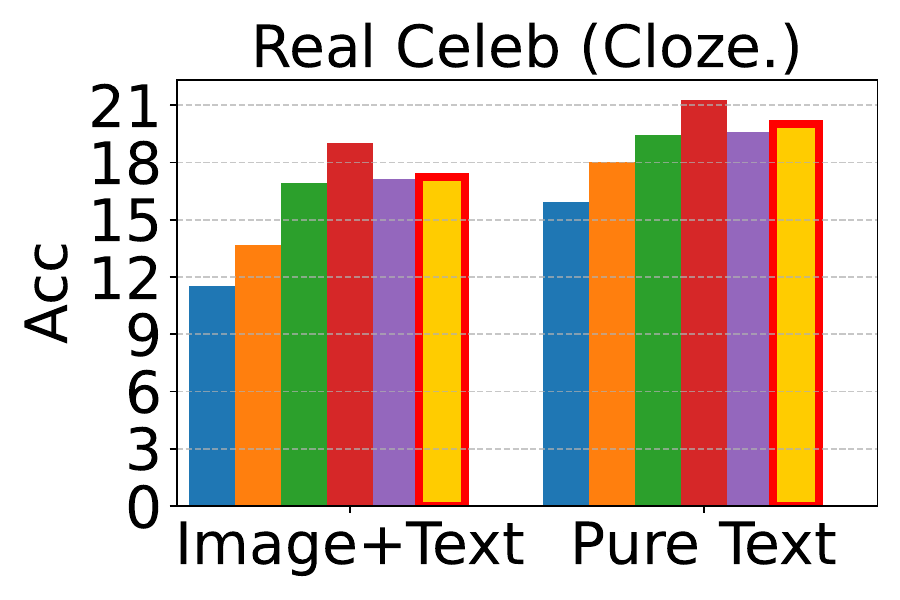}
    \subcaption{Real Celeb (Cloze)}
    \label{fig:idefics_15_cloze_real}
\end{subfigure}
\vspace{-0.1in}
\caption{
Classification, generation, and cloze performance of \method and baselines in multimodal and unimodal setups with 15\% forget data, using Idefics2 as the base model. In subplots (a), (b), (e), (f), (i), and (j), the $y$-axis represents the change in classification accuracy, ROUGE-L score, and cloze accuracy relative to the vanilla model, evaluated on the Forget and Test sets. In the remaining subplots, the $y$-axis indicates classification accuracy, ROUGE-L score, and cloze accuracy, respectively. The $x$-axis represents performance across different modalities.}
\vspace{-0.1in}
\label{fig:idefics_15_compare}
\end{figure*}

\subsection{Appendix: Unlearning v.s. Utility}
\label{appendix:unlearn_vs_utility}

\begin{figure*}
\centering
\begin{subfigure}[b]{\textwidth}
    \centering    \includegraphics[width=0.9\textwidth]{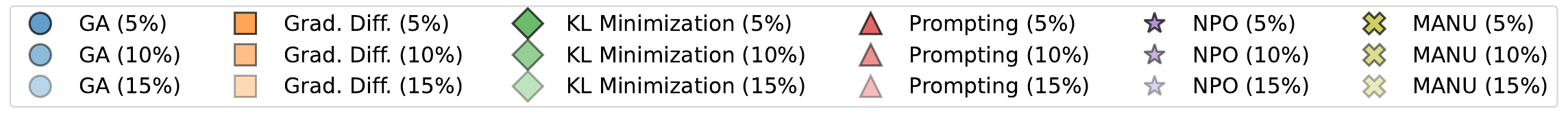}
\end{subfigure}
\begin{subfigure}{0.244\textwidth}
    \includegraphics[width=\textwidth]{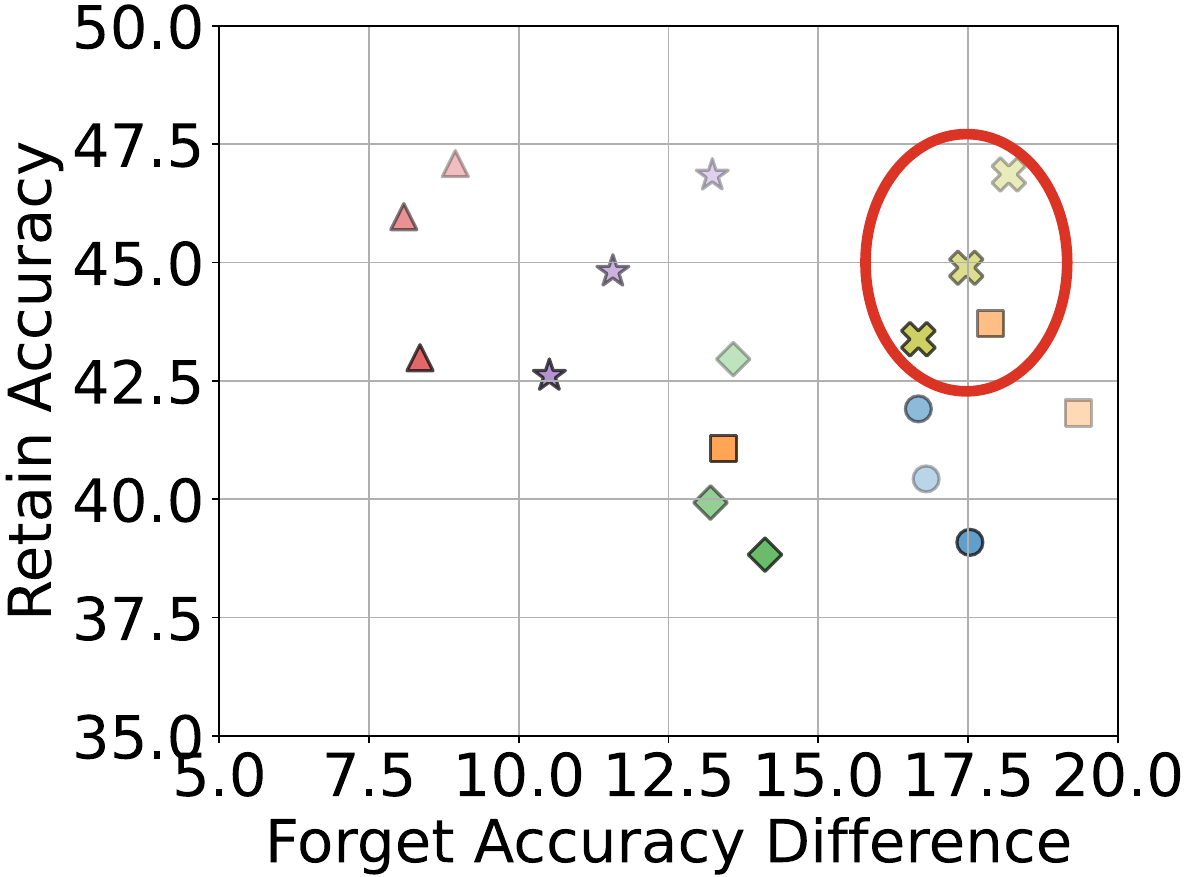}
    \subcaption{Forget Acc vs Retain Acc}
    \label{fig:idefics_forget_retain}
\end{subfigure}    
\begin{subfigure}{0.244\textwidth}
    \includegraphics[width=\textwidth]{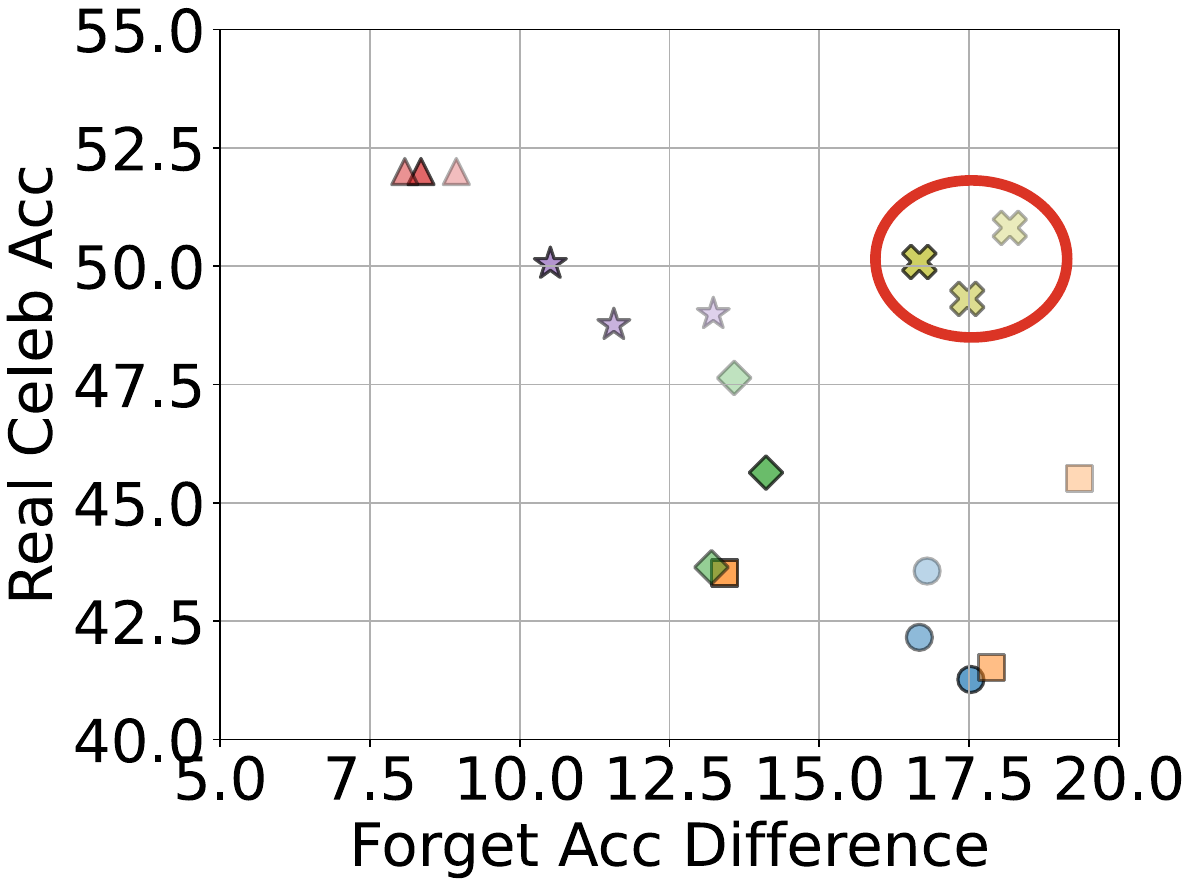}
    \subcaption{Forget Acc vs Real Celeb}
    \label{fig:idefics_forget_real}
\end{subfigure}
\begin{subfigure}{0.244\textwidth}
    \includegraphics[width=\textwidth]{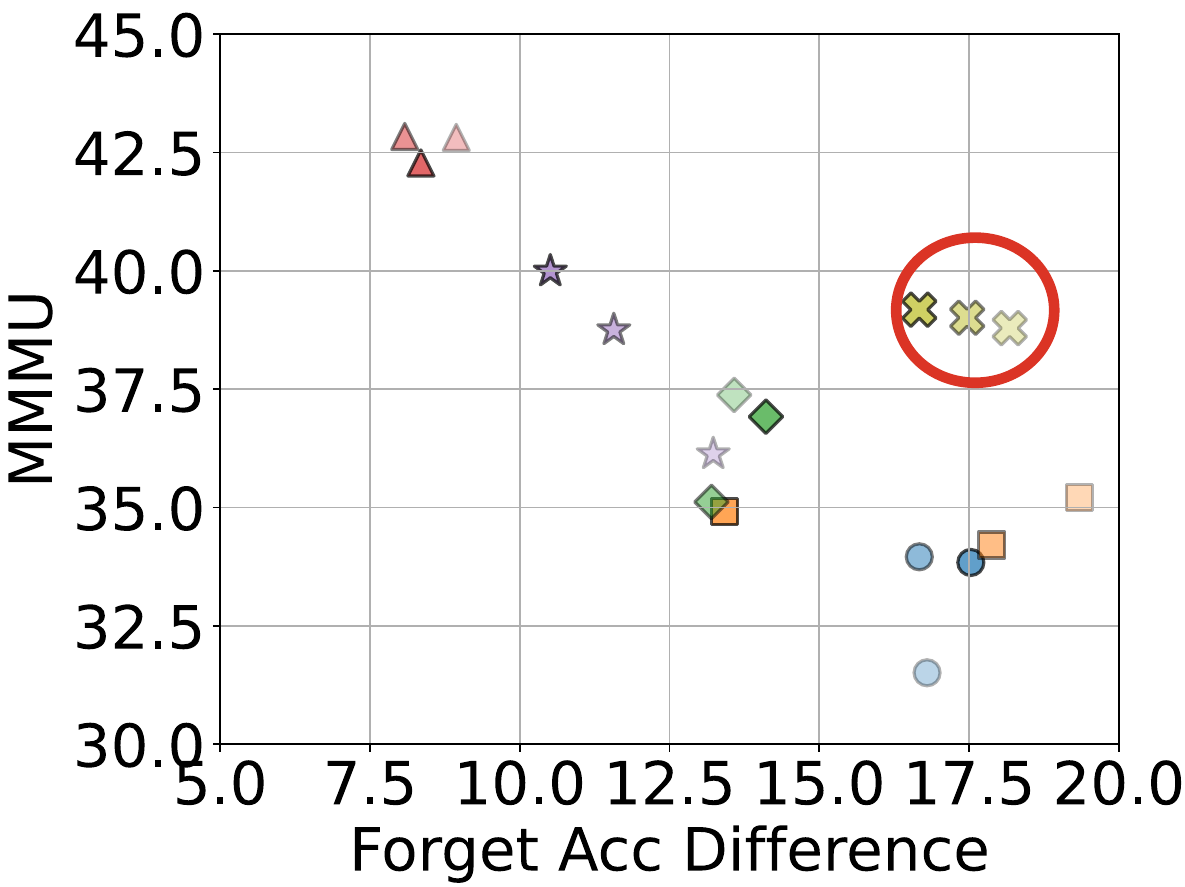}
    \subcaption{Forget Acc vs MMMU}
    \label{fig:idefics_forget_mmmu}
\end{subfigure}
\begin{subfigure}{0.244\textwidth}
    \includegraphics[width=\textwidth]{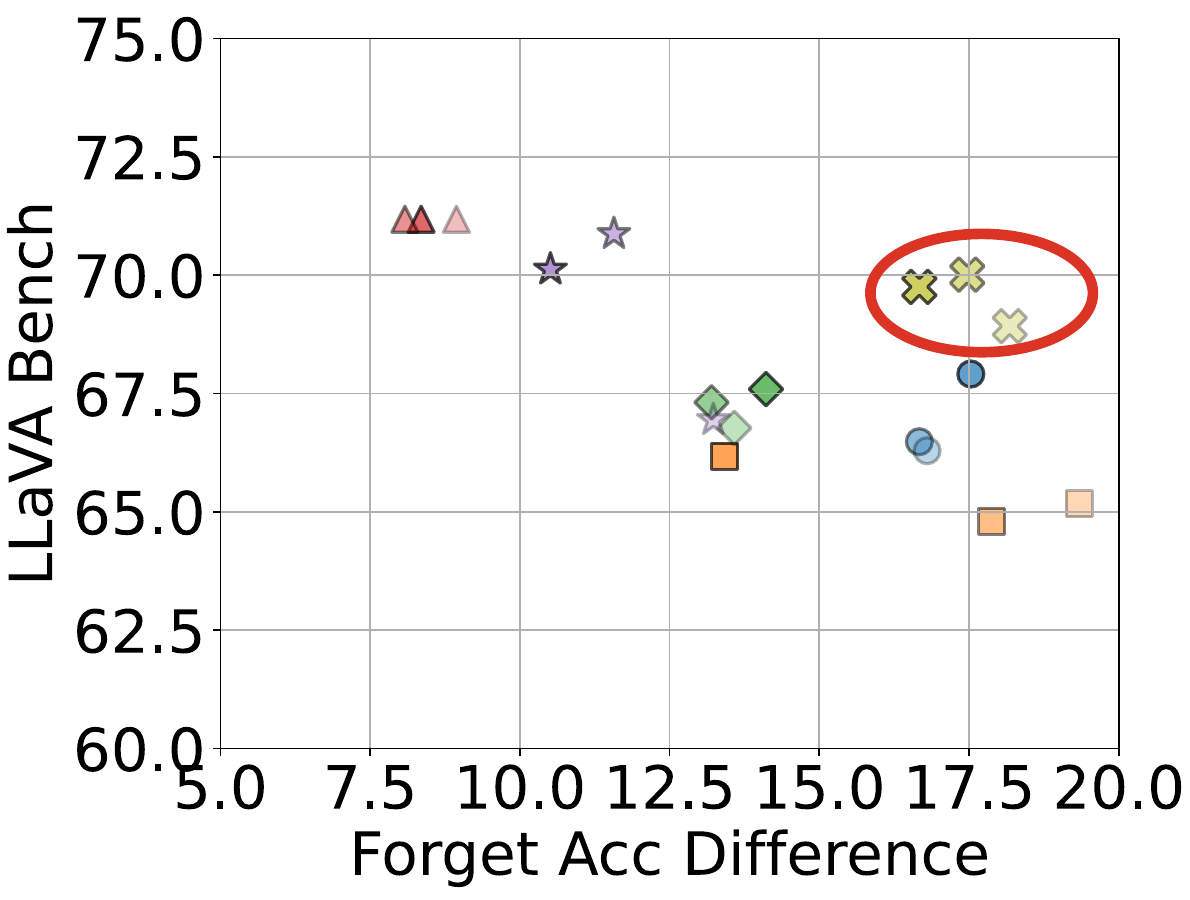}
    \subcaption{Forget Acc vs LLaVABench}
    \label{fig:idefics_forget_llavaB}
\end{subfigure}
\vspace{-0.1in}
\caption{
The overall trade-off between unlearning effectiveness and model utility across all baselines using different forget data, with Idefics2 as the base model. The $x$-axis shows the difference in forget classification accuracy relative to the vanilla model, while the $y$-axis reflects model utility from various perspectives. From left to right, these perspectives include retain accuracy, real celebrity accuracy, MMMU, and LLaVA-Bench performance, respectively.}
\label{fig:idefics_class_tradeoff}
\end{figure*}

In this section, we present additional experiments analyzing the trade-off between unlearning effectiveness and model utility using Idefics2 as the base model. The detailed results are shown in Figure \ref{fig:idefics_class_tradeoff}. Same as the observations in Figure \ref{fig:llava_class_tradeoff}, \method consistently outperforms other baselines, as it is typically closest to the top-right corner—indicating a better balance between unlearning effectiveness and model utility. Notably, \method achieves unlearning performance comparable to GA-based methods while maintaining competitive model utility across different perspectives.

\subsection{Appendix: Additional Utility Datasets}
In addition to the evaluations already included—specifically on MMMU and LLaVA-Bench, we present results on three additional downstream benchmarks to further assess the functional utility of \method across a diverse range of multimodal tasks. These benchmarks include: \textbf{MIA-Bench}~\cite{qian2024mia} for evaluating conversational abilities, \textbf{MM-Vet}~\cite{yu2023mm} for assessing integrated multimodal reasoning, and \textbf{MMStar}~\cite{chen2024we} for testing vision-indispensable capabilities.

In Table \ref{tab:additional_benchmark-results}, we report the average performance of the vanilla model and unlearned models using various unlearning methods on these benchmarks. As shown in the Table, \method performs competitively across all three benchmarks, often outperforming or matching the baselines. These results suggest that \method effectively preserves cross-modal alignment and functional utility, even after selective neuron pruning.

\begin{table}[t!]
\centering
\small
\resizebox{0.9\columnwidth}{!}{%
\begin{tabular}{@{}lccc@{}}
\toprule
\multicolumn{4}{c}{\textbf{LLaVA (5\% Forget Set)}} \\
\midrule
\textbf{Model} & \textbf{MIA-Bench (Avg)} & \textbf{MM-Vet (Avg)} & \textbf{MMStar (Avg)} \\
\midrule
Vanilla        & 69.60 & 34.08 & 34.80 \\
GA             & 57.29 & 28.92 & 29.83 \\
Grad. Diff     & 59.02 & 30.28 & 30.13 \\
KL Minim.      & 62.24 & 31.29 & 31.29 \\
NPO            & 63.44 & 32.38 & 32.32 \\
\textbf{MANU}  & \textbf{64.02} & \textbf{32.76} & \textbf{32.41} \\
\midrule
\multicolumn{4}{c}{\textbf{Idefics2 (5\% Forget Set)}} \\
\midrule
\textbf{Model} & \textbf{MIA-Bench (Avg)} & \textbf{MM-Vet (Avg)} & \textbf{MMStar (Avg)} \\
\midrule
Vanilla        & 54.42 & 43.90 & 49.27 \\
GA             & 47.77 & 35.12 & 40.90 \\
Grad. Diff     & 48.58 & 36.67 & 41.02 \\
KL Minim.      & 49.64 & 38.61 & 41.09 \\
NPO            & 50.89 & 40.22 & 42.88 \\
\textbf{MANU}  & \textbf{51.33} & \textbf{40.83} & \textbf{44.71} \\
\bottomrule
\end{tabular}
}
\caption{Evaluation results on MIA-Bench, MM-Vet, and MMStar for LLaVA and Idefics2 under the 5\% Forget Set setting. Higher scores indicate better performance.}
\vspace{-0.2in}
\label{tab:additional_benchmark-results}
\end{table}

\subsection{Appendix: Ablations on Importance Functions}
\label{appendix:ablations_importance_functions}
To further investigate the contribution of each importance function, here we provide additional ablation studies where we iteratively zeroed out each importance function in the scoring formula. The results are displayed in Table \ref{tab:ablation_study_importance}. From the ablation results, we observe that each importance function contributes uniquely to the overall effectiveness of \method, and removing any single component results in a noticeable trade-off between unlearning performance and utility preservation. Since the trends for LLaVA and Idefics2 are consistent, we use the LLaVA results as a representative example. 

Specifically, removing Frequency Importance ($I_{\text{freq}}$) or Variance Importance ($I_{\text{var}}$) substantially worsens unlearning on the Forget and Test sets—e.g., classification accuracy rises from 41.25\% (\method) to 47.35\% and 46.67\%, respectively, indicating a failure to sufficiently erase the target knowledge. These two metrics are particularly valuable for identifying neurons consistently and distinctively activated by forget data, thus supporting targeted unlearning. On the other hand, removing Absolute Importance ($I_{\text{abs}}$) or RMS Importance ($I_{\text{rms}}$) more prominently degrades performance on the Retain and Real Celebrity sets. For instance, when $I_{\text{abs}}$ is excluded, Retain classification accuracy drops from 43.38\% to 42.37\%, and Real Celebrity classification accuracy declines to 46.59\%. This suggests that $I_{\text{abs}}$ and $I_{\text{rms}}$ are important for preserving high-activation neurons that contribute broadly to general reasoning, thus maintaining utility. These findings support our equal-weighting strategy, where each importance score captures a distinct and complementary signal. While we acknowledge that learned or tuned weightings might yield further improvements, we leave such model-specific enhancements for future work.

\begin{table*}[t!]
    \centering
\scalebox{0.51}{
\begin{tabular}{l|cccc|cccc|cccc|cccc}
        \toprule
        \multirow{3}{*}{\textbf{Models}} 
        & \multicolumn{4}{c|}{\textbf{Forget Set}} 
        & \multicolumn{4}{c|}{\textbf{Test Set}} 
        & \multicolumn{4}{c|}{\textbf{Retain Set}} 
        & \multicolumn{4}{c}{\textbf{Real Celebrity}} \\
        \cline{2-17}
        & \begin{tabular}[c]{@{}c@{}}Class.\\ Acc (\textcolor{blue}{$\downarrow$})\end{tabular} 
        & \begin{tabular}[c]{@{}c@{}}Rouge\\ Score (\textcolor{red}{$\downarrow$})\end{tabular} 
        & \begin{tabular}[c]{@{}c@{}}Fact.\\ Score (\textcolor{red}{$\downarrow$})\end{tabular} 
        & \begin{tabular}[c]{@{}c@{}}Cloze\\ Acc (\textcolor{teal}{$\downarrow$})\end{tabular} 
        & \begin{tabular}[c]{@{}c@{}}Class.\\ Acc (\textcolor{blue}{$\downarrow$})\end{tabular} 
        & \begin{tabular}[c]{@{}c@{}}Rouge\\ Score (\textcolor{red}{$\downarrow$})\end{tabular} 
        & \begin{tabular}[c]{@{}c@{}}Fact.\\ Score (\textcolor{red}{$\downarrow$})\end{tabular} 
        & \begin{tabular}[c]{@{}c@{}}Cloze\\ Acc (\textcolor{teal}{$\downarrow$})\end{tabular} 
        & \begin{tabular}[c]{@{}c@{}}Class.\\ Acc (\textcolor{blue}{$\uparrow$})\end{tabular} 
        & \begin{tabular}[c]{@{}c@{}}Rouge\\ Score (\textcolor{red}{$\uparrow$})\end{tabular} 
        & \begin{tabular}[c]{@{}c@{}}Fact.\\ Score (\textcolor{red}{$\uparrow$})\end{tabular} 
        & \begin{tabular}[c]{@{}c@{}}Cloze\\ Acc (\textcolor{teal}{$\uparrow$})\end{tabular} 
        & \begin{tabular}[c]{@{}c@{}}Class.\\ Acc (\textcolor{blue}{$\uparrow$})\end{tabular} 
        & \begin{tabular}[c]{@{}c@{}}Rouge\\ Score (\textcolor{red}{$\uparrow$})\end{tabular} 
        & \begin{tabular}[c]{@{}c@{}}Fact.\\ Score (\textcolor{red}{$\uparrow$})\end{tabular} 
        & \begin{tabular}[c]{@{}c@{}}Cloze\\ Acc (\textcolor{teal}{$\uparrow$})\end{tabular} \\
        \midrule
        \multicolumn{17}{c}{\textbf{LLaVA-7B (5\% Forget)}} \\
        \midrule
        Vanilla        & 51.70 & 0.645 & 6.78 & 25.81 & 47.86 & 0.539 & 4.89 & 23.01 & 46.11 & 0.632 & 6.41 & 27.83 & 51.80 & 0.479 & 5.47 & 17.35 \\
        w/o $I_{abs}$  & 40.33 & 0.472 & 3.30 & 16.83 & 41.67 & 0.354 & 3.62 & 14.15 & 42.37 & 0.458 & 4.23 & 19.22 & 46.59 & 0.375 & 4.12 & 13.38 \\
        w/o $I_{freq}$ & 47.35 & 0.612 & 4.09 & 25.01 & 45.67 & 0.472 & 4.22 & 21.78 & 45.12 & 0.522 & 5.16 & 24.16 & 49.86 & 0.459 & 4.99 & 15.73 \\
        w/o $I_{var}$  & 46.67 & 0.636 & 4.14 & 24.92 & 44.17 & 0.460 & 4.04 & 22.75 & 44.09 & 0.564 & 5.39 & 24.64 & 50.61 & 0.461 & 4.85 & 16.07 \\
        w/o $I_{rms}$  & 40.82 & 0.500 & 3.19 & 15.17 & 40.33 & 0.382 & 3.70 & 16.03 & 41.02 & 0.471 & 4.19 & 20.68 & 46.08 & 0.466 & 4.05 & 12.03 \\
        MANU           & 41.25 & 0.491 & 3.27 & 17.08 & 41.67 & 0.334 & 3.81 & 15.78 & 43.38 & 0.542 & 4.45 & 24.08 & 49.57 & 0.448 & 4.67 & 16.01 \\
        \midrule
        \multicolumn{17}{c}{\textbf{Idefics2-8B (5\% Forget)}} \\
        \midrule
        Vanilla        & 53.80 & 0.630 & 6.22 & 4.75 & 47.86 & 0.434 & 5.00 & 24.97 & 46.11 & 0.644 & 6.51 & 42.35 & 52.75 & 0.459 & 5.75 & 20.05 \\
        w/o $I_{abs}$  & 35.98 & 0.403 & 3.17 & 3.32 & 33.86 & 0.354 & 3.89 & 18.55 & 41.06 & 0.542 & 4.50 & 39.22 & 48.82 & 0.366 & 3.92 & 14.91 \\
        w/o $I_{freq}$ & 48.81 & 0.608 & 4.72 & 4.41 & 42.39 & 0.407 & 4.07 & 23.49 & 44.14 & 0.583 & 5.87 & 41.09 & 51.54 & 0.428 & 5.09 & 19.33 \\
        w/o $I_{var}$  & 46.65 & 0.588 & 4.66 & 4.83 & 44.77 & 0.395 & 4.14 & 23.82 & 45.34 & 0.626 & 5.80 & 41.43 & 52.22 & 0.417 & 4.99 & 19.15 \\
        w/o $I_{rms}$  & 36.18 & 0.420 & 3.05 & 2.18 & 32.19 & 0.324 & 3.32 & 17.77 & 40.91 & 0.507 & 4.44 & 37.19 & 46.88 & 0.342 & 3.73 & 14.11 \\
        MANU (full)    & 37.13 & 0.413 & 3.20 & 2.11 & 35.55 & 0.361 & 3.91 & 20.97 & 42.71 & 0.538 & 4.60 & 40.01 & 50.09 & 0.399 & 4.11 & 18.80 \\
        \bottomrule
    \end{tabular}}
    \vspace{-0.1in}
    \caption{Ablation study of \method on two base MLLM models under a 5\% forget data setup. Lower scores on the Forget/Test sets indicate better unlearning, while higher scores on the Retain/Celebrity sets indicate better utility preservation.}
    \label{tab:ablation_study_importance}
    \vspace{-0.15in}
\end{table*}

\subsection{Appendix: Generalizability with Larger MLLMs}
\label{appendix:generalizability_MLLMs}

\begin{table*}[t!]
    \centering
\scalebox{0.51}{
\begin{tabular}{l|cccc|cccc|cccc|cccc}
        \toprule
        \multirow{3}{*}{\textbf{Model}} 
        & \multicolumn{4}{c|}{\textbf{Forget Set}} 
        & \multicolumn{4}{c|}{\textbf{Test Set}} 
        & \multicolumn{4}{c|}{\textbf{Retain Set}} 
        & \multicolumn{4}{c}{\textbf{Real Celebrity}} \\
        \cline{2-17}
        & \begin{tabular}[c]{@{}c@{}}Class.\\ Acc (\textcolor{blue}{$\downarrow$})\end{tabular} 
        & \begin{tabular}[c]{@{}c@{}}Rouge\\ Score (\textcolor{red}{$\downarrow$})\end{tabular} 
        & \begin{tabular}[c]{@{}c@{}}Fact.\\ Score (\textcolor{red}{$\downarrow$})\end{tabular} 
        & \begin{tabular}[c]{@{}c@{}}Cloze\\ Acc (\textcolor{teal}{$\downarrow$})\end{tabular} 
        & \begin{tabular}[c]{@{}c@{}}Class.\\ Acc (\textcolor{blue}{$\downarrow$})\end{tabular} 
        & \begin{tabular}[c]{@{}c@{}}Rouge\\ Score (\textcolor{red}{$\downarrow$})\end{tabular} 
        & \begin{tabular}[c]{@{}c@{}}Fact.\\ Score (\textcolor{red}{$\downarrow$})\end{tabular} 
        & \begin{tabular}[c]{@{}c@{}}Cloze\\ Acc (\textcolor{teal}{$\downarrow$})\end{tabular} 
        & \begin{tabular}[c]{@{}c@{}}Class.\\ Acc (\textcolor{blue}{$\uparrow$})\end{tabular} 
        & \begin{tabular}[c]{@{}c@{}}Rouge\\ Score (\textcolor{red}{$\uparrow$})\end{tabular} 
        & \begin{tabular}[c]{@{}c@{}}Fact.\\ Score (\textcolor{red}{$\uparrow$})\end{tabular} 
        & \begin{tabular}[c]{@{}c@{}}Cloze\\ Acc (\textcolor{teal}{$\uparrow$})\end{tabular} 
        & \begin{tabular}[c]{@{}c@{}}Class.\\ Acc (\textcolor{blue}{$\uparrow$})\end{tabular} 
        & \begin{tabular}[c]{@{}c@{}}Rouge\\ Score (\textcolor{red}{$\uparrow$})\end{tabular} 
        & \begin{tabular}[c]{@{}c@{}}Fact.\\ Score (\textcolor{red}{$\uparrow$})\end{tabular} 
        & \begin{tabular}[c]{@{}c@{}}Cloze\\ Acc (\textcolor{teal}{$\uparrow$})\end{tabular} \\
        \midrule
        \multicolumn{17}{c}{\textbf{LLaVA-13B (5\% Forget Set)}} \\
        \midrule
        Vanilla       & 44.40 & 0.604 & 6.10 & 21.90 & 36.80 & 0.493 & 4.72 & 20.82 & 46.11 & 0.657 & 6.38 & 27.80 & \textbf{68.89} & \textbf{0.633} & \textbf{7.01} & \textbf{26.80} \\
        GA   & \underline{34.04} & \textbf{0.423} & \textbf{2.98} & \textbf{15.19} & 26.35 & 0.386 & \underline{3.74} & \textbf{15.01} & 35.69 & 0.477 & 3.04 & 16.94 & 48.84 & 0.447 & 4.56 & 18.02 \\
        Grad. Diff    & 34.16 & 0.435 & 3.95 & 15.55 & \underline{25.99} & \textbf{0.362} & 3.80 & 16.13 & 37.54 & 0.498 & 3.55 & 18.80 & 49.25 & 0.493 & 4.97 & 18.34 \\
        KL Minim.     & 38.64 & 0.561 & 4.17 & 19.97 & 31.06 & 0.432 & 4.13 & 18.62 & 40.00 & 0.571 & 4.18 & 20.07 & 55.29 & 0.555 & 5.44 & 22.15 \\
        Prompting     & 42.99 & 0.588 & 5.36 & 20.80 & 34.80 & 0.454 & 4.52 & 20.91 & \underline{42.81} & \textbf{0.622} & \textbf{5.23} & \textbf{27.66} & \textbf{67.72} & \textbf{0.633} & \textbf{6.93} & \textbf{25.76} \\
        NPO           & 35.12 & 0.473 & 3.41 & 18.77 & 29.73 & 0.397 & 3.89 & 17.77 & 40.79 & 0.565 & 4.55 & 20.23 & 59.83 & 0.579 & 5.86 & 22.54 \\
        \rowcolor{gray!12}MANU & \textbf{33.65} & \underline{0.434} & \underline{3.13} & \underline{15.33} & \textbf{25.41} & \underline{0.382} & \textbf{3.70} & \underline{16.08} & \textbf{42.88} & \underline{0.593} & \underline{4.72} & \underline{22.75} & \underline{59.97} & \underline{0.585} & \underline{6.08} & \underline{22.80} \\
        \bottomrule
    \end{tabular}}
    \vspace{-0.1in}
    \caption{Overall results of \method with varying pruning ratios on LLaVA-13B model under a 5\% forget data setup. Lower scores on the Forget/Test sets indicate better unlearning, while higher scores on the Retain/Celebrity sets indicate better utility preservation.}
    \label{tab:llava13b_results}
    \vspace{-0.2in}
\end{table*}

To further evaluate the scalability and generalizability of \method to larger MLLMs, we conducted an additional set of experiments using the LLaVA-13B architecture. In this section, we only report results for the 5\% Forget Set split for reference, which required fine-tuning a separate vanilla LLaVA-13B model on the MLLMU-Bench dataset. The quantitative results are presented in Table~\ref{tab:llava13b_results}.

As it shown in the table, the observed performance trends on LLaVA-13B are consistent with those reported for the smaller 7B and 8B variants. Specifically, \method consistently ranks among the top-performing methods across all evaluation tasks. While GA and Gradient Difference occasionally achieve marginally better scores in terms of unlearning effectiveness, these methods generally underperform in preserving model utility. Conversely, the Prompting-based approach demonstrates strong utility preservation but exhibits significantly lower forgetting capability. \method offers a robust compromise, maintaining competitive unlearning performance while preserving downstream utility across all evaluation settings.

These findings reaffirm the central design hypothesis behind \method—that modality-specific importance signals can be effectively extracted and leveraged even within larger, more entangled model architectures. Additionally, we note that italicized emphasis used in our main tables to indicate second-best values may not be easily distinguishable in print. In the revised manuscript, we will replace italics with underlined formatting for improved visual clarity. This extension strengthens the empirical validation of \method and demonstrates its applicability to state-of-the-art, large-scale MLLMs.

\end{document}